%% file: main.tex
\documentclass[a4paper]{article}

\usepackage{cite}
\usepackage{amsmath,amssymb,amsfonts}
\usepackage{algorithmic}
\usepackage{graphicx}
\usepackage{textcomp}
\usepackage{hyperref}

\usepackage{subcaption}
\usepackage{multirow}
\usepackage{adjustbox}
\usepackage{float}
\usepackage{makecell}
\usepackage{xspace}
\usepackage{cleveref}
\usepackage{rotating}
\usepackage{booktabs}
\usepackage[toc,page]{appendix}
\crefname{figure}{Fig.}{Figs.}
\Crefname{figure}{Fig.}{Figs.}
\crefname{table}{Table}{Tables}
\Crefname{table}{Table}{Tables}

\newcommand{\etal}{\textit{et al.}\xspace}

\begin{document}
\title{Evaluation of Language Models in the Medical Context Under Resource-Constrained Settings}
\author{Andrea Posada, Daniel Rueckert, Felix Meissen, and Philip Müller}

\maketitle

\begin{abstract}
Since the Transformer architecture emerged, language model development has grown, driven by their promising potential. Releasing these models into production requires properly understanding their behavior, particularly in sensitive domains like medicine. Despite this need, the medical literature still lacks practical assessment of pre-trained language models, which are especially valuable in settings where only consumer-grade computational resources are available. To address this gap, we have conducted a comprehensive survey of language models in the medical field and evaluated a subset of these for medical text classification and conditional text generation. The subset includes $53$ models with $110$ million to $13$ billion parameters, spanning the Transformer-based model families and knowledge domains. Different approaches are employed for text classification, including zero-shot learning, enabling tuning without the need to train the model. These approaches are helpful in our target settings, where many users of language models find themselves. The results reveal remarkable performance across the tasks and datasets evaluated, underscoring the potential of certain models to contain medical knowledge, even without domain specialization. This study thus advocates for further exploration of model applications in medical contexts, particularly in computational resource-constrained settings, to benefit a wide range of users. The code is available on \url{https://github.com/anpoc/Language-models-in-medicine}.

\end{abstract}

\section{Introduction}
\label{sec:intro}
\input{sections/01_introduction}

\section{Preliminaries}
\label{sec:prelim}
\input{sections/02.1_preliminaries}

\section{Related Work}
\label{sec:relwork}
\input{sections/02.2_relatedwork}

\section{Methodology}
\label{sec:method}
\input{sections/03_methodology}

\section{Results and Discussion}
\label{sec:results}
\input{sections/04_results}

\section{Conclusion}
\label{sec:concl}
\input{sections/05_conclusion}

\bibliographystyle{IEEEtran.bst}
\bibliography{references.bib}

\clearpage
\onecolumn
\begin{appendices}
\section{Data}
\label{app:data}
\input{sections/90_appendix_data}

\clearpage
\section{Models}
\label{app:models}
\input{sections/91_appendix_models}

\clearpage
\section{Prompts}
\label{app:prompts}
\input{sections/92_appendix_prompts}

\clearpage
\section{Supplementary results}
\label{app:supplementary}
\input{sections/93_appendix_supplementary}
\end{appendices}

\end{document}

%% file: sections/01_introduction.tex
Natural language processing (NLP) holds great promise in the medical field. The medical community has recently shown substantial interest in leveraging state-of-the-art language models to address various medical challenges \cite{wiggings, shah}. In particular, generative large language models (LLMs) have showcased emergent abilities beyond their original training objectives, such as text summarization and question answering \cite{wei_ec}. These newfound abilities have enabled LLMs to perform tasks of significant clinical importance, including passing medical examinations, summarizing clinical and radiological reports, as well as medical dialogues, extracting drug names from medical notes, responding to patient inquiries, and writing medical histories and physical assessments \cite{shah, omiye}.

The versatility of language models can be attributed to a convergence of factors \cite{shah, omiye, lievin}. The first factor is their ability to learn valuable patterns within large amounts of unlabeled data via self-supervision. The second factor revolves around the Transformer architecture \cite{transformer} and its suitability for efficient parallel processing on modern computing hardware. Lastly, the third factor encompasses the crucial process of fine-tuning language models to align their responses with human expectations through instruction tuning.

Integration of language models in medical settings is becoming a reality as partnerships between developers and healthcare systems continue to grow \cite{kuling}. The potential benefits are significant, as they can derive broadly applicable representations from extensive medical corpora at scale and encapsulate clinical knowledge \cite{singhal}. Nevertheless, it is essential to recognize that our understanding of the behavior of both small pre-trained and large language models still needs to be completed \cite {omiye}. Deploying these models also carries risks, such as the generation of inaccurate results, a phenomenon known as hallucinations, and the potential amplification of existing biases \cite{wiggings, omiye}. Language models' implementation in sensitive fields, such as healthcare, should therefore be approached with the utmost care \cite{lievin}.

Computing and energy resources required by language models for their development and operation are another critical and limiting factor, especially in LLMs. The standard computing resources available in hospitals are of the consumer-grade type, where it is currently infeasible to handle models with hundreds of billions of parameters. Such resource-constrained settings, i.e., with consumer-grade computing resources, are presented not only by healthcare agents and institutions but also by research groups. 

When large language models do not represent a cost-effective or viable solution, smaller pre-trained language models can be an alternative. LLMs, albeit more massive, have similar architectures and pre-training tasks to smaller pre-trained language models \cite{zhao}. With the same computing budget, a smaller model trained with more high-quality data can perform better than its larger counterparts due to undertraining \cite{hoffmann}. Using curated scientific and biomedical corpora in pre-trained language models has also been effective for discriminative and generative language modeling \cite{liu_rad}. Furthermore, these smaller models align with the crucial imperative of environmental sustainability and open up the possibilities for organizations to develop applications that can run directly on commodity hardware and small devices rather than relying on cloud-based services \cite{phi3}. Language models in resource-constrained settings thereby address practical challenges and have great potential in local computing.

To further understand the performance of language models in clinical scenarios with limited computational resources, we conducted a comprehensive evaluation focusing on the classification and conditional generation of medical texts in open-source models. The datasets employed enable the assessment of general and radiology-specific medical knowledge. In total, $53$ models are tested, ranging from $110$M to $13$B parameters, spanning all Transformer-based model families and knowledge domains from general to clinical. For conditional text generation, solely decoder-only models are used. The approaches adopted for text classification, together with prompt engineering, allow for improved model performance without the need for training or fine-tuning. An analysis of the impact of the prompts on performance is also included. To the best of our knowledge, this is the first work to evaluate such a large number of small pre-trained language models for medical tasks.

%% file: sections/02.1_preliminaries.tex
The evolution of natural language processing can be condensed into four major groups of models: (1) statistical models, (2) neural language models, (3) pre-trained language models, and (4) large language models \cite{zhao}. Each of these groups represents a paradigm shift in natural language modeling and has contributed significantly to the conception of language models as we know them today. 

The first transition, from statistical to neural language models, entailed a shift from word prediction based on minimal local context to probabilistic evaluation of word sequences using neural networks. This transition also introduced the representation of words as low-dimensional continuous embeddings based on their contextual usage (distributional semantics). The second transition, from neural to pre-trained language models, involved turning from developing task-specific models to pre-training and fine-tuning methodologies. The third transition to large language models moved the focus from discriminative AI to generative AI, from model-centric to data-centric approaches, and from fine-tuning to prompt engineering and prompt tuning \cite{zhao, he, tang}. These advances have paved the way for more sophisticated language models with broader applications and improved capabilities.

\subsection{Pre-trained language models}
\begin{figure*}
    \centering
    \begin{subfigure}{.45\textwidth}
        \includegraphics[width=\textwidth]{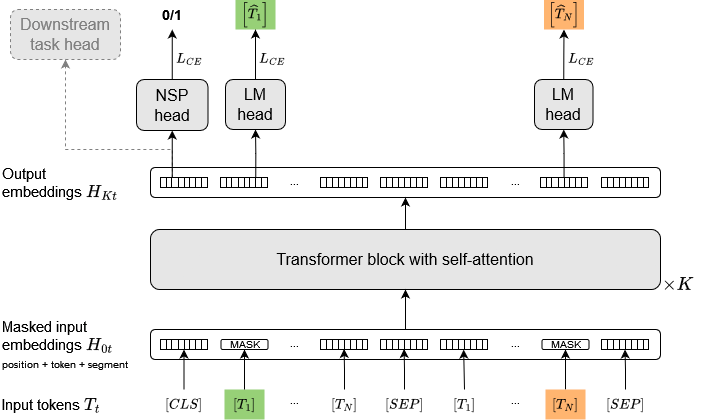}
        \caption{Encoder-only models}
    \end{subfigure}\hspace{25pt}%
    \begin{subfigure}{.45\textwidth}
        \includegraphics[width=\textwidth]{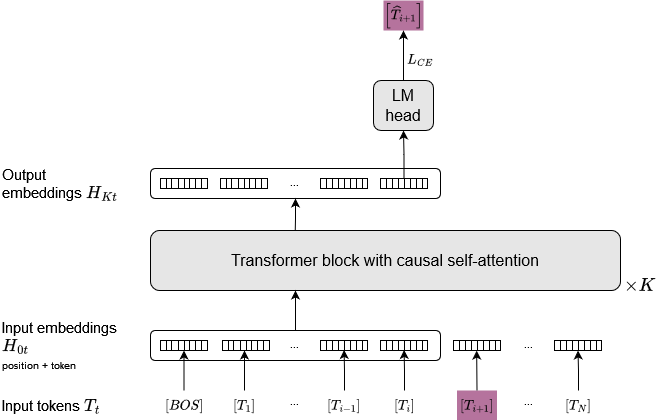}
        \caption{Decoder-only models}
    \end{subfigure}
    \par\bigskip
    \begin{subfigure}{.90\textwidth}
        \includegraphics[width=\textwidth]{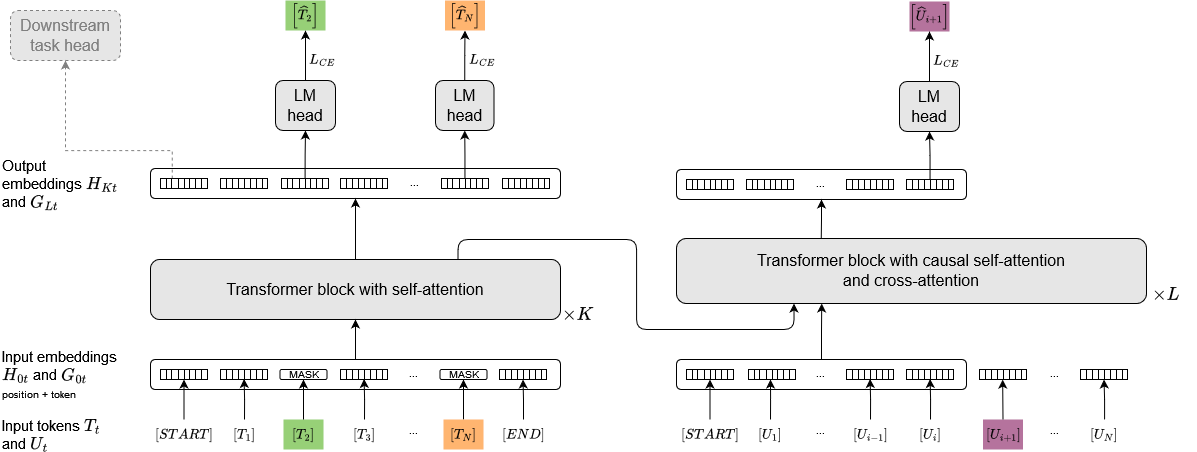}
        \caption{Encoder-decoder models}
    \end{subfigure}
    \caption{Graphical representation of the three families of Transformer-based models: encoder-only, decoder-only, and encoder-decoder models. Colors signal the correspondence between outputs and targets. Encoder-only models are mainly used for discriminative tasks. Their input is tokenized, and some of these tokens are masked. They are then fed into Transformer blocks with self-attention to obtain contextualized output embeddings, which are further processed by next sentence prediction (NSP) and language model (LM) heads or used by downstream task-specific heads. Depending on the training objective, the NSP head may or may not be necessary. Decoder-only models focus on generation tasks. Their input is tokenized and fed to Transformer blocks with causal self-attention. The causal self-attention ensures that the information flows unidirectionally from left to right. Encoder-decoder models are used for text-to-text tasks. Their encoder processes the input text, similar to encoder-only models but excluding the NSP head, and flows information to the decoder via the cross-attention mechanism. This information is used with the target output so that the decoder learns to produce the latter generatively.}
    \label{fig:transformerLM}
\end{figure*}

The emergence of pre-trained language models represented a paradigm shift, driving research toward designing more efficient architectures and refining pre-training strategies. These pre-trained models have been commonly adapted or specialized to downstream tasks via fine-tuning, which involves transferring knowledge by further training a model on new data. There are significant advantages demonstrated by these models in language understanding and model performance in various tasks \cite{he, zhao}.

ELMo is one of the earliest attempts at pre-trained language models\cite{elmo}. This model was developed to capture context-aware word representations by pre-training a bidirectional Long Short-Term Memory (biLSTM) network and fine-tuning it for subsequent downstream tasks. Later the Transformer architecture was introduced, revolutionizing the NLP field by offering highly parallelizable structures and self-attention mechanisms. The Transformer \cite{transformer} follows the autoencoder archetype, from which three families of models arose: (1) BERT-family or encoder-only models, (2) GPT-family or decoder-only models, and (3) text-to-text or encoder-decoder models. In \cref{fig:transformerLM}, the graphical representations of these families are shown.

\subsubsection{Encoder-only models}
Encoder-only models, exemplified by BERT \cite{bert}, are based on masked language modeling (MLM), where parts of the input are masked to encourage the model to reconstruct the original sequence, leveraging contextual information bidirectionally. These models can be stated as $v_{1:n} \rightarrow \phi(v_{1:n})$. In particular, their contextual embeddings have been proven highly effective as general-purpose semantic features, significantly boosting performance in discriminative NLP tasks.

\subsubsection{Decoder-only models}
Decoder-only models focus on autoregressive language modeling, i.e., predicting the next token in a sequence based on previous tokens. These models produce contextual embeddings and distribution over the subsequent tokens \(v_{i+1}\), which can be stated as $v_{1:i} \rightarrow \phi(v_{1:i}), \mathbb{P}(v_{i+1}|v_{1:i})$. However, the contextual embeddings they generate depend solely on the left context. Most research efforts are currently directed toward decoder-only models due to their exceptional performance in conditional generation tasks and their demonstrated emergent capabilities.

\subsubsection{Encoder-decoder models}
Text-to-text models, or encoder-decoder models, are trained to learn the correspondence between a pair of texts and can be stated as $v_{1:n} \rightarrow \phi(v_{1:n}), \mathbb{P}(w_{1:m}|\phi(v_{1:n}))$. These models combine bidirectional contextual embeddings with the capability to generate output sequences, making them versatile in various text-to-text tasks without requiring additional heads for fine-tuning. Moreover, by having a broad spectrum of language tasks that can be translated into text-to-text representation, these models can potentially be used for a wide range of applications.

\subsection{Large language models}
Scaling of language models has often resulted in improved model capabilities in various tasks \cite{flant5, hoffmann, brown, gpt2, palm, rae, wei}, including those requiring specialized scientific knowledge and reasoning \cite{hendrycks}. Research by Kaplan \textit{et al.} \cite{scalelaw} revealed that there is an empirical power-law relationship between the language model performance, in terms of cross-entropy loss, and the model size, dataset size, and amount of compute used for training. It was further found that architectural details, such as network width or depth, had minimal effects on performance. Scaling laws have further been studied by Hoffmann \textit{et al.} \cite{hoffmann} and Bahri \textit{et al.} \cite{scalelaw2}.

Following these empirical results, several studies have trained progressively larger language models of up to hundreds of billion parameters, such as GPT-3 \cite{brown}, PaLM \cite{palm}, Galactica \cite{galactica}, LLaMA \cite{llama,llama2}, Claude \cite{claude}, Gemini 1.5 \cite{gemini}, and Mistral \cite{mistral}. Among all, GPT-3 and ChatGPT can be considered the precursors of the large language models, the name by which these large-scale language models are known \cite{he, zhao}. GPT-4, a latter version of GPT-3, stands out for its exceptional performance, often matching or surpassing human performance on a variety of tasks \cite{liu_rad, bubeck, nori}, even in specialized domains \cite{mao}. Extensive evaluations have been conducted to GPT-4 \cite{hendrycks, lopez, liang, liu__}, exploring even the path toward Artificial General Intelligence (AGI) \cite{bubeck}.

LLMs can be adapted to different tasks via prompt engineering, which, unlike fine-tuning, does not require retraining the model and updating its weights. These prompting techniques have led to observing unexpected emergent capabilities in LLMs, demonstrating the potential to address a wide range of complex tasks and exhibit apparent reasoning abilities \cite{schaeffer, wei_ec, singhal, tang, agrawal, brown, wei, sanh, lampinen, kojima}. In the medical field, for example, Chain of Thought (CoT) has been used for explainability \cite{wei_cot} and in-context learning to mitigate the need for costly medical annotations \cite{he}. Numerous studies have even highlighted the competence of large language models as implicit knowledge bases \cite{singhal, hendrycks, galactica, joshi}. 

In-context learning techniques, such as zero-shot and few-shot learning, have also proven to be remarkably effective on instruction-tuned models and models to which reinforcement learning techniques have been applied \cite{wei, agrawal, singhal, rev_scarcity}. Zero-shot learning consists of asking a trained model to complete a task without providing explicit examples of this task, whereas in few-shot learning, some examples are provided. Nonetheless, prompting techniques are not exclusive to LLMs but are also applicable to smaller pre-trained language models, especially encoder-decoder and decoder-only models.

Despite their advantages, LLMs also have limitations. Their high computational resource requirements and associated computational challenges represent a major limitation of these models. For example, conducting studies to better understand the behavior of LLMs and assess important criteria, such as faithfulness and biases, can be costly and time-consuming. The detection biases and hallucinations, i.e., generating inaccurate results, is crucial in sensitive domains such as medicine. 

Due to the significance of the computing limitation, alternatives such as model quantization \cite{quantization} have been introduced. Quantization is a technique that reduces the computational and memory costs of model inference by representing its weights and activations with low-precision data types, such as 8-bit integers, instead of the usual 32-bit floating point. In natural language processing, this technique is currently being extensively studied, with \cite{quant1,quant2,quant3,quant4,quant5,quant6} being some examples in the literature. 

Recommendations on the optimal use of computational resources have also been proposed. Chinchilla's scaling law \cite{hoffmann}, one of these recommendations, states that the optimal model size and the number of tokens for training a language model should scale equally for compute-optimal training under a given computational budget. In \cite{hoffmann}, it is further proved that current large language models are significantly undertrained due to the recent focus on scaling language models while keeping the amount of training data constant. A smaller model trained with more high-quality data can thus achieve better performance than its larger counterparts with the same computing budget.

\subsection{Language models in the biomedical/clinical context}
Broadly speaking, language models used in specialized domains are (i) trained models solely on target domain data, (ii) pre-trained models on general domain corpus with tuning strategies, and (iii) pre-trained models on specialized domain corpus with(out) tuning strategies. Examples of tuning strategies are fine-tuning with target domain data and prompt engineering. Pre-training on (ii) can also be domain-adaptive continual pre-training (i.e., pre-trained on specialized domain corpus after pre-training on general domain corpus) or mixed-domain pre-training (i.e., pre-trained on a mix of general and specialized domain corpus, simultaneously).

GPT-4 is an example of a general domain language model that has been studied in medical applications. Research has covered from its utility as a medical chatbot \cite{lee} and in medical competency exams \cite{nori} to its applications in radiology \cite{fink, liu_rad, lyu, adams, bhayana, wu, ranjit}, among others \cite{mesko, gala, atallah, cheng}. Nevertheless, the models studied in the medical context are mostly domain-specific, either biomedical or clinical. These models include pre-trained models such as BioBERT \cite{biobert}, SciBERT \cite{scibert}, BioMedBERT \cite{pubmedbert}, BioMegatron \cite{biomegatron}, ScholarBERT \cite{scholarbert}, BioGPT \cite{biogpt}, and ClinicalBERT \cite{bioclinicalbert}; as well as large language models as Galactica \cite{galactica}, MedAlpaca \cite{medalpaca}, PMC-LLaMA \cite{pmcllama}, Med-PaLM 2 \cite{medpalm2}, GatorTron \cite{gatortron}, GatorTronGPT \cite{gatortrongpt}, and ClinicalGPT \cite{clinicalgpt}. 

Domain-specific models usually contain general domain data within the pre-training data, with exceptions such as BioMedBERT, Galactica, GatorTron, and GatorTronGPT. For large language models, instruction fine-tuning is the most common tuning technique, as in MedAlpaca, Med-PaLM 2, GatorTronGPT, and ClinicalGPT. Reinforcement learning from human feedback (RLHF) and reinforcement learning from AI feedback (RLAIF) have also been adopted, although less frequently, being HuatuoGPT \cite{huatuogpt} an example of this. Recent research studies indicate as well a multimodal trend that supports various types of healthcare data, including electronic health records (EHR), medical images, and medical sequence signals. Examples of these developments include LLaVAMed \cite{llavamed}, MedAGI \cite{medagi}, OphGLM \cite{ophglm}, Visual Med-Alpaca \cite{visualmedalpaca}, MedFlamingo \cite{medflaming}, and CheXzero \cite{chexzero}.

%% file: sections/02.2_relatedwork.tex
Comparative studies investigating language models are crucial to advance our understanding of them, shed light on their functionalities and pinpoint their constraints. Despite previous research, a notable gap persists in the literature due to, among other cause, current pace of development in NLP. This gap is particularly significant in fields that require heightened sensitivity, such as medicine, where a thorough understanding of models is imperative \cite{rev_scarcity}. Existing research in medicine is mainly focused on specific tasks or datasets or models \cite{lievin, agrawal, tang, liu_rad, singhal}. Moreover, most of the discursive and practical assessments focus on LLMs, as can be seen below. To the best of our knowledge, there is no practical assessment in the clinical context that includes a wide number of pre-trained models, covering all Transformer-based model families, targeting settings where only consumer grade computing resources are available.

The work by He \etal \cite{he} stands out among exiting descriptive studies, comprehensively addressing the capabilities, limitations, development and integration of language models in healthcare. The language models in scope are pre-trained and large language models. The development process is explained in detail, covering aspects such as training data, methodologies, and optimization strategies. Concerns related to the integration of LLMs into healthcare are also investigated, as fairness, accountability, transparency, and ethics.

Zhou \etal \cite{zhou} also provide a comprehensive overview of the development and deployment of LLMs in medicine, together with the challenges and opportunities these models face. Their study is both discursive and practical, being one of its highlights. The authors detail the principles of existing medical LLMs, comprising basic model structures, number of parameters, and data sources and scales used for model development. A comparison of the performance of different LLMs across various medical tasks, also against state-of-the-art lightweight models, is also included. 

Continuing with practical reviews, Soni \etal \cite{soni} assessed the cost-effectiveness of pre-training and fine-tuning in BERT, BioBERT, Clinical BERT, and XLNet for medical question answering tasks. Their results indicate that BERT-based models exhibit superior performance when fine-tuned with mixed datasets (i.e., general and clinical domain data), highlighting a gap in well-generalizable medical QA datasets. The results also suggest that initial fine-tuning on general domain datasets, such as SQuAD, before doing it on clinical datasets can enhance performance. Prompting techniques were not included in their evaluations.

In a similar vein, Jahan \textit{et al.} \cite{rev_scarcity} studied the impact of data size for fine-tuning and that of prompts in zero-shot learning on model performance. Four large language models are evaluated on six benchmark biomedical text processing tasks across $26$ datasets. Zero-shot LLMs outperform state-of-the-art fine-tuned models, such as BioBERT, BioGPT, and BioBART, when fine-tuning data is scarce. As the amount of fine-tuning data increases, so does the performance of these state-of-the-art fine-tuned models, surpassing zero-shot LLMs. The study also highlights LLMs' sensitivity to prompts, as variations in these led to significant differences in outcomes. No single LLM consistently excelled across all datasets and tasks. The authors advocate the training of biomedical LLMs on domain-specific corpora while recognizing LLMs' potential for biomedical applications that lack large annotated data.

Lehman \etal \cite{lehman} further explored whether LLMs trained primarily on general web text are suitable for highly specialized, safety-critical domains such as medicine, or if domain-specific models are a better alternative. A total of $12$ language models, ranging from $220$ million to $175$ billion parameters, are evaluated on three clinical tasks. As part of the experiments, T5 models were trained from scratch using MIMIC-III and MIMIC-IV clinical notes to investigate the efficiency of clinical tokens. Their findings suggest that relatively small, specialized clinical models significantly outperform all in-context learning approaches, even when fine-tuned on limited annotated data. Neither the models' ability to handle long texts nor decoder-only and instruction-tuned models are accounted for in their work.

Lastly, Li \etal \cite{rev_long} focuses on pre-trained language models for long clinical text. A core limitation of Transformer-based models is their substantial memory consumption, leading to performance degradation in long clinical texts. To overcome this limitation, the authors pre-trained Longformer and BigBird, two long-sequence Transformers, on a large-scale clinical corpus, extending the maximum input length from $512$ to $4\,096$. These models consistently and significantly outperformed ClinicalBERT and other short sequence Transformers across ten tasks. Long-sequence Transformers enriched with clinical knowledge are thus capable of learning long-term dependencies in long clinical texts according to the results. No generative tasks and solely encoder-only models are considered in their evaluations.

%% file: sections/03_methodology.tex
A series of experiments on medical text classification and conditional text generation are carried out to understand better the behavior of language models under resource-constrained settings, i.e., settings with consumer-grade computing resources. In total, $53$ language models are evaluated, whose size ranges from $110$ million to $13$ billion parameters. The selection of these models spans the general, biomedical, and clinical knowledge domains and includes the three families of Transformer-based models. Moreover, only open-source, smaller than $13$ B parameters models are considered. Details on the selected models are found in Table \ref{tab:models} and \Cref{app:models}.

All experiments are performed using a Quadro RTX 8000 GPU and CUDA version 12.2. To guarantee that the selected models align with consumer-grade computing resources, models with more than $8$ billion parameters (i.e., \texttt{OpenLLaMA 13B}, \texttt{Flan-T5-XXL}, \texttt{T5-V1.1-11B}, and \texttt{T0++}) are run with float$16$ precision. By halving the floating-point precision, these $11$ and $13$ billion parameter model versions are still viable in computational resource-constrained settings.

The three families of Transformer-based models are considered for the text classification task via different approaches (described in \Cref{subsec:approaches}), whereas solely decoder-only models are used for the conditional text generation task. Transcriptions, MIMIC-CXR, and MS-CXR have been chosen as evaluation datasets. Transcriptions covers a broad spectrum of medical specialties, allowing a general assessment of medical knowledge. MIMIC-CXR and its labeled version, MS-CXR, enable testing focused on radiology, one of the most promising fields for AI integration, narrowing the evaluation to specialized medical knowledge.

\begin{table*}
\centering
\rotatebox{90}{
\begin{minipage}{.99\textheight}
\caption{The models used in this study are categorized by their type, domain, and size. Each model is presented with its number of parameters and may have one or more superscripts. Superscripts are 0: model used for contextual embedding similarity, 1: model used for natural language inference (NLI), 2: model used for multiple-choice questions, 3: model used for text generation, \dag: instruction-tuned model, \ddag: cross-encoder model.} 
\label{tab:models}
\begin{adjustbox}{width=.99\textheight}
\begin{tabular}{llllrllrllrllrllr}
\midrule
 &  & \multicolumn{3}{c}{\textbf{Small (S)}} & \multicolumn{3}{c}{\textbf{Medium (M)}} & \multicolumn{3}{c}{\textbf{Large (L)}} & \multicolumn{3}{c}{\textbf{XL}} & \multicolumn{3}{c}{\textbf{XXL}}\\ \cmidrule(lr){3-5} \cmidrule(lr){6-8} \cmidrule(lr){9-11} \cmidrule(lr){12-14} \cmidrule(l){15-17}
 &  & \multicolumn{1}{c}{\textbf{ID}} & \multicolumn{1}{c}{\textbf{Model}} & \multicolumn{1}{c}{\textbf{Size}} & \multicolumn{1}{c}{\textbf{ID}} & \multicolumn{1}{c}{\textbf{Model}} & \multicolumn{1}{c}{\textbf{Size}} & \multicolumn{1}{c}{\textbf{ID}} & \multicolumn{1}{c}{\textbf{Model}} & \multicolumn{1}{c}{\textbf{Size}} & \multicolumn{1}{c}{\textbf{ID}} & \multicolumn{1}{c}{\textbf{Model}} & \multicolumn{1}{c}{\textbf{Size}} & \multicolumn{1}{c}{\textbf{ID}} & \multicolumn{1}{c}{\textbf{Model}} & \multicolumn{1}{c}{\textbf{Size}}\\ \midrule

\multirow{10}{*}{\textbf{Encoder-only}} & \textbf{General} & m00 & \texttt{BERT$_{\texttt{BASE}}$} \textsuperscript{0} \cite{bert} & 110 M & m01 & \texttt{BERT$_{\texttt{LARGE}}$} \textsuperscript{0} \cite{bert} & 340 M & \multicolumn{1}{c}{\multirow{2}{*}{-}} & \multicolumn{1}{c}{\multirow{2}{*}{-}} & \multicolumn{1}{c}{\multirow{2}{*}{-}} & \multicolumn{1}{c}{\multirow{2}{*}{-}} & \multicolumn{1}{c}{\multirow{2}{*}{-}} & \multicolumn{1}{c}{\multirow{2}{*}{-}} & \multicolumn{1}{c}{\multirow{2}{*}{-}} & \multicolumn{1}{c}{\multirow{2}{*}{-}} & \multicolumn{1}{c}{\multirow{2}{*}{-}} \\ 
  &  & m11 & \texttt{NLI-DeBERTa$_\texttt{base}$}  \textsuperscript{\ddag1} \cite{deberta} & 100 M  &  m12 & \texttt{RoBERTa$_\texttt{LARGE}$-MNLI} \textsuperscript{\ddag1} \cite{roberta} & 355 M  &  &  &  &  &  &  &  &  &  \\ \cmidrule(lr){3-5} \cmidrule(lr){6-8} \cmidrule(lr){9-11} \cmidrule(lr){12-14} \cmidrule(l){15-17}
 & \textbf{Biomedical} & m02 & \texttt{BiomedBERT} & 110 M & m04 & \texttt{BiomedBERT-large} & 340 M & \multicolumn{1}{c}{\multirow{9}{*}{-}} & \multicolumn{1}{c}{\multirow{9}{*}{-}} & \multicolumn{1}{c}{\multirow{8}{*}{-}} & \multicolumn{1}{c}{\multirow{8}{*}{-}} & \multicolumn{1}{c}{\multirow{8}{*}{-}} & \multicolumn{1}{c}{\multirow{8}{*}{-}} & \multicolumn{1}{c}{\multirow{8}{*}{-}} & \multicolumn{1}{c}{\multirow{8}{*}{-}} & \multicolumn{1}{c}{\multirow{8}{*}{-}} \\
  &  &  & \texttt{(abstracts + full text)} \textsuperscript{0} \cite{pubmedbert} &  &  & \texttt{(abstracts only)} \textsuperscript{0} \cite{pubmedbert} &  &  &  &  &  &  &  &  &  &  \\
 &  &  m03 &  \texttt{BiomedBERT} & 110 M &  &  &  &  &  &  &  &  &  &  &  &  \\
  &  &  & \texttt{(abstracts only)} \textsuperscript{0} \cite{pubmedbert} &  &  &  &  &  &  &  &  &  &  &  &  &  \\
 &  & m05 & \texttt{SciBERT} \textsuperscript{0} \cite{scibert} & 110 M &  &  &  &  &  &  &  &  &  &  &  &  \\
 &  & m06 & \texttt{SapBERT} \textsuperscript{0} \cite{sapbert} & 110 M &  &  &  &  &  &  &  &  &  &  &  &  \\
 &  & m07 & \texttt{BioLORD-STAMB2-v1} \textsuperscript{0} \cite{biolord} & 110 M &  &  &  &  &  &  &  &  &  &  &  &  \\
 &  & m08 & \texttt{BioLORD-STAMB2-v1-STS2} \textsuperscript{0} \cite{biolord} & 110 M &  &  &  &  &  &  &  &  &  &  &  &  \\
 &  & m09 & \texttt{BioLORD-PMB} \textsuperscript{0} \cite{biolord} & 110 M &  &  &  &  &  &  &  &  &  &  &  &  \\ \cmidrule(lr){3-5} \cmidrule(lr){6-8} \cmidrule(lr){9-11} \cmidrule(lr){12-14} \cmidrule(l){15-17}
 & \textbf{Clinical} & m10 & \texttt{Bio+Clinical BERT} \textsuperscript{0} \cite{bioclinicalbert} & 110 M & \multicolumn{1}{c}{-} & \multicolumn{1}{c}{-} & \multicolumn{1}{c}{-} & \multicolumn{1}{c}{-} & \multicolumn{1}{c}{-} & \multicolumn{1}{c}{-} & \multicolumn{1}{c}{-} & \multicolumn{1}{c}{-} & \multicolumn{1}{c}{-} & \multicolumn{1}{c}{-} & \multicolumn{1}{c}{-} & \multicolumn{1}{c}{-}\\ \midrule

\multirow{5}{*}{\parbox{2cm}{\textbf{Encoder-\\decoder}}} & \textbf{General} & m14 & \texttt{T5-V1.1-Base} \textsuperscript{2} \cite{t5, t5v1} & 220 M &  m13 &  \texttt{BART Large-MNLI} \textsuperscript{1} \cite{bart} & 407 M & m15 & \texttt{T5-V1.1-Large} \textsuperscript{2} \cite{t5, t5v1} & 770 M & m16 & \texttt{T5-V1.1-3B} \textsuperscript{2} \cite{t5, t5v1} & 3.0 B & m17 & \texttt{T5-V1.1-11B} \textsuperscript{2} \cite{t5, t5v1} & 11.0 B \\
 &  & m18 & \texttt{Flan-T5-Base} \textsuperscript{\dag2} \cite{flant5} & 220 M &  &  &  & m19 & \texttt{Flan-T5-Large} \textsuperscript{\dag2} \cite{flant5} & 770 M & m20 & \texttt{Flan-T5-XL} \textsuperscript{\dag2} \cite{flant5} & 3.0 B & m21 & \texttt{Flan-T5-XLL} \textsuperscript{\dag2} \cite{flant5} & 11.0 B \\
 &  &  &  &  &  &  &  &  &  &  & m22 & \texttt{T0 3B} \textsuperscript{\dag2} \cite{t0} & 3.0 B & m23 & \texttt{T0++} \textsuperscript{\dag2} \cite{t0} & 11.0 B \\ \cmidrule(lr){3-5} \cmidrule(lr){6-8} \cmidrule(lr){9-11} \cmidrule(lr){12-14} \cmidrule(l){15-17}
 & \textbf{Biomedical} & \multicolumn{1}{c}{-} & \multicolumn{1}{c}{-} & \multicolumn{1}{c}{-} & \multicolumn{1}{c}{-} & \multicolumn{1}{c}{-} & \multicolumn{1}{c}{-} & \multicolumn{1}{c}{-} & \multicolumn{1}{c}{-} & \multicolumn{1}{c}{-} & \multicolumn{1}{c}{-} & \multicolumn{1}{c}{-} & \multicolumn{1}{c}{-} & \multicolumn{1}{c}{-} & \multicolumn{1}{c}{-} & \multicolumn{1}{c}{-}\\ \cmidrule(lr){3-5} \cmidrule(lr){6-8} \cmidrule(lr){9-11} \cmidrule(lr){12-14} \cmidrule(l){15-17}
 & \textbf{Clinical} & m24 & \texttt{ClinicalT5-base} \textsuperscript{2} \cite{clinicalt5} & 220 M  & \multicolumn{1}{c}{-} & \multicolumn{1}{c}{-} & \multicolumn{1}{c}{-} & m25 & \texttt{ClinicalT5-large} \textsuperscript{2} \cite{clinicalt5} & 700 M & \multicolumn{1}{c}{-} & \multicolumn{1}{c}{-} & \multicolumn{1}{c}{-} & \multicolumn{1}{c}{-} & \multicolumn{1}{c}{-} & \multicolumn{1}{c}{-}\\ \midrule

\multirow{18}{*}{\textbf{Decoder-only}} & \textbf{General} & \multicolumn{1}{c}{\multirow{15}{*}{-}} & \multicolumn{1}{c}{\multirow{15}{*}{-}} & \multicolumn{1}{c}{\multirow{15}{*}{-}} & m26 & \texttt{GPT-2 Medium}\textsuperscript{3} \cite{gpt2} & 355 M & m27 &  \texttt{GPT-2 Large} \textsuperscript{3} \cite{gpt2} & 774 M & m28 & \texttt{GPT-2 XL} \textsuperscript{3} \cite{gpt2} & 1.5 B & m29 & \texttt{Palmyra Base 5B} \textsuperscript{23} \cite{palmyra} & 5.0 B \\
 &  &  &  &  &  &  &  &  &  &  & m41 & \texttt{OpenLLaMA 3B} \textsuperscript{3} \cite{openllama} & 3.0 B & m30 & \texttt{Camel 5B}\textsuperscript{\dag2} \cite{camel} & 5.0 B \\
 &  &  &  &  &  &  &  &  &  &  & m42 & \texttt{OpenLLaMA 3Bv2} \textsuperscript{3} \cite{openllama} & 3.0 B & m31 & \texttt{GPT-J 6B} \textsuperscript{23} \cite{gptj} & 6.0 B \\
 &  &  &  &  &  &  &  &  &  &  &  &  &  & m32 & \texttt{Instruct GPT-J} \textsuperscript{\dag2} \cite{igptj} & 6.0 B \\
 &  &  &  &  &  &  &  &  &  &  &  &  &  & m33 & \texttt{Falcon-7B} \textsuperscript{23} \cite{falcon} & 7.0 B \\
 &  &  &  &  &  &  &  &  &  &  &  &  &  & m34 & \texttt{Falcon-7B-Instruct} \textsuperscript{\dag2} \cite{falcon} & 7.0 B \\
 &  &  &  &  &  &  &  &  &  &  &  &  &  & m35 & \texttt{MPT-7B} \textsuperscript{23} \cite{MPT} & 7.0 B \\
 &  &  &  &  &  &  &  &  &  &  &  &  &  & m36 & \texttt{MPT-7B-Instruct} \textsuperscript{\dag2} \cite{MPT} & 7.0 B \\
 &  &  &  &  &  &  &  &  &  &  &  &  &  & m37 & \texttt{LLaMA-7B} \textsuperscript{23} \cite{llama} & 7.0 B \\
 &  &  &  &  &  &  &  &  &  &  &  &  &  & m38 & \texttt{LLaMA 2-7B} \textsuperscript{23} \cite{llama2} & 7.0 B \\
 &  &  &  &  &  &  &  &  &  &  &  &  &  & m39 & \texttt{Alpaca 7B} \textsuperscript{\dag2} \cite{alpaca} & 7.0 B \\
 &  &  &  &  &  &  &  &  &  &  &  &  &  & m40 & \texttt{LLaMA 2-CHAT-7B} \textsuperscript{\dag2} \cite{llama2} & 7.0 B \\
 &  &  &  &  &  &  &  &  &  &  &  &  &  & m43 & \texttt{OpenLLaMA 7B} \textsuperscript{3} \cite{openllama} & 7.0 B \\
 &  &  &  &  &  &  &  &  &  &  &  &  &  & m44 & \texttt{OpenLLaMA 7Bv2} \textsuperscript{3} \cite{openllama} & 7.0 B \\
 &  &  &  &  &  &  &  &  &  &  &  &  &  & m45 & \texttt{OpenLLaMA 13B} \textsuperscript{3} \cite{openllama} & 13.0 B \\ \cmidrule(lr){3-5} \cmidrule(lr){6-8} \cmidrule(lr){9-11} \cmidrule(lr){12-14} \cmidrule(l){15-17}
 & \parbox{2cm}{\textbf{Biomedical\\/ Scientific}} & \multicolumn{1}{c}{\multirow{2}{*}{-}} & \multicolumn{1}{c}{\multirow{2}{*}{-}} & \multicolumn{1}{c}{\multirow{2}{*}{-}} & m48 & \texttt{BioGPT} \textsuperscript{3} \cite{biogpt} & 347 M & m47 & \texttt{GPT-2-PubMed Large} \textsuperscript{3} \cite{gpt2p} & 774 M & m50 & \texttt{Galactica 1.3B} \textsuperscript{3} \cite{galactica} & 1.3 B & m51 & \texttt{Galactica 6.7B} \textsuperscript{3} \cite{galactica} & 6.7 B \\
 &  &  &  &  & m46 & \texttt{GPT-2-PubMed Medium} \textsuperscript{3} \cite{gpt2p} & 355 M &  &  &  & m49 & \texttt{BioGPT-Large} \textsuperscript{3} \cite{biogpt} & 1.5 B &  &  &  \\ \cmidrule(lr){3-5} \cmidrule(lr){6-8} \cmidrule(lr){9-11} \cmidrule(lr){12-14} \cmidrule(l){15-17}
 & \textbf{Clinical} & \multicolumn{1}{c}{-} & \multicolumn{1}{c}{-} & \multicolumn{1}{c}{-} & \multicolumn{1}{c}{-} & \multicolumn{1}{c}{-} & \multicolumn{1}{c}{-} & \multicolumn{1}{c}{-} & \multicolumn{1}{c}{-} & \multicolumn{1}{c}{-} & \multicolumn{1}{c}{-} & \multicolumn{1}{c}{-} & \multicolumn{1}{c}{-} & m52 & \texttt{MedAlpaca 7b} \textsuperscript{\dag2} \cite{medalpaca} & 7.0 B \\ \midrule
\end{tabular}
\end{adjustbox}
\end{minipage}
}
\end{table*}

\subsection{Text classification}
Text classification is addressed using the Transcriptions and MS-CXR datasets and three different approaches: (i) contextual embedding similarity, (ii) natural language inference (NLI), and (iii) multiple-choice question answering (MCQA). The contextual embedding similarity approach is intended for encoder-only models, the NLI approach for encoder-only and encoder-decoder models pre-trained for NLI, and the MCQA approach for encoder-decoder and decoder-only models..

Model tuning is implemented through zero-shot learning. To analyze the impact of prompting on text classification performance, different prompts are applied during inference. These prompts, grouped into two sets, are defined according to the classification approach. The first set of prompts is used for contextual embedding similarity and NLI. Since neither of these approaches requires a prompt to work, its non-use is also included in the analysis. The second set of prompts is used for MCQA, approach that needs a prompt to work. Prompts from the second set are defined based on those most commonly used in instruction-tuning models for multiple-choice question answering tasks.

Let $x \in X$ be a text sample and $y \in Y$ be a class, not necessarily corresponding to $x$. A prompt from the first prompt set, $p \in P_1$, is defined as a function of a prompt template and a label. For example, $p_1(y)=$ ``This is an example of $y$''. The set $P_1$ is only applied to the classes. Meanwhile, a prompt from the second prompt set, $p \in P_2$, combines a prompt template (consisting of the prompt structure and a question), a text sample, and the classes. For example, $p(x,Y)=$ ``You are a doctor and have the following information about a patient from a chest x-ray: $x$. What is the diagnosis? $Y$. (''. In this example, the prompt template consists of the question ``What is the diagnosis?'' and the prompt structure, which is the rest of the text. Prompts are presented in detail in \Cref{app:prompts}.

\subsubsection{Datasets} The datasets evaluated in text classification are Transcriptions and MS-CXR. Each of these datasets is introduced below. Their preprocessing and characterization details are given in \Cref{app:data}.

\noindent\textbf{Transcriptions} is a multi-label collection of electronic health records (EHRs) covering many medical specialties. Preprocessing is applied to the data, removing null entries, organizing the EHR format, and selecting the final set of labels. After all, $2\,074$ samples and $29$ classes are available. Performance is measured by the AUC score since the dataset is multi-label.

Due to the length of some EHRs, certain token vectors exceed the maximum input length allowed by some models. To cope with this limit, the input sequence is processed using a non-overlapping sliding window method \cite{rev_long}, as detailed in \Cref{subsec:approaches}. 

\noindent\textbf{MS-CXR} is a multi-class dataset composed of X-ray report sections, each accompanied by annotations made by a radiologist \cite{ms-cxr-1, ms-cxr-2, PhysioNet}. There are $718$ unique samples representing eight well-distributed classes. Preprocessing of this dataset consists of removing samples with missing information and duplicates. Contrary to Transcriptions, no sample exceeds the maximum allowed input length for any of the models. Performance is measured by accuracy, F1-score, precision, and recall in their macro-averaged version to ensure a comprehensive assessment.

\subsubsection{Approaches}
\label{subsec:approaches}
Text classification is performed through (i) contextual embedding similarity, (ii) natural language inference, and (iii) multiple-choice question answering.

\noindent\textbf{Contextual embedding similarity} is grounded in the cosine similarity between the contextual embeddings of the sample text and the classes. The contextual or sentence embedding is determined by three distinct pooling strategies: CLS-token embedding, average token-level embedding pooling, and maximum token-level embedding pooling.

For this approach, encoder-only models are employed, with a total of $11$ models evaluated. These models have a maximum input token size of $512$ tokens. Therefore, the samples' token vectors that exceed this limit are processed with the non-overlapping sliding window method. The fragments are aggregated according to the pooling strategy, as follows.

\begin{itemize}
\item \textit{CLS pooling:} The contextual embedding of each fragment from a sample is computed as its CLS-token output embedding. These embeddings are then aggregated using the element-wise average to obtain the contextual embedding representing the sample.

\item \textit{Maximum pooling:} The contextual embedding of each fragment from a sample is computed by applying element-wise maximum at token level over the output embeddings. These embeddings are then aggregated using again the element-wise maximum to obtain the contextual embedding representing the sample.

\item \textit{Average pooling:} The contextual embedding of each fragment from a sample is computed by applying the element-wise average at the token level over the output embeddings. These embeddings are then aggregated using the element-wise weighted average to obtain the contextual embedding representing the sample. Average's weights indicate the number of non-padding tokens in each fragment.
\end{itemize}

\noindent\textbf{Natural language inference} is the task of determining whether a hypothesis is true (entailment), false (contradiction), or indeterminate (neutral) given a premise. When applied for text classification, the premise represents a test sample, and the hypothesis represents the classes. For multi-class datasets, the predicted label is calculated from the entailment logits of each hypothesized class. For multi-label datasets, the entailment and contradiction logits are transformed into binary probabilities, which indicate whether or not a particular hypothesized class is predicted. This could be viewed as having $n$ binary text classifiers, where $n$ is the number of classes.

This approach employs encoder-only (cross-encoder) and encoder-decoder models. These models have a lower maximum input token size than some of the test samples. Therefore, the token vectors of these particular samples are processed with the non-overlapping sliding window method. They are divided into fragments, whose scores are calculated individually, and then these scores are averaged to get the score of the whole sample.

\noindent\textbf{Multiple-choice question answering} enables generative models, i.e., encoder-decoder and decoder-only models, to perform text classification. A total of $27$ models are assessed in this approach, including both pre-trained models and their instruction-tuned versions. 

As multiple-choice question answering is not intended for a extensive number of choices, the Transcriptions dataset is evaluated using a reduced version with eleven classes instead of the 29 available. These eleven classes consist of the ten most frequent labels plus an ``Other'' class. The number of samples evaluated is not affected. Additionally, the models' logit space has been constrained to align with the response options of a multiple-choice scenario and, thereby, allow for automated evaluation. The token identifiers associated with the feasible response options are determined and used to filter the logit space.

\subsection{Conditional text generation task}
Conditional text generation is assessed with the MIMIC-CXR dataset, using perplexity as the performance evaluation metric. Perplexity (PPL) is a measure of uncertainty on the value of a sample from a discrete probability distribution. Let $X=(x_0,x_1,..,x_T)$ be a tokenized sequence, then 
$$\textnormal{PPL}(X) = \exp\{-\frac{1}{T}\sum_{t=1}^{T}\log p_\theta(x_t\mid x_{<t})\}$$
where $\log p_\theta(x_t \mid x_{<t})$ is the log-likelihood of the $t$-th token conditioned on the preceding tokens $x_{<t}$.

Decoder-only models are employed for evaluation, with a total of $20$ models considered. Among these models is Galactica, whose tokenizer lacks special tokens. As a consequence, two scenarios are analyzed: the inclusion and the non-inclusion of the start-of-sequence (BOS) token. The BOS token is a special token typically used by generative models to indicate the start of a text. In the first scenario, this token is included during tokenization, and perplexity is calculated from the first token in the texts. When the model's tokenizer does not have the BOS token predefined, such as \texttt{Falcon-7B}, it is then defined as the tokenizer's first special token. In the second scenario, the BOS token is excluded, and perplexity is calculated from the text's second token.

\subsubsection{Dataset} The dataset evaluated in conditional text generation is MIMIC-CXR, introduced below. Details on its preprocessing and characterization are in \Cref{app:data}.

\noindent\textbf{MIMIC-CXR} is an X-ray reports dataset \cite{mimic-cxr-1, mimic-cxr-2, PhysioNet}. Relevant sections of these reports are extracted using the code provided by Johnson \etal \cite{mimicrepo,mimiccode}. Subsequently, null and duplicate samples are removed, with the resulting dataset having $57\,711$ samples. None of these samples exceeds the maximum input size allowed for the proposed models.

%% file: sections/04_results.tex
The main findings are outlined below. For comparability, AUC scores reported in this section correspond to evaluating the eleven class-reduced version of Transcriptions (see \cref{subsec:approaches}). In addition, to ensure the robustness of the results, bootstrapping with $1\,000$ iterations is applied to each experiment. Supplementary results are found in \Cref{app:supplementary}.

\subsection{Text classification analysis}
\begin{table*}[ht]
\centering
\caption{Highest-performing models for text classification per approach and metric. The scores presented correspond to the mean and, in parenthesis, its standard deviation of $1\,000$ bootstrap iterations. Approaches are encoded as follows: CES stands for contextual embedding similarity, NLI for natural language inference, and MCQA for multiple-choice question answering.} 
\label{tab:best}
\resizebox{\textwidth}{!}{
\begin{tabular}{lllccc lcc lc}
\midrule
\multicolumn{1}{c}{\multirow{2}{*}{\textbf{Dataset}}} & \multicolumn{1}{c}{\multirow{2}{*}{\textbf{Metric}}} & \multicolumn{4}{c}{\textbf{CES}} & \multicolumn{3}{c}{\textbf{NLI}} & \multicolumn{2}{c}{\textbf{MCQA}} \\ \cmidrule(lr){3-6} \cmidrule(lr){7-9} \cmidrule(l){10-11}
\multicolumn{1}{c}{} & \multicolumn{1}{c}{} & \multicolumn{1}{c}{\textbf{Model}} & \textbf{Score} & \textbf{Prompt} & \textbf{Pooling} & \multicolumn{1}{c}{\textbf{Model}} & \textbf{Score} & \textbf{Prompt} & \multicolumn{1}{c}{\textbf{Model}} & \multicolumn{1}{c}{\textbf{Score}} \\ \midrule
MS-CXR & Accuracy & \texttt{BioLORD-STAMB2-v1-STS2} & $69.68 ~ (1.70)$ & x & Avg. & \texttt{RoBERTa$_\texttt{LARGE}$-MNLI} & $76.49 ~ (1.59)$ & x & \texttt{T0++} & $\mathbf{81.74 ~ (1.45)}$ \\
 & F1-score & \texttt{BioLORD-STAMB2-v1-STS2} & $69.24 ~ (1.67)$ &  & Avg. & \texttt{RoBERTa$_\texttt{LARGE}$-MNLI} & $78.15 ~ (1.44)$ & x & \texttt{T0++} & $\mathbf{83.86 ~ (1.24)}$ \\
\multirow{2}{*}{} & Precision & \texttt{BioLORD-PMB} & $83.34 ~ (1.11)$ &  & CLS & \texttt{RoBERTa$_\texttt{LARGE}$-MNLI} & $80.72 ~ (1.42)$ & x & \texttt{Alpaca 7B} & $\mathbf{85.83 ~ (0.95)}$ \\
 & Recall & \texttt{BioLORD-STAMB2-v1-STS2} & $72.62 ~ (1.34)$ &  & Avg. & \texttt{RoBERTa$_\texttt{LARGE}$-MNLI} & $82.27 ~ (1.33)$ & x & \texttt{T0++} & $\mathbf{89.22 ~ (0.82)}$ \\ \midrule
Transcriptions & AUC score & \texttt{BioLORD-STAMB2-v1-STS2} & $89.03 ~ (0.31)$ & x & Avg. &  \texttt{BART Large-MNLI} & $80.75 ~ (0.46)$ & x & \texttt{Flan-T5-XXL} & $\mathbf{92.37 ~ (0.26)}$ \\ \midrule
\end{tabular}}
\end{table*}

\begin{figure*}[ht]
    \centering
        \begin{subfigure}{\textwidth}
        \centering
        \includegraphics[width=.98\textwidth]{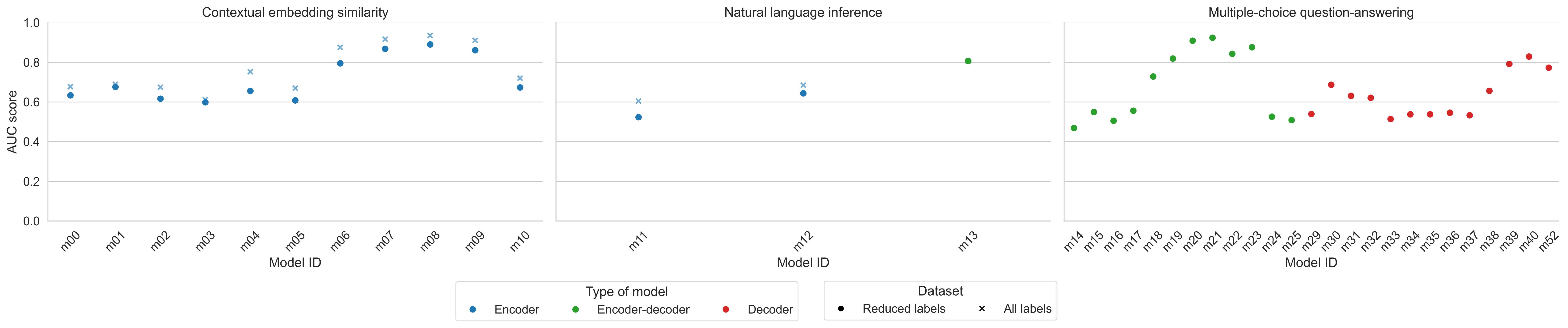}
        \caption{Results for MS-CXR dataset}
    \end{subfigure}
    \\\vspace{5pt}
    \begin{subfigure}{\textwidth}
        \centering
        \includegraphics[width=.98\textwidth]{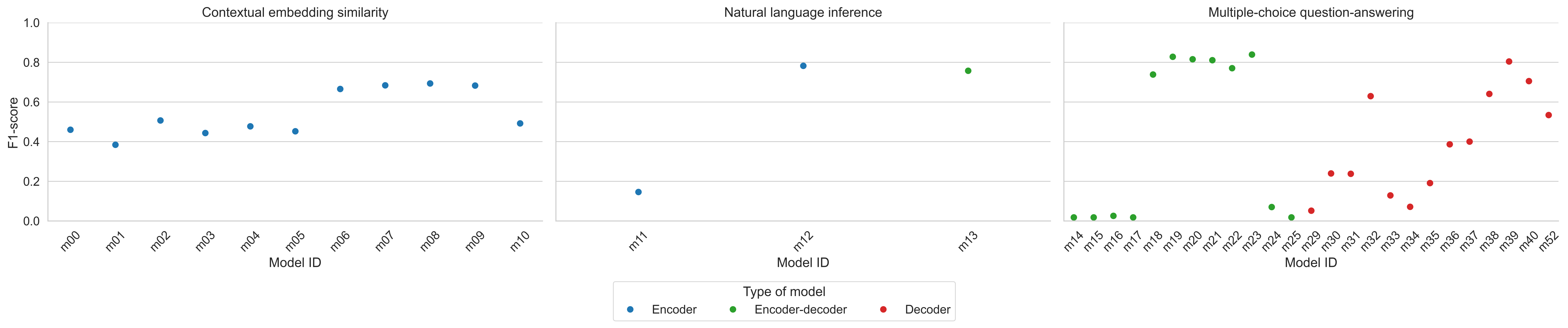}
        \caption{Results for Transcriptions dataset}
    \end{subfigure}
    \caption{Highest model classification scores achieved by approach for the evaluated datasets. Each point corresponds to the mean of $1\,000$ bootstrap iterations. Error bars are calculated as three times the standard deviation of the mean. The highest-performing models are consistent across datasets: BioLORD models (m07-m09) for contextual embedding similarity, MNLI fine-tuned RoBERTa and BART (m12-m13) for NLI, and the largest instruction-tuned models within the T5 family (m20-m23) and instruction-tuned models within the LLaMA family (m39-m40, m52) for multiple-choice QA. Overall, the larger instruction-tuned T5 models emerge as the top performers. The correspondence between the model and ID is found in \Cref{tab:models}.}  
    \label{fig:r4}
\end{figure*}

The highest F1 and AUC scores are achieved with the largest instruction-tuned T5 models, i.e., Flan-T5 (m19-m21) and T0 (m22-m23), as shown in \cref{tab:best} and \cref{fig:r4}. Some of these scores are above $80\%$ in the F1-score and $90\%$ in the AUC score. These instruction-tuned T5 models in question stand among all models considered, ranking within the top 10 highest-performing models in both datasets. Nevertheless, the optimal choice of models may vary when considering precision as the target metric, where \texttt{Alpaca} (m39) and \texttt{LLaMA 2-CHAT-7B} (m40) demonstrate high competence.

Conversely, the lowest F1 and AUC scores are paradoxically obtained with the base T5 models (m14-m17), as evidenced in \cref{fig:r4}. These models, along with their clinically fine-tuned versions (m24-m25), rank within the top 10 lowest-performing models on both datasets. Moreover, the $100\%$ (1/1) and $75\%$ (6/8) of the models underperforming a random evaluator in Transcriptions and MS-CXR datasets, respectively, belong to base and clinically fine-tuned T5 models. 

Delving into each approach, BioLORD models (m07-m09) are consistently the best choice for the contextual embedding similarity approach. MS-CXR dataset is relatively more complex than the Transcriptions dataset for these models, as reflected by their ranking in performance: 11th versus 3rd place, respectively. \texttt{BART Large-MNLI} (m13) represents the best overall model for the NLI approach. For both datasets, \texttt{BART Large-MNLI} is included in the top 10 highest-performing models, while \texttt{RoBERTa$_\texttt{LARGE}$-MNLI} (m12) only does so for MS-CXR dataset. For the multiple choice QA approach, instruction-tuned models stand out, which include instruction-tuned LLaMa models (m39-m40) to the aforementioned instruction-tuned T5 models. Notably, \texttt{LLaMA 2-CHAT-7B} (m40) is within the top 10 highest-performing models in both datasets. These highest performers per approach consistently show results indicative of clinical knowledge or clinical notions. 

The results of instruction-tuned T5 models support the feasibility of representing discriminative tasks as generative ones by framing them as instructions. These results also underline that generative tasks are not exclusive to decoder-only models, and text-to-text models may be a promising architecture to explore further. For example, versions of T5 tuned to instructions with 3B parameters (m20, m22) provide superior results to decoder-only models, almost three times larger, on both evaluated datasets.

\subsubsection*{Model size -- More parameters alone do not always translate into better results}

\begin{figure*}[ht]
    \centering
    \begin{subfigure}{\textwidth}
        \centering
        \includegraphics[width=.95\textwidth]{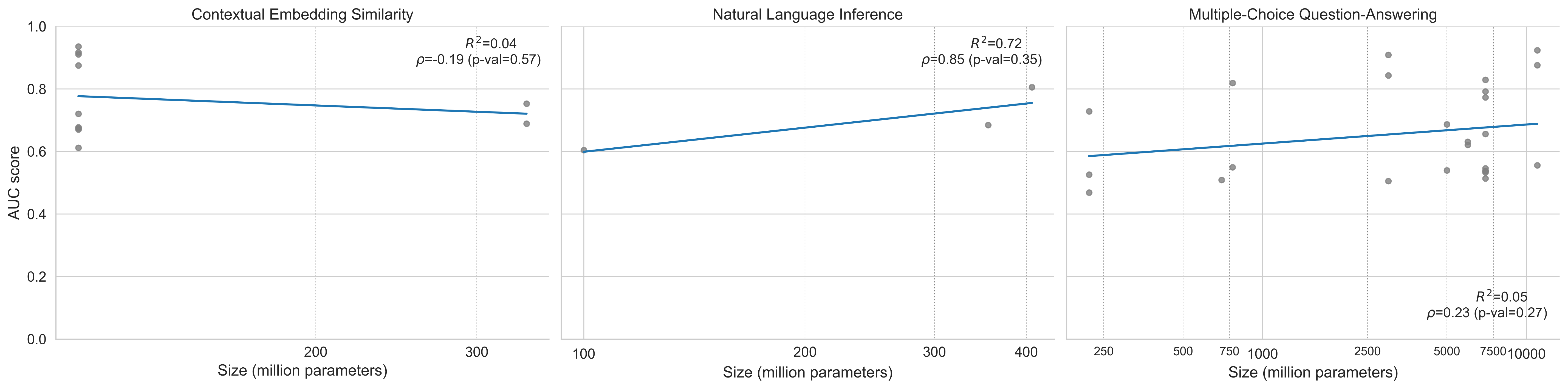}
        \caption{Results for MS-CXR dataset}
    \end{subfigure}
    \\\vspace{5pt}
    \begin{subfigure}{\textwidth}
        \centering
        \includegraphics[width=.95\textwidth]{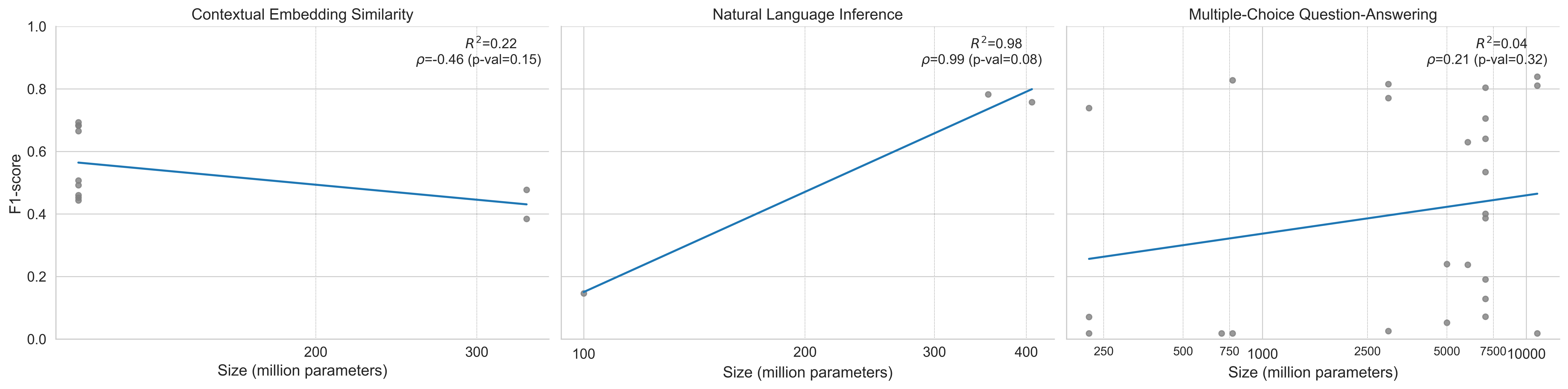}
        \caption{Results for Transcriptions dataset}
    \end{subfigure}
    \caption{Analysis of the impact of the logarithm of size on model performance. Model performance is defined as the highest performance achieved per model over the configurations evaluated. Due to either the lack of size diversity or the low number of samples, Spearman's coefficient, i.e., testing for monotonic relationships, is only reported for the multiple choice QA approach. An analysis of this coefficient suggests that there is not enough evidence to establish the statistical significance of the correlation, as reflected by the p-values.}
    \label{fig:r1}
\end{figure*}

The experiments yield findings questioning the claim that larger models consistently deliver superior performance. The performance of the models as a function of the logarithm of their size is depicted in \cref{fig:r1}. Testing for monotonic relationships via Spearman's correlation is only reported for the multiple-choice QA approach due to the lack of diversity in sizes or number of samples. There is insufficient evidence to conclude that the Spearman's correlation between size and performance is statistically significant in either dataset.

The trend of performance improvement with increasing size is almost nonexistent in the contextual embedding similarity approach. As seen in \Cref{fig:r4}, for instance, models such as \texttt{SapBERT} (m06) and BioLORD models (m07-m09), which excel in this approach, outperform even three times larger models on both datasets. Within the same models, the deltas in performance associated with increasing the number of parameters are inconclusive. \texttt{BERT$_{\texttt{LARGE}}$} (m01) marginally outperforms \texttt{BERT$_{\texttt{BASE}}$} (m00) on the Transcriptions dataset, whereas the opposite is observed in all metrics on the MS-CXR dataset. \texttt{BiomedBERT-large (abstracts only)} (m04) surpasses, albeit marginally, \texttt{BioMedBERT (abstracts only)} (m03) on both evaluated datasets, excluding in precision. Furthermore, performance gains are evidenced when more training data is used, as shown by comparing \texttt{BioMedBERT (abstracts only)} and \texttt{BiomedBERT (abstracts + full text)} (m02).

Similarly, the effect of increasing the size on performance is not sufficiently clear or strong in the multiple-choice question answering approach. While positive Spearman's correlations are obtained, there is insufficient evidence to deem them statistically significant. None of the p-values are $<0.05$, so the null hypothesis that the two variables have no ordinal correlation cannot be rejected. Within T5 models, the effect is minimal or inconsistent when considering their non-instruction-tuned versions (m14-m17, m24-m25). Within the instruction-tuned T5 models (m18-m23), a consistent positive effect of size on performance is observed for both FlanT5 and T0 models on Transcriptions and only for the latter on MS-CXR.

On the other hand, the results in NLI align with the expectation that larger models lead to better performance. However, more models are needed to draw a (solid) conclusion. To have a notion about the lower bound in performance between \texttt{NLI-DeBERTa$_\texttt{base}$} (m11) and the largest models evaluated, the difference between the highest and the lowest values obtained among the evaluated prompts, respectively, is calculated. Considering all metrics, these differences range between $[36.32, 51.59]$ on MS-CXR, having thus that the largest models always lead to performance improvement. Reaching the same conclusion on the Transcriptions dataset is not straightforward, given the results obtained for \texttt{RoBERTa$_\texttt{LARGE}$-MNLI} (m12), as depicted in \Cref{fig:r4}. This model's performance is closer to the performance of \texttt{NLI-DeBERTa$_\texttt{base}$} than to that of \texttt{BART Large-MNLI} (m13).

Altogether, the results do not provide sufficient evidence that only increasing the model size, in number of parameters, leads to an improvement in performance, whether comparing different or the same models. Although model size may be a relevant factor in determining performance, it is hypothesized that training data and objectives are more decisive in small pre-trained language models. This hypothesis aligns with findings in \cite{hoffmann} and \cite{phi3}. Expanding the sample size and diversity could be essential to validate these observations, considering a minimum of 30 or 35 models per approach.

\subsubsection*{Model domain -- More than a specialized domain; model architecture, training data, and training objective}
Current medical datasets remain relatively small compared to those of the general domain, covering only a tiny region of the medical knowledge space \cite{zhou}. Domain specialization of models using only one of these datasets in question may limit their generalization ability \cite{rev_scarcity}.

The effectiveness of domain specialization in improving performance is not evident in the contextual embedding similarity approach, as displayed in \cref{fig:r4}. The domain-specific models considered in this approach are \texttt{Bio+Clinical BERT} (m10), BiomedBERT models (m02-m04), and \texttt{SciBERT} (m05). \texttt{Bio+Clinical BERT} achieves lower scores than expected, positioning around the middle of the performance ranking for this approach. Similarly, some of the BiomedBERT models are outperformed by \texttt{BERT$_{\texttt{BASE}}$} (m00) and \texttt{BERT$_{\texttt{LARGE}}$} (m01), their general domain counterparts. These findings, present in both datasets, challenge the superiority of domain-specific models over general domain ones in the task being evaluated via contextual embedding similarity.

Although existing, evidence supporting the effectiveness of domain specialization is still limited and unclear in the multiple-choice question answering approach. The models to be compared are T5 models (m14-m15) versus their clinical specialized versions (m24-m25), and \texttt{Alpaca} (m39) versus \texttt{MedAlpaca} (m52). Differences between ClinicalT5 and T5 models are $5.75$ and $-4.11$ in AUC scores and $5.24$ and $0.00$ in F1-scores. Similarly, differences between \texttt{MedAlpaca} and \texttt{Alpaca} are $-1.86$ in AUC scores and $26.97$ in F1 scores. Due to these values, it can not be clearly stated that domain specialization positively impacts performance.

Considering the insights discussed and the remarkable performance of BioLORD (m07-m09) models, \texttt{SapBERT} (m06), Flan-T5 (m18-m21) models, and T0 (m22-m23) models in their respective approaches, the training data, training objectives, and model architectures are possibly critical in determining model generalization. Continual pre-training for named entity recognition or medical entity linkage using contrastive learning on UMLS data is likely one of the factors for the success of \texttt{SapBERT} and BioLORD models. Likewise, employing instruction-tuned text-to-text models represents a compelling approach to achieving high performance in multiple-choice QA. Due to the impossibility of concluding on the NLI approach, expanding the analysis to incorporate domain-specialized NLI models in biomedical and clinical domains could be valuable.

\subsubsection*{Prompting and instruction-tuning key to model performance}

\begin{figure*}[ht]
    \centering
    \begin{subfigure}{.5\textwidth}
        \centering
        \includegraphics[width=\textwidth]{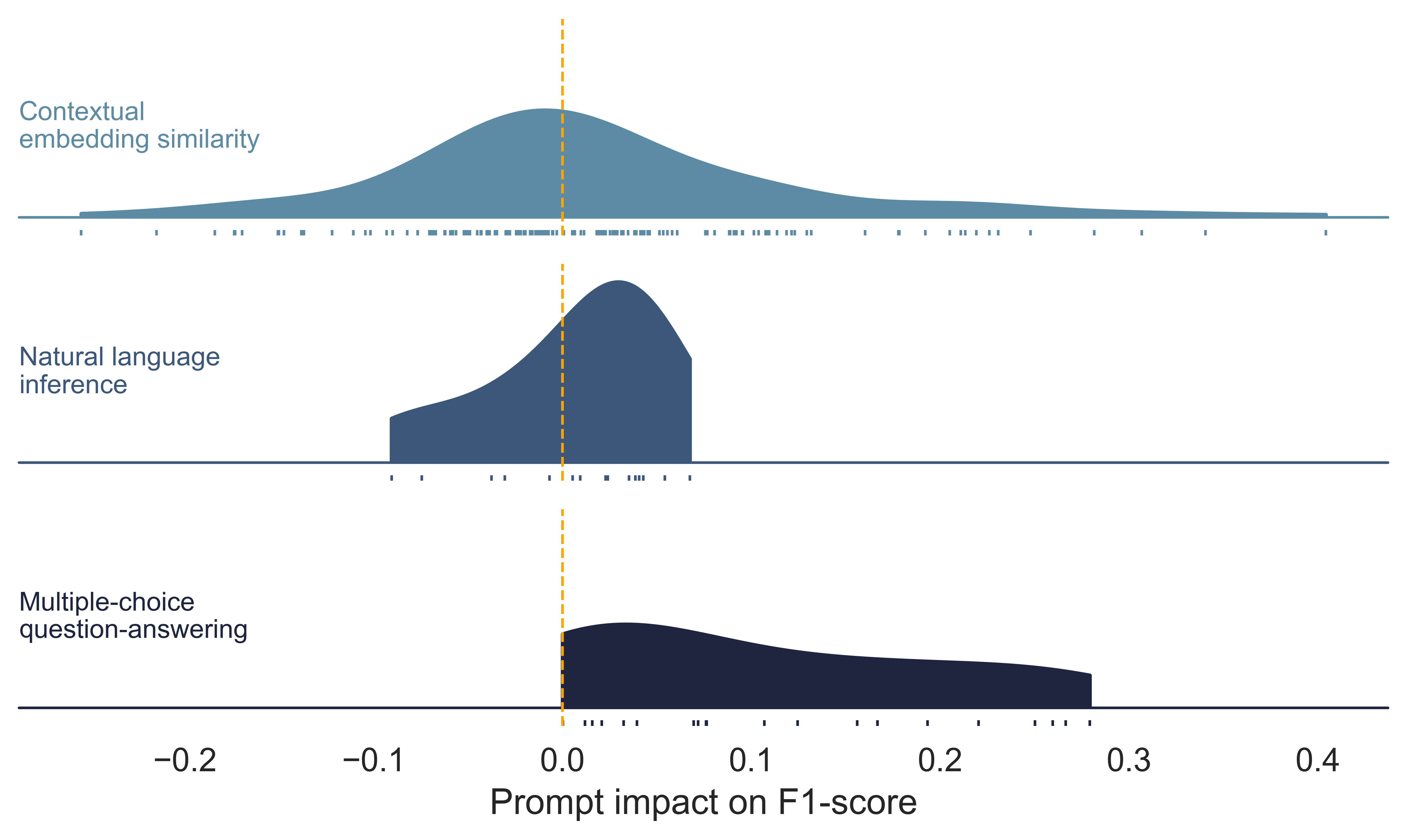}
        \caption{Results for MS-CXR dataset}
    \end{subfigure}%
    \begin{subfigure}{.5\textwidth}
        \centering
        \includegraphics[width=\textwidth]{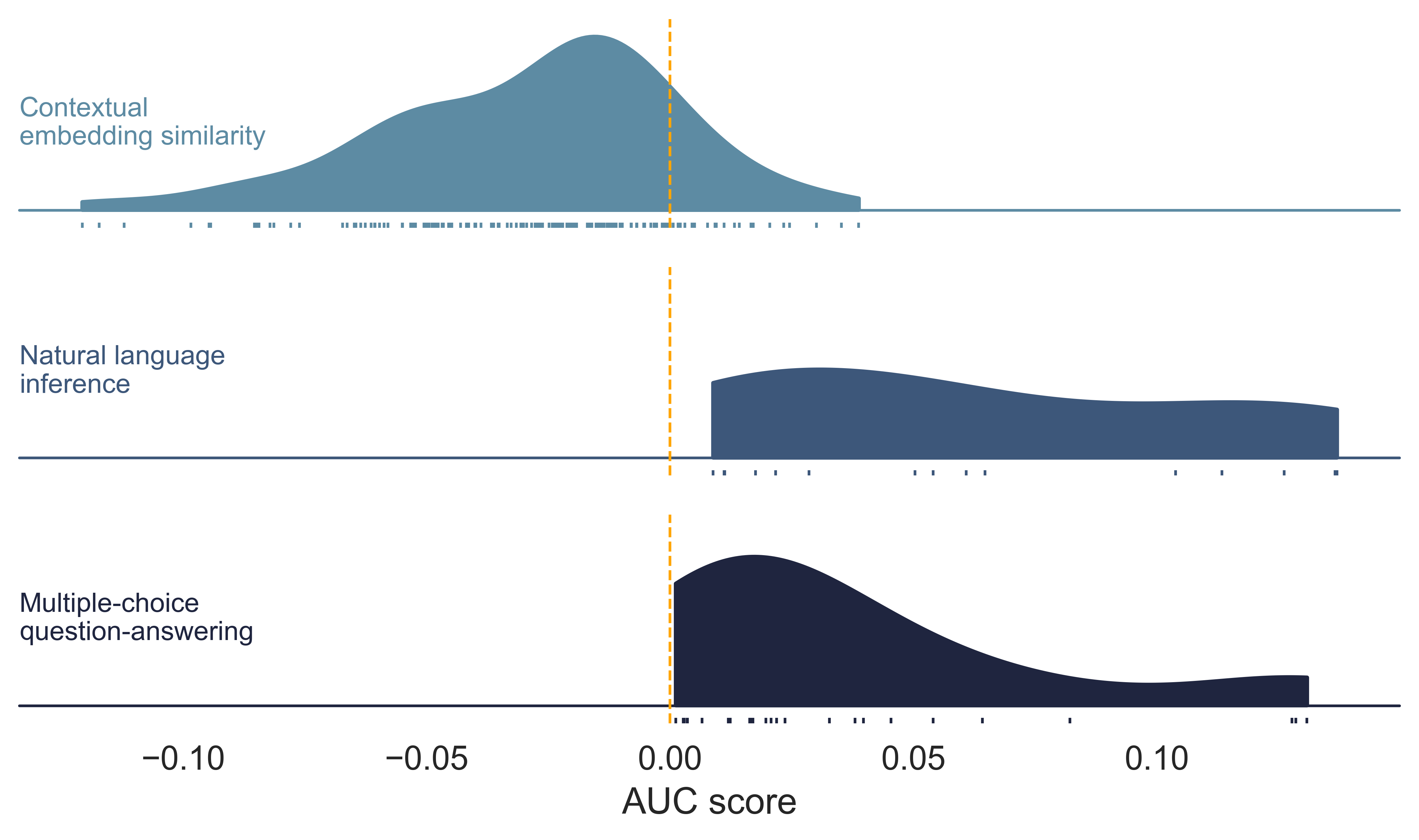}
        \caption{Results for Transcriptions dataset}
    \end{subfigure}
    \caption{Distributions of the impact of prompting on model performance. In contextual embedding similarity and NLI, the impact of prompting is quantified as the difference in performance resulting from prompt usage, with positive values indicating improvement. As the distributions reveal, its usage only sometimes enhances performance. In multiple-choice QA, the impact of prompting is calculated as the variation in performance, expressed in standard deviations, when using different prompts. Optimal scenarios entail non-extreme values, suggesting that there is no strong dependence of performance on prompt wording. The distributions unveil some significant prompt-sensitive models in this case. These distributions are cut to the minimum and maximum observed values to avoid misleading remarks.}
    \label{fig:r3}
\end{figure*}

One of the central points of the study is to analyze the influence of prompting on the models and text classification approaches under investigation. Prompting impact is quantified as the difference in performance resulting from the prompt usage, with positive values indicating an improvement, in contextual embedding similarity and NLI. In multiple-choice QA, this impact is calculated as the variation in performance, expressed in standard deviations, when different instructions are used. The resulting distributions are shown in \cref{fig:r3}.

Using a prompt does not always confer benefits in contextual embedding similarity, as reflected by \cref{fig:r3}. On Transcriptions, the average impact on the AUC score is $-2.25$ points, with values ranging from $-9.32$ to $5.91$. Using any of the proposed prompts improves performance for $45.45\%$ of the model + pooling strategy combinations. In contrast, none of these prompts led to AUC score improvements for \texttt{BioLORD-PMB}, BiomedBERT models, \texttt{BERT$_\texttt{BASE}$}, and \texttt{SciBERT}. On MS-CXR, the impact of the prompt on performance is more positive on average, albeit with more variability. The average impact on the F1-score is $1.30$ points, with values ranging from $-25.43$ to $40.40$. Similar values are reported on accuracy, precision, and recall. Employing any of the proposed prompts represents benefits for the $69.70\%$ to $84.85\%$ of the model + pooling strategy combinations, depending on the metric. The performance of \texttt{BioMedBERT (abstracts only)} and \texttt{BiomedBERT-large (abstracts only)} is enhanced with any of the prompts, whereas the performance of the BioLORD models and \texttt{Bio+Clinical BERT} is hindered.

More consistent benefits are observed than in contextual embedding similarity when examining the prompt impact in the NLI approach. On Transcriptions, any of the proposed prompts yields performance improvements, profiting larger models the most from its usage. The average impact on the AUC score is $8.03$ for \texttt{BART Large-MNLI}, $2.15$ for \texttt{NLI-DeBERTa$_\texttt{base}$}, and $4.87$ for \texttt{RoBERTa$_\texttt{LARGE}$-MNLI}. On MS-CXR, using a prompt only sometimes results in gains, particularly for \texttt{NLI-DeBERTa$_\texttt{base}$}. For this model, the average impact on the F1-score is $-3.42$; while for \texttt{BART Large-MNLI} and \texttt{RoBERTa$_\texttt{LARGE}$-MNLI} is $2.18$ and $2.78$, respectively. Moreover, positive prompt impacts are only observed on precision for \texttt{NLI-DeBERTa$_\texttt{base}$}. In both datasets, there are certain prompts with a high positive impact, whereas others do not, mostly independent of the model.

Similarly, prompt importance is also evident in the multiple-choice question answering approach, given its significant observed influence on model performance. The proportion of models performing better than a random evaluator (AUC score $50\%$) on Transcriptions increases from $52\%$ to $96\%$ with appropriate prompts. Similarly, the proportion of better than a random evaluator (F1-score $12.5\%$) on MS-CXR rises from $25\%$ to $85\%$. Prompting importance is thus highlighted not only by the high performance achieved but also by the brittleness of the models. The latter is reflected by the variability in \Cref{fig:r3}, and further supported by \Cref{fig:transcriptions_cg,fig:ms_cg}  in \Cref{app:supplementary}. Between datasets, the highest sensitivity to the prompt is found when evaluating Transcriptions, such that, with certain prompts, the instruction-tuned models yield similar results to their base counterparts. Overall, no single prompt works universally well for all models.

Regarding instruction-tuning, these models generally outperform their non instruction-tuned counterparts. The instruction-tuned T5 versions, whether T0 or Flan-T5, in any size considered, exhibit superior performance than their base counterparts. Instruction-tuning also improves performance consistently for the LLaMA models, whereas this is not always the case for other generative models: MPT and GPT-J are exceptions on the Transcriptions dataset and Falcon on the MS-CXR dataset. Overall, this tuning technique represents a gain, with an average increase of $21.45$ points in the AUC score and $43.55$ points in the F1-score.

Summarizing, the results endorse the crucial role of the prompt and its wording in the model's performance, with both positive and negative effects presented. Consequently, we advocate using prompts and advanced prompting techniques to guide the model toward better results. This process should also not be limited to a single prompt due to the observed and well-known phenomenon of prompt brittleness \cite{he}. Regarding instruction tuning, this technique proves to be beneficial for the models. More details on the prompt impact can be found in \Cref{fig:ms_cp,fig:transcriptions_cp} in \Cref{app:supplementary}.


\subsection{Conditional text generation analysis}
\begin{figure*}[ht]
\centering
\includegraphics[width=.8\textwidth]{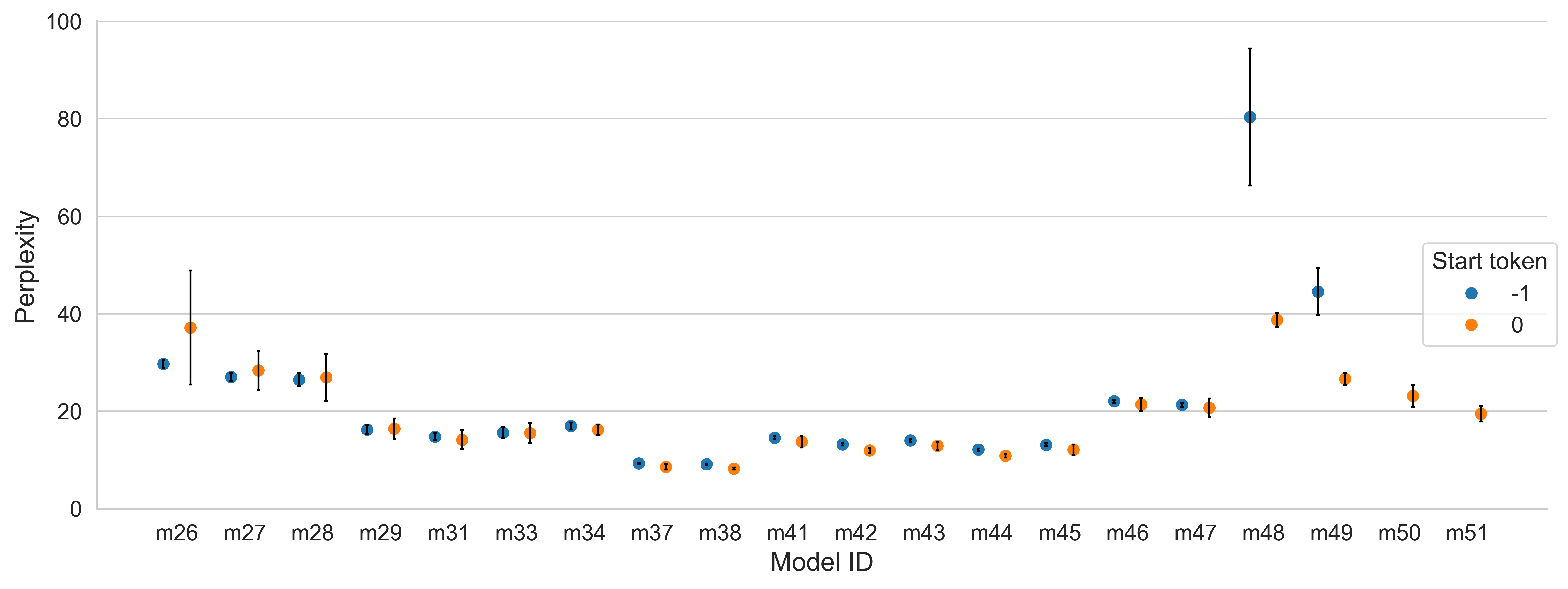}
\caption{Mean perplexity scores for the MIMIC-CXR dataset, disaggregated by BOS token usage. Each point corresponds to the mean of $1\,000$ bootstrap iterations. Error bars are calculated as three times the standard deviation of the mean. The highest-performers are the LlaMA models (m38-m39), whereas the lowest-performers are the BioGPT models (m48-m49). Not using the BOS token is beneficial for $77.78\%$ (14/18) of the models, with the exceptions of the GPT-2 models (m26-m28) and \texttt{Palmyra Base 5B} (m29). The correspondence between model and ID is found in \Cref{tab:models}.}
\label{fig:mimic_red}
\end{figure*}

LLaMA models (m38-m39) stand out as the ones with the highest predictive capacity among the models evaluated. Particularly, \texttt{LLaMA 2-7B}  (m38) is the highest performer, with a mean perplexity of $9.12$ when including the BOS token and $8.21$ when not. LLaMA models are also notable for the low standard deviation of their mean, with approximate values of $0.05$ and $0.13$ depending on the BOS token usage. These standard deviations indicate higher confidence in the estimated value of the mean.

Conversely, BioGPT models (m48-m49) are the models with the most significant difficulty in comprehending the dataset. \texttt{BioGPT} (m48), the lowest performer, presents a mean perplexity of $80.34$ when including the BOS token and $38.70$ when not. The variability on the mean of these models is among the highest observed, with approximate standard deviations of $3.15$ and $0.44$ depending on the BOS token usage. These results are paradoxical considering that \texttt{BioGPT} is domain-specific while \texttt{LLaMA 2-7B} is not.

Similarly to previous findings for text classification, domain specialization does not necessarily imply surpassing general domain models. For the medium-size domain-specific models, it is observed that \texttt{BioGPT} (m48) does not outperform any of the general domain models, while \texttt{GPT-2-PubMed Medium} (m46) does. For the large size models, domain specialization proves beneficial; whereas for the XL and XXL sizes, neither Galactica (m50-m51) nor \texttt{BioGPT-Large} (m49) clearly outperforms general domain models. Consequently, the only specialized models that prove advantageous are GPT-2-PubMed (m46-m47).

On the other hand, increasing the model size contributes to improved performances, regardless of whether or not the BOS token is included. A slight performance improvement is also observed for the second versions (m42, m44) versus the first versions (m41, m43) of OpenLLaMA. This improvement is on average $1.58$ and $1.95$ points on the perplexity for the 3B and 7B parameter versions, respectively. Considering that the difference between these versions of OpenLLaMA is the dataset used for pre-training, the results obtained for conditional text generation do not contradict those for text classification. 

Furthermore, the standard deviations of perplexity reveal the presence of exceptionally challenging samples for the models, that is, outliers, which is visually depicted in \cref{fig:mimic_p} in \Cref{app:supplementary}. Moderate outliers, above quantile $0.75$ by $1.5$ times the IQR, represent between $7\%$ and $11\%$ of the data, with BioGPT models having the highest percentages. Extreme outliers, above quantile $0.75$ by three times the IQR, make up between $4\%$ and $7\%$ of the data, with most models exhibiting percentages around $4\%$ and $5\%$. 

\subsubsection*{Groups of generative models -- LLaMA and GPT-2}
\begin{figure*}[ht]
\centering
\begin{subfigure}{0.45\textwidth}
    \centering
    \includegraphics[width=\textwidth]{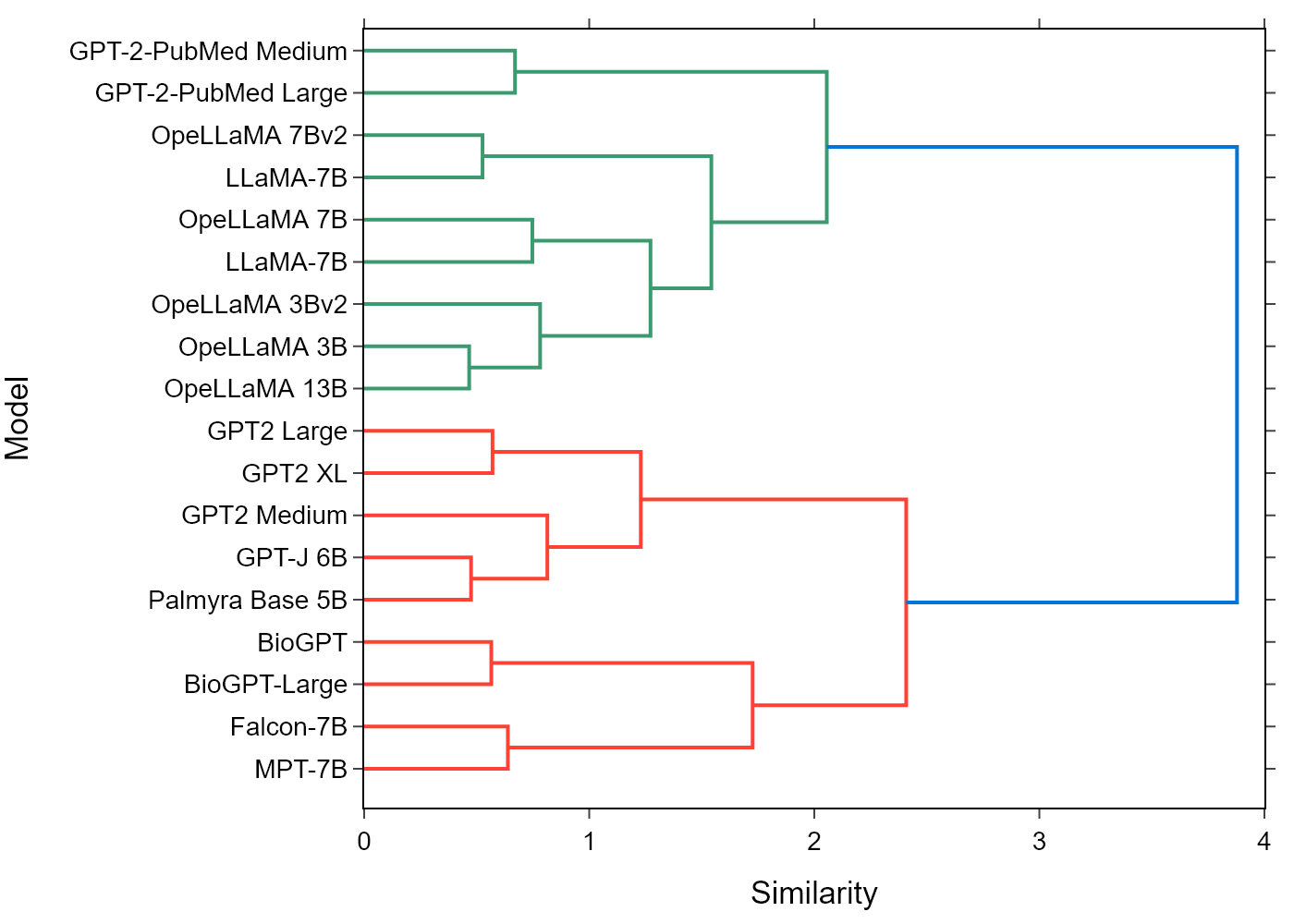}
    \subcaption{Perplexities with BOS token}
    \label{fig:mimic_uw}
\end{subfigure}\hspace{25pt}%
\begin{subfigure}{0.45\textwidth}
    \centering
    \includegraphics[width=\textwidth]{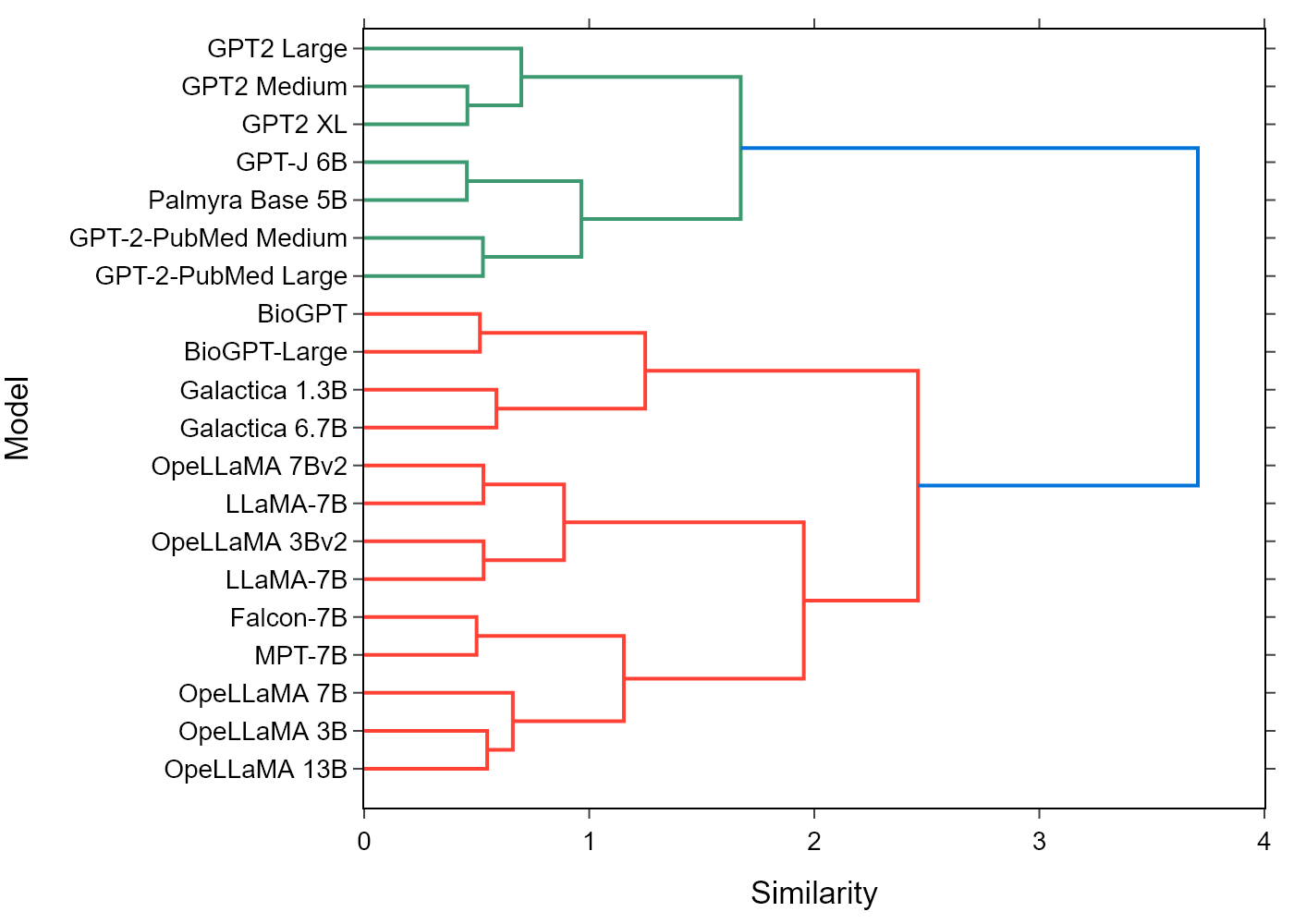}
    \subcaption{Perplexities without BOS token}
    \label{fig:mimic_uwo}
\end{subfigure}
\caption{Dendrograms of the UMAP's principal components after being applied to the perplexities. Two major clusters of models are observed: the GPT-2 models and the LLaMA models.}
\label{fig:mimic_u}
\end{figure*}

Two procedures are carried out to determine whether the models exhibit similar perplexity behavior and identify potential clusters among them. The first procedure involves calculating the correlations between the models. Spearman's and Pearson correlations are considered, assessing monotonic and linear relationships, respectively. The second procedure consists of dimensionality reduction via UMAP, followed by hierarchical clustering, represented by dendrograms in \cref{fig:mimic_u}. Both procedures reveal the existence of two main groups of models: the GPT-2 and the LLaMA models.

In general, all models are positively correlated, indicating that most samples have a similar relative difficulty for these models. BioGPT models (m48-m49) are the only exception to this. Further looking at the Pearson correlations, clustering patterns are present, where groups such as the LLaMA, the OpenLLaMA, and the GPT-2 models are identified. Although these previous clusters are somewhat expected, some unexpected associations are also evident, such as between \texttt{Falcon-7B} and \texttt{MPT-7B} and between \texttt{Palmyra Base 5B} and \texttt{GPT-J 6B}. Moreover, linear relationships between the LLaMA and OpenLLaMA models weaken, interestingly, when the BOS token is used, indicating more pronounced performance disparities.Possibly, training data plays a role, as it is essentially their main difference \cite{openllama}. 

%% file: sections/05_conclusion.tex
This study comprehensively explores small pre-trained language models with varying sizes, architectural families, and domains. These models, being $52$ considered, are tested for two fundamental medical natural language processing tasks: text classification and conditional text generation. The size of the models ranges from $110$ million to $13$ billion parameters, which is relatively small compared to recent language models but suitable for consumer-grade computing resources. Our findings have significant implications, particularly for researchers and organizations operating under computational resource-constrained settings.

For the text classification task, three distinct approaches are explored: context embedding similarity, natural language inference, and multiple-choice question answering. BioLORD and SapBERT models have demonstrated remarkable performance in text classification via contextual embedding similarity. Similarly, the instruction-tuned versions of T5, Flan-T5 and T0, followed by the instruction-tuned versions of LLaMA, have exhibited outstanding results in the multiple-choice question answering approach. Flan-T5 and T0 are remarkably good in both general medical and radiology-specific knowledge assessments. To fully understand NLI models' potential, further exploration of this approach is needed, particularly in specialized domains.

A common thread running through our findings is the significance of the prompt in improving text classification performance across different datasets and approaches. This significance extends beyond performance gains; they present a viable alternative to the resource-intensive processes of training and fine-tuning language models, which are often associated with substantial financial and environmental costs. Effective prompt engineering is also essential to mitigate prompt brittleness, ensuring more robust and reliable outcomes. As prompt brittleness is evidenced during the study, and given its importance, further exploration in this line of research is recommended.

Medical datasets often remain relatively small and cover only a small region of the medical knowledge space \cite{zhou}, so domain-specific models specialized using these datasets might see their generalization ability hindered. This practice could explain, to some extent, the results obtained. The results also suggest that the architecture, training data, and training objectives are crucial in determining the model's generalization abilities, possibly outweighing the relevance of model size as a single variable.

For the conditional text generation task, LLaMA models stand out due to their low perplexities with minimal variation. Two groups of models are also identified based on the perplexities obtained in MIMIC-CXR: a group consisting of GPT-2 models and another of LLaMA models. Further research is needed to identify and understand the outliers within these results, as they could hold important insights.

In conclusion, this research highlights the critical role of prompts in language model inference and reaffirms the effectiveness of instruction-tuned generative models in addressing downstream tasks. It also underscores the relevance of model architecture, training data, and training objectives, potentially even more so than model size alone, in its generalization capacity. We advocate for further investigations into topics such as model calibration, i.e., how certain the model is about output, prompt engineering and tuning, and performance concerning issues like hallucinations and biases, among others. Such studies can lead to more effective and ethical applications of language models in healthcare. Extensions to include quantized models and more medical NLP tasks will be considered in further research. Quantification is an interesting and promising approach to making LLMs viable in consumer-grade computing resources.

%% file: sections/90_appendix_data.tex
This section describes the data employed and outlines the corresponding preprocessing procedure.

\subsection{Transcriptions}
Transcriptions is a multi-label dataset with 40 different labels and $2,358$ data samples. The data were extracted from \href{https://www.kaggle.com/datasets/tboyle10/medicaltranscriptions}{Kaggle}, and additional information about the labels can be found in \href{https://mtsamples.com/}{MTSamples.com}.

\subsubsection{Preprocessing}
The preprocessing procedure involves the removal of samples that lack associated reports, adjusting the formatting of the report, and selecting and renaming labels. Formatting adjustments are necessary because line breaks are encoded as comma patterns. To ascertain the final format, we considered the original data source \href{https://mtsamples.com/}{MTSamples.com} and the results generated by ChatGPT as a guide to knowledge of language models.

In terms of labels, less relevant categories were excluded due to their broad level of generality or lack of association with a specific medical specialty. Precisely, the eliminated labels are: ``Consult - History and Phy.'', ``Discharge Summary'', ``Emergency Room Reports'', ``General Medicine'', ``Hospice - Palliative Care'', ``IME-QME-Work Comp etc.'', ``Letters'', ``Office Notes'', ``Pain Management'', ``SOAP / Chart / Progress Notes''. Additionally, several labels contained the ``/'' character, indicating ``or'', which we explicitly replaced with the latter. For example, ``Allergy / Immunology'' was transformed into ``Allergy or Immunology''. Subsequently, the labels ``Chiropractic'' and ``Physical Medicine - Rehab'' were merged into a unified category called ``Physical Medicine and Rehabilitation, or Chiropractic''. Other modifications include transforming ``ENT - Otolaryngology'' into ``Otolaryngology'', ``Hematology - Oncology'' into ``Hematology or Oncology'', ``Lab Medicine - Pathology'' into ``Laboratory Medicine or Clinical Pathology'', ``Pediatrics - Neonatal'' into ``Pediatrics or Neonatal'', and ``Speech - Language'' into ``Speech and Language''.

Upon completion of the preprocessing, the initial count of 40 different labels is reduced to 29, and the number of samples to consider is $2,074$.

\subsubsection{Description}
\begin{figure}[ht]
    \begin{minipage}{0.48\linewidth}
        \centering
        \includegraphics[width=\textwidth]{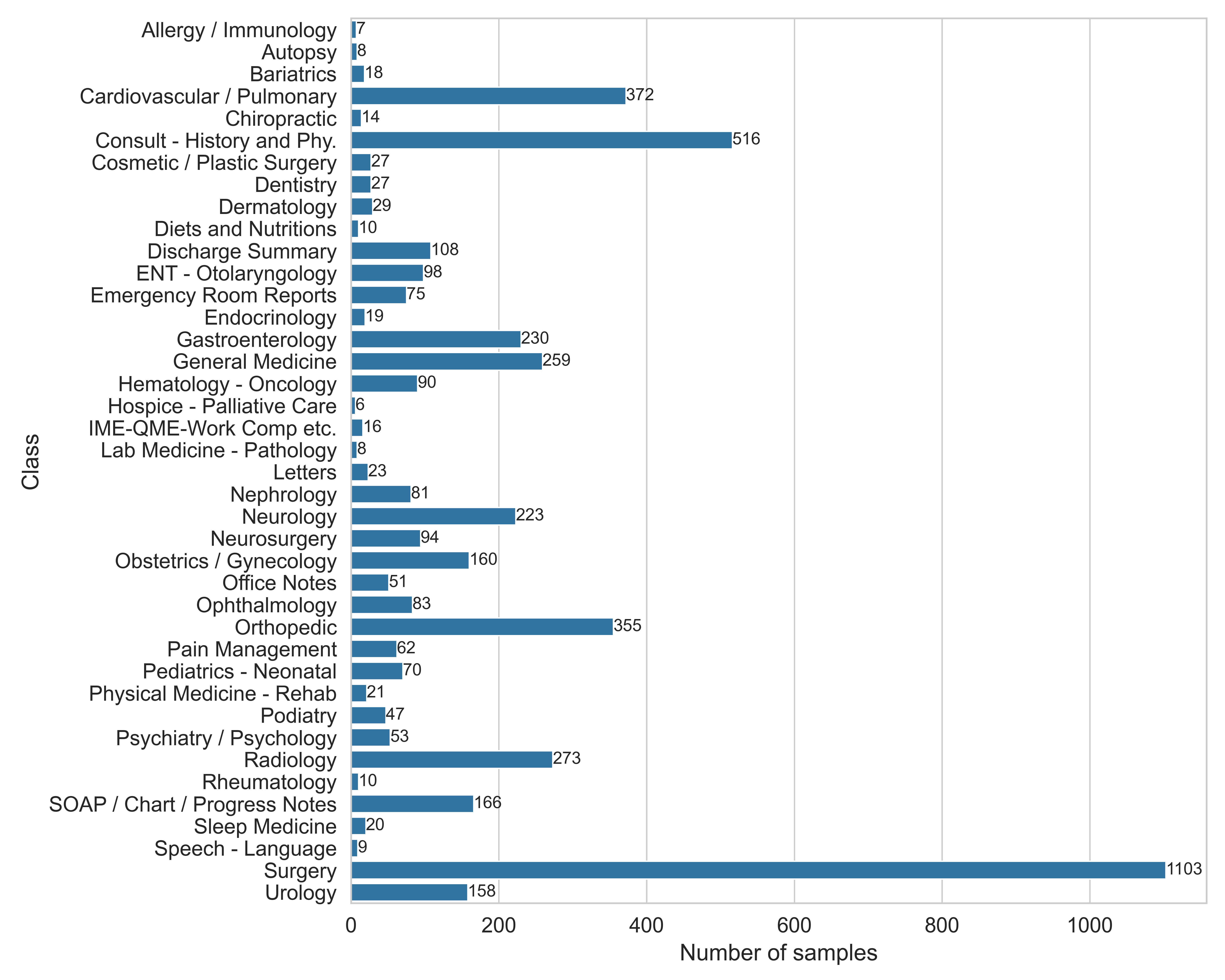}
        \caption{Label distribution of the transcriptions dataset after preprocessing}
        \label{fig:transcriptions_freq}
    \end{minipage}
    \hfill
    \begin{minipage}{0.48\linewidth}
        \centering
        \includegraphics[width=\textwidth]{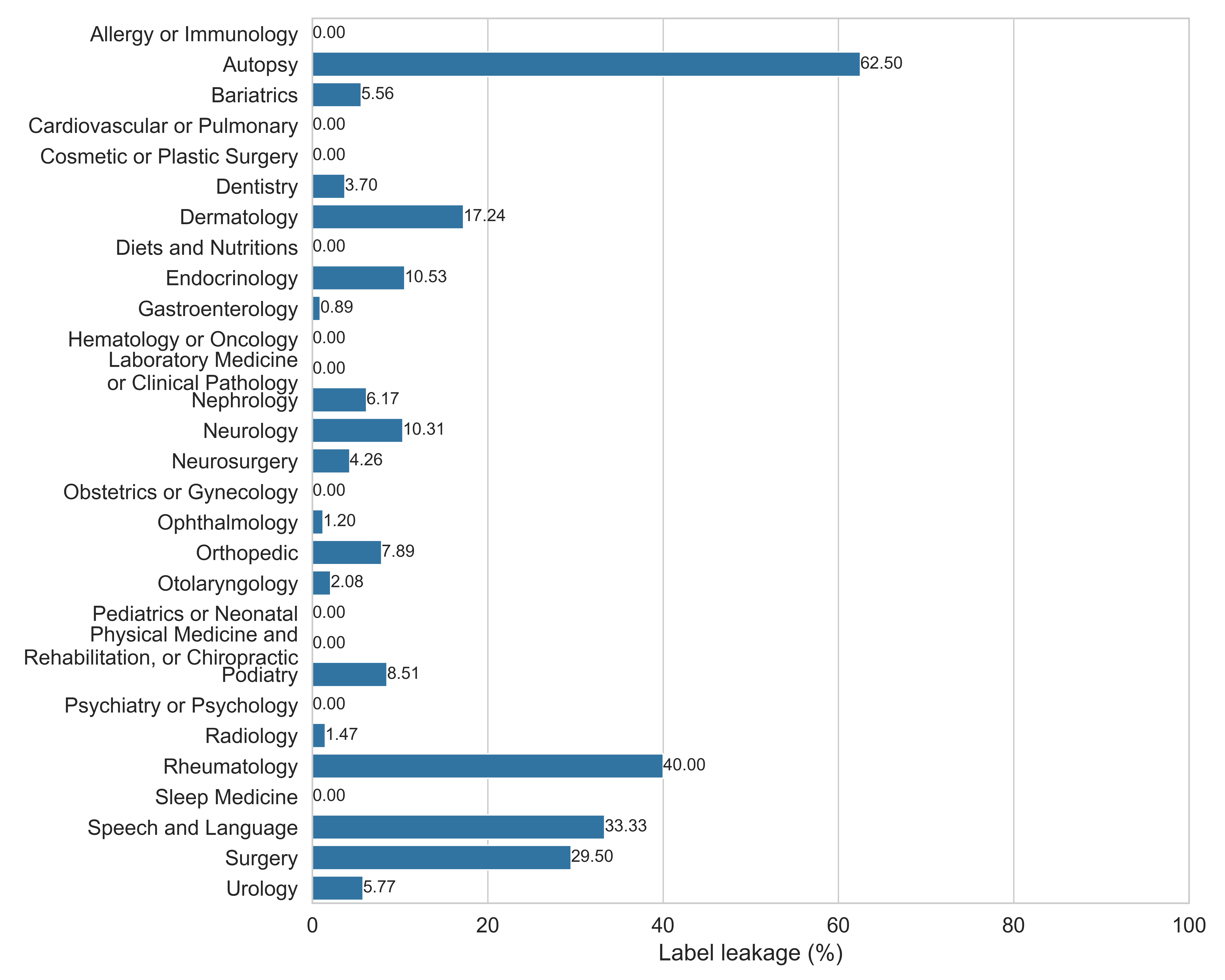}
        \caption{Label leakage for the transcriptions dataset}
        \label{fig:transcriptions_leakage}
    \end{minipage}
\end{figure}

The class distribution is visualized in Fig. \ref{fig:transcriptions_freq}. Surgery is the most prevalent category (in $52.46\%$ of the samples), followed by Cardiovascular or Pulmonary ($17.89\%$) and Orthopedic ($17.11\%$). On the other hand, Allergy or Immunology ($0.33\%$), preceded by Autopsy ($0.38\%$) and Laboratory Medicine or Clinical Pathology ($0.38\%$), are the least frequent categories. The number of labels per sample ranges from 1 to 4, with an average of 2 labels per sample. Additionally, some labels never co-occur within the same sample.

The results of analyzing label leakage, which refers to whether a label appears explicitly in the text to be classified, are shown in Fig. \ref{fig:transcriptions_leakage}. For most labels, label leakage is minimal, except for Autopsy ($62.50\%$), Rheumatology ($40.00\%$), Speech and Language ($33.33\%$), and Surgery ($29.50\%$). Labels without label leakage are Allergy or Immunology, Cardiovascular or Pulmonary, Cosmetic or Plastic Surgery, Diets and Nutritions, Hematology or Oncology, Laboratory Medicine or Clinical Pathology, Obstetrics or Gynecology, Pediatrics or Neonatal, Physical Medicine and Rehabilitation, or Chiropractic, Psychiatry or Psychology, and Sleep Medicine. The presence of labels in texts of other labels is not considered, given that this is a multi-label dataset, and the analysis and interpretation of such occurrences are inherently complex.

\subsection{MS-CXR}
MS-CXR \cite{ms-cxr-1, ms-cxr-2, PhysioNet} is a multi-class dataset with 8 different classes and a corpus of $1,448$ data samples, comprising 718 unique samples. The data can be obtained from \cite{PhysioNet}.

\subsubsection{Preprocessing}
The preprocessing procedure involves removing instances without associated reports and eliminating duplicates. To be precise, 730 samples ($50.41\%$) were identified as duplicates, with a maximum of 82 and an average of 3 duplicates, considering only repeated reports. In addition, when duplicate reports do not agree with the assigned label, either of these labels is evaluated as the true one.

\subsubsection{Description} 
\begin{figure}[ht]
    \begin{minipage}{0.48\linewidth}
        \centering
        \includegraphics[width=\textwidth]{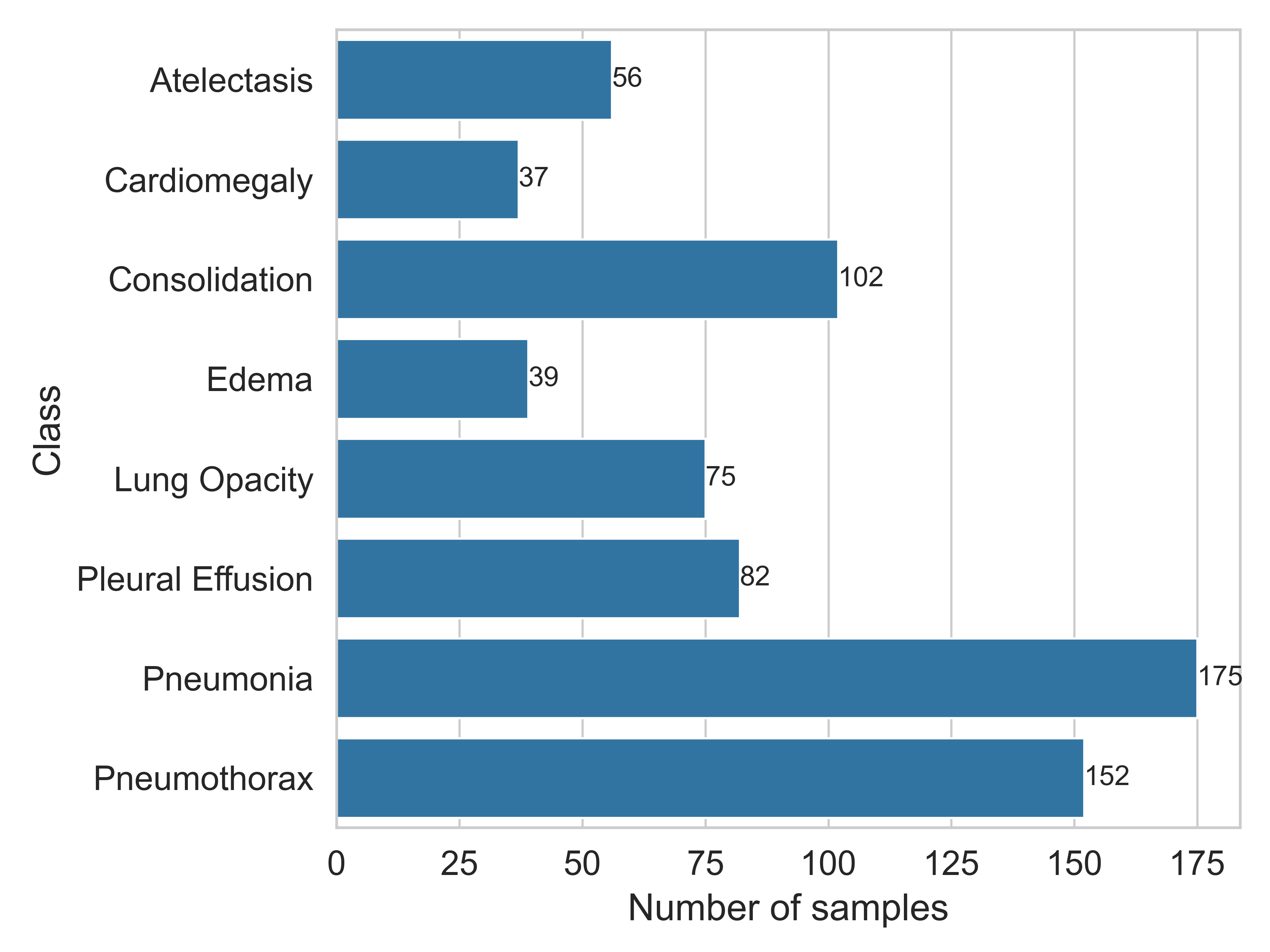}
        \caption{Label distribution of the MS-CXR dataset}\label{fig:ms-cxr_freq}
    \end{minipage}
    \hfill
    \begin{minipage}{0.48\linewidth}
        \centering
        \includegraphics[width=\textwidth]{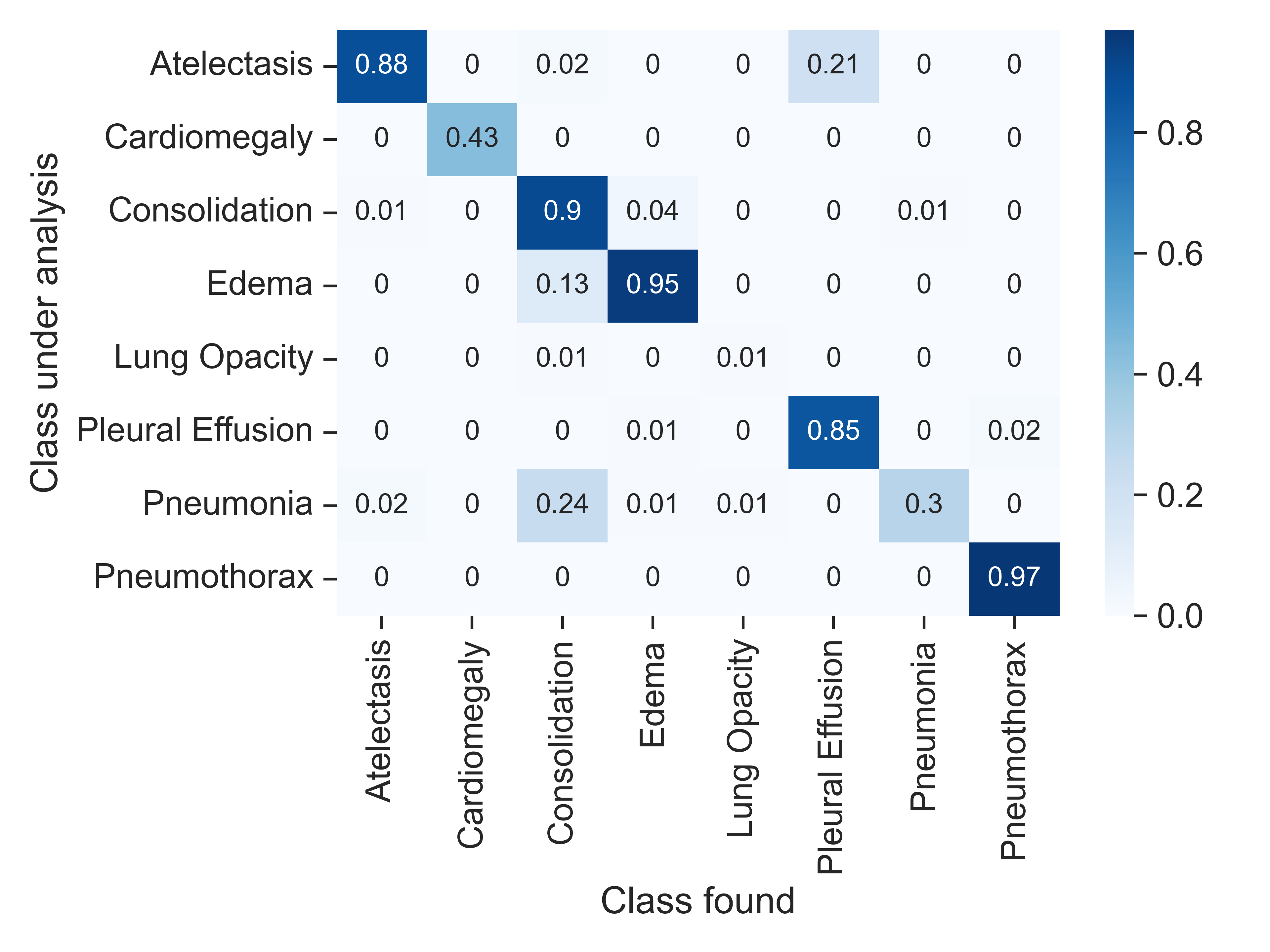}
        \caption{Label leakage and presence of class label on other classes for MS-CXR dataset}
        \label{fig:ms-cxr_leakage}
    \end{minipage}
\end{figure}

The class distribution is depicted in Fig. \ref{fig:ms-cxr_freq}. Overall, the dataset does not exhibit class imbalance. The most frequent classes are Pneumonia ($24.37\%$), closely followed by Pneumothorax ($21.17\%$), while the less frequent classes are Cardiomegaly ($5.15\%$), preceded by Edema ($5.43\%$).

Upon analysis of label leakage, as presented in Fig. \ref{fig:ms-cxr_leakage}, a high label leakage is observed, except for Lung Opacity, which has a low leakage rate of $1.33\%$. In particular, Consolidation, Edema, and Pneumothorax exhibit leakage rates that exceed $90\%$. Classes with leakage rates below $50\%$ include Pneumonia, Cardiomegaly, and Lung Opacity, as mentioned earlier. Regarding the presence of labels in text from other labels, notable occurrences include Consolidation in the classes of Edema ($12.82\%$) and Pneumonia ($24.00\%$), and Pleural Effusion in the Atelectasis class ($21.43\%$).

To conclude, each class's word count per text is measured, and their distributions are presented in Fig. \ref{fig:ms-cxr_freqw}. Classes with shorter texts include Cardiomegaly and Pneumothorax. Although classes with longer texts are not observed, there are flatter distributions with heavy tails, suggesting that the length of texts in these classes is less concentrated around a specific value.

\subsection{MIMIC-CXR}
MIMIC-CXR \cite{mimic-cxr-1, mimic-cxr-2, PhysioNet} is a dataset of radiographic reports that encompasses $78,584$ samples. After extracting the most pertinent sections, $75,029$ samples are identified as informative. This dataset is accessible through \cite{PhysioNet}.

\subsubsection{Preprocessing}
The preprocessing procedure involves extracting the most relevant sections from chest X-ray reports using the codes \cite{mimicrepo} designed for this purpose and publicly available on GitHub \cite{mimiccode}. In addition, texts lacking content and duplicate samples are removed. Texts lacking information are defined as those that are empty or match one of the following: ``.'', ``As above'', ``As above.'', ``As above..'', ``None.'', ``See above.'', ``No changes.'', ``\_\_\_'', ``\_\_\_ earlier'', ``\_\_\_,'', or ``\_\_\_.''. Those mentioned above were identified after meticulously examining texts with a maximum length of two words. In total, these non-informative texts represent merely $0.26\%$ of the dataset. Regarding duplicates, $1.69\%$ of the total samples are duplicated, comprising $23.07\%$ of the dataset. The text with the most duplicates is "No acute cardiopulmonary process." representing $7.88\%$ of the samples. On average, each text appears twice in the dataset.

Upon completion of the preprocessing steps, the dataset results in $57,711 samples$, composed mainly of impressions ($81.92\%$) and findings ($17.48\%$).

\subsubsection{Description}
\begin{figure}[H]
    \begin{minipage}{0.48\linewidth}
        \centering
        \includegraphics[width=\textwidth]{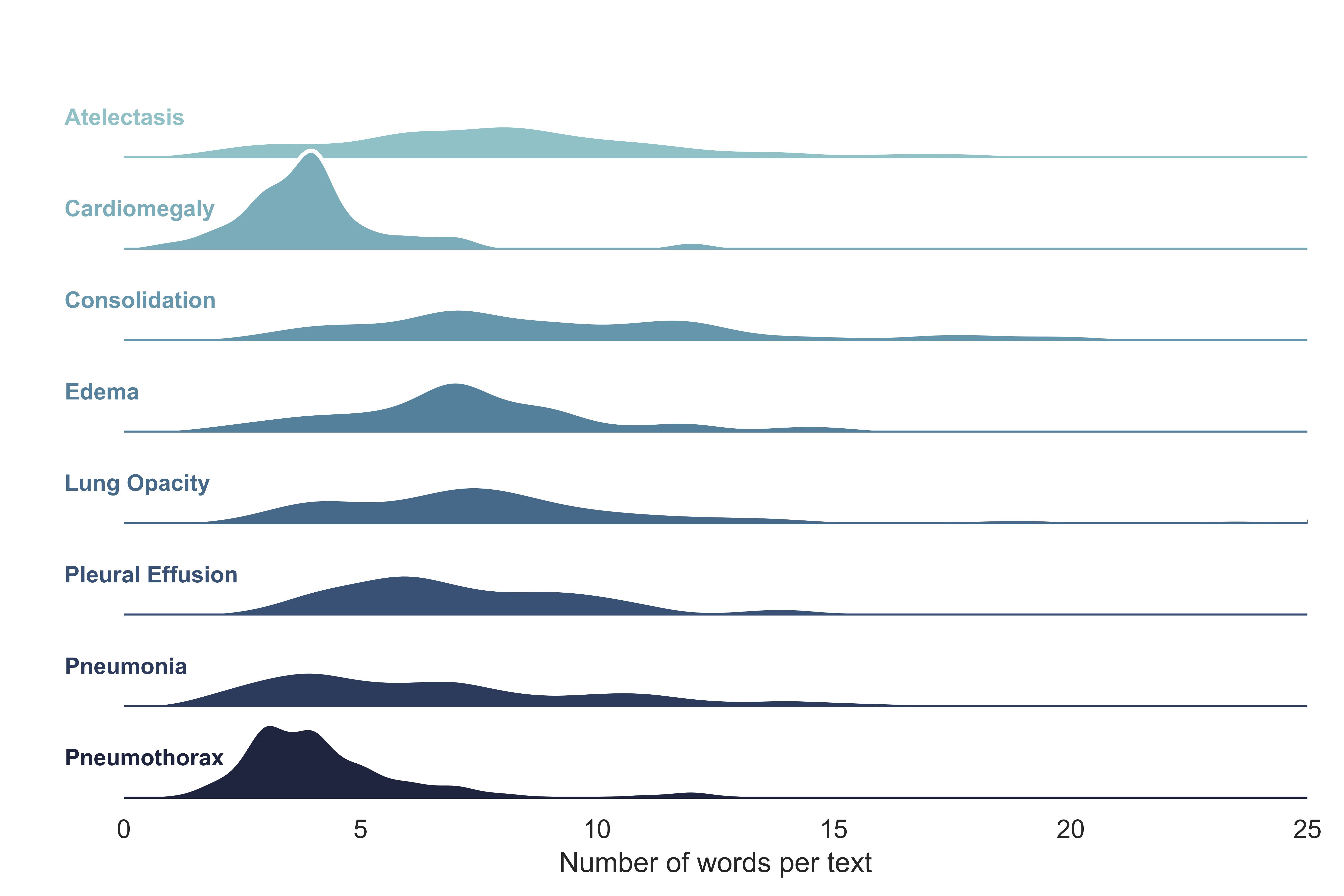}
        \caption{Length distribution of reports, in number of words, for MS-CXR dataset}
        \label{fig:ms-cxr_freqw}
    \end{minipage}
    \hfill
    \begin{minipage}{0.48\linewidth}
        \centering
        \includegraphics[width=\textwidth]{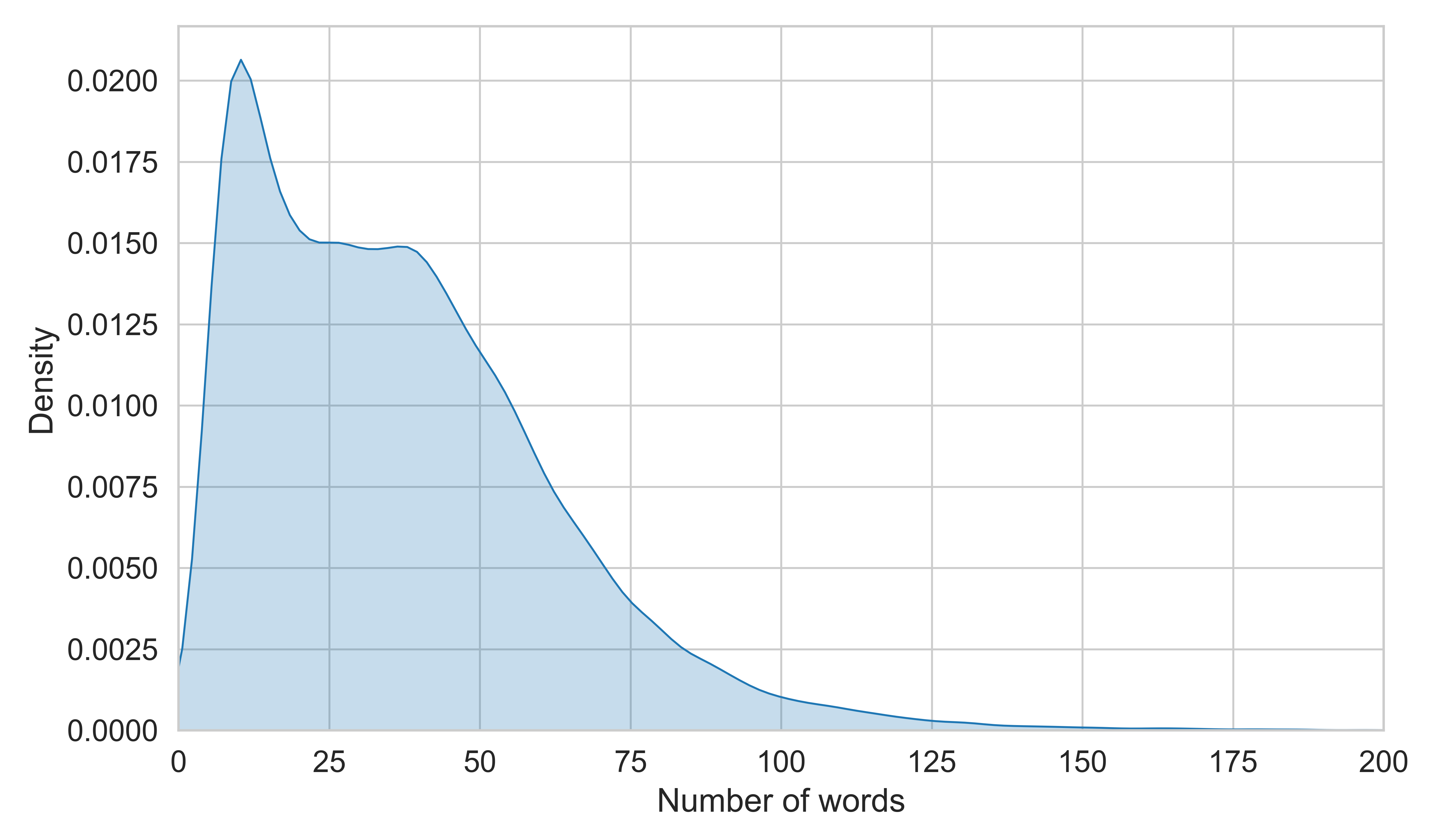}
        \caption{Length distribution of reports in number of words for MIMIC-CXR dataset}
        \label{fig:mimic-cxr}
    \end{minipage}
\end{figure}

Considering the nature of this dataset, its description focuses mainly on the distribution of the number of words per sample, as shown in Fig. \ref{fig:mimic-cxr}. This distribution is left-skewed, with a peak of around 10 words per sample. Moreover, there is a significant plateau between 20 and 40 words per sample. Interestingly, the distribution's right tail extends beyond 150 words per sample. In summary, most texts ($75\%$) contain at most 51 words, with a pronounced peak of around 10 words per sample. However, this dataset also includes longer texts, some reaching up to 307 words.

%% file: sections/91_appendix_models.tex
\begin{table}[H]
\caption{Details on the models studied. The total inference time represents the average time to process the entire dataset per experiment.}
\resizebox{.99\textwidth}{!}{\begin{tabular}{clllcccc}
\midrule
\multirow{2}{*}{\textbf{ID}} & \multicolumn{1}{c}{\multirow{2}{*}{\textbf{Model}}} & \multicolumn{1}{c}{\multirow{2}{*}{\textbf{Type}}} & \multicolumn{1}{c}{\multirow{2}{*}{\textbf{Domain}}} & \multicolumn{1}{c}{\multirow{2}{*}{\textbf{\begin{tabular}[c]{@{}c@{}}Model size\\ (no. parameters)\end{tabular}}}} & \multicolumn{1}{c}{\multirow{2}{*}{\textbf{\begin{tabular}[c]{@{}c@{}}Input max. size\\ (no. tokens)\end{tabular}}}} & \multicolumn{2}{c}{\textbf{Total inference time (seconds)}} \\
 & \multicolumn{1}{c}{} & \multicolumn{1}{c}{} & \multicolumn{1}{c}{} & \multicolumn{1}{c}{} & \multicolumn{1}{c}{} & \multicolumn{1}{c}{\textbf{\begin{tabular}[c]{@{}c@{}}Classification\\ MS-CXR\end{tabular}}} & \multicolumn{1}{c}{\textbf{\begin{tabular}[c]{@{}c@{}}Generation\\ MIMIC-CXR\end{tabular}}} \\ \midrule
m00 & \texttt{BERT$_{\texttt{BASE}}$}  \cite{bert} & Encode-only & General & $110$ M & $512$ & $3.33$ &  \multicolumn{1}{c}{--} \\
m01 & \texttt{BERT$_{\texttt{LARGE}}$}  \cite{bert} & Encode-only & General & $340$ M & $512$ & $5.25$ & \multicolumn{1}{c}{--} \\
m02 & \begin{tabular}[c]{@{}l@{}}\texttt{BiomedBERT}\\ \texttt{(abstracts + full text)} \cite{pubmedbert}\end{tabular}
 & Encode-only & Biomedical & $110$ M & $512$ & $3.10$ &  \multicolumn{1}{c}{--} \\
m03 & \begin{tabular}[c]{@{}l@{}}\texttt{BiomedBERT}\\ \texttt{(abstracts only)} \cite{pubmedbert}\end{tabular} & Encode-only & Biomedical & $110$ M & $512$ & $3.09$ & \multicolumn{1}{c}{--} \\
m04 & \begin{tabular}[c]{@{}l@{}}\texttt{BiomedBERT-large}\\ \texttt{(abstracts only)} \cite{pubmedbert}\end{tabular} & Encode-only & Biomedical & $340$ M & $512$ & $4.56$ & \multicolumn{1}{c}{--} \\
m05 & \texttt{SciBERT}  \cite{scibert} & Encode-only & Biomedical & $110$ M & $512$ &  $3.28$ &  \\
m06 & \texttt{SapBERT}  \cite{sapbert} & Encode-only & Biomedical & $110$ M & $512$ & $3.14$ & \multicolumn{1}{c}{--} \\
m07 & \texttt{BioLORD-STAMB2-v1}  \cite{biolord} & Encode-only & Biomedical & $110$ M & $512$ & $3.35$ & \multicolumn{1}{c}{--} \\
m08 & \texttt{BioLORD-STAMB2-v1-STS2}  \cite{biolord} & Encode-only & Biomedical & $110$ M & $512$ & $3.31$ & \multicolumn{1}{c}{--} \\
m09 & \texttt{BioLORD-PMB}  \cite{biolord} & Encode-only & Biomedical & $110$ M & $512$ & $3.29$ & \multicolumn{1}{c}{--} \\
m10 & \texttt{Bio+Clinical BERT}  \cite{bioclinicalbert} & Encode-only & Clinical & $110$ M & $512$ & $3.15$
 &  \multicolumn{1}{c}{--} \\
m11 & \texttt{NLI-DeBERTa$_\texttt{base}$}  \cite{deberta} & \begin{tabular}[c]{@{}l@{}}Encoder-only\\ (cross-encoder)\end{tabular} & General & $100$ M & $512$ & $8.84$ & \multicolumn{1}{c}{--} \\
m12 & \texttt{RoBERTa$_\texttt{LARGE}$-MNLI}  \cite{roberta} & \begin{tabular}[c]{@{}l@{}}Encoder-only\\ (cross-encoder)\end{tabular} & General & $355$ M & $512$ & $20.38$ & \multicolumn{1}{c}{--} \\ \midrule
m13 & \texttt{BART Large-MNLI}  \cite{bart} & Encoder-decoder & General & $407$ M & $1\,024$ & $23.93$ & \multicolumn{1}{c}{--} \\
m14 & \texttt{T5-V1.1-Base}  \cite{t5, t5v1} & Encoder-decoder & General & $220$ M & $512$ & $6.16$ & \multicolumn{1}{c}{--} \\
m15 & \texttt{T5-V1.1-Large}  \cite{t5, t5v1} & Encoder-decoder & General & $770$ M & $512$ & $14.52$ & \multicolumn{1}{c}{--} \\
m16 & \texttt{T5-V1.1-3B}  \cite{t5, t5v1} & Encoder-decoder & General & $3.0$ B & $512$ & $38.57$ & \multicolumn{1}{c}{--} \\
m17 & \texttt{T5-V1.1-11B}  \cite{t5, t5v1} & Encoder-decoder & General & $11.0$ B & $512$ & $64.88$ & \multicolumn{1}{c}{--} \\
m18 & \texttt{Flan-T5-Base}  \cite{flant5} & \begin{tabular}[c]{@{}l@{}}Encoder-decoder \\ (instruction-tuned)\end{tabular} & General & $220$ M & $512$ & $6.74$ & \multicolumn{1}{c}{--} \\
m19 & \texttt{Flan-T5-Large}  \cite{flant5} & \begin{tabular}[c]{@{}l@{}}Encoder-decoder \\ (instruction-tuned)\end{tabular} & General & $770$ M & $512$ & $16.18$ &  \multicolumn{1}{c}{--} \\
m20 & \texttt{Flan-T5-XL}  \cite{flant5} & \begin{tabular}[c]{@{}l@{}}Encoder-decoder \\ (instruction-tuned)\end{tabular} & General & $3.0$ B & $512$ & $40.71$ & \multicolumn{1}{c}{--} \\
m21 & \texttt{Flan-T5-XLL}  \cite{flant5} & \begin{tabular}[c]{@{}l@{}}Encoder-decoder \\ (instruction-tuned)\end{tabular} & General & $11.0$ B & $512$ & $69.1$ & \multicolumn{1}{c}{--} \\
m22 & \texttt{T0 3B}  \cite{t0} & \begin{tabular}[c]{@{}l@{}}Encoder-decoder \\ (instruction-tuned)\end{tabular} & General & $3.0$ B & $512$ & $38.62$ & \multicolumn{1}{c}{--} \\
m23 & \texttt{T0++}  \cite{t0} & \begin{tabular}[c]{@{}l@{}}Encoder-decoder \\ (instruction-tuned)\end{tabular} & General & $11.0$ B & $512$ & $63.89$ & \multicolumn{1}{c}{--} \\
m24 & \texttt{ClinicalT5-base}  \cite{clinicalt5} & Encoder-decoder & Clinical & $220$ M & $512$ & $5.56$ & \multicolumn{1}{c}{--} \\
m25 & \texttt{ClinicalT5-large}  \cite{clinicalt5} & Encoder-decoder & Clinical & $700$ M & $512$ & $11.94$ & \multicolumn{1}{c}{--} \\ \midrule
\multicolumn{1}{l}{m26} & \texttt{GPT-2 Medium} \cite{gpt2} & Decoder-only & General & $355$ M & $1\,024$ & \multicolumn{1}{c}{--} & $3\,169.67$ \\
\multicolumn{1}{l}{m27} & \texttt{GPT-2 Large}  \cite{gpt2} & Decoder-only & General & $774$ M & $1\,024$ & \multicolumn{1}{c}{--} & $5\,206.18$ \\
\multicolumn{1}{l}{m28} & \texttt{GPT-2 XL}  \cite{gpt2} & Decoder-only & General & $1.5$ B & $1\,024$ & \multicolumn{1}{c}{--} & $5\,330.05$ \\
\multicolumn{1}{l}{m29} & \texttt{Palmyra Base 5B}  \cite{palmyra} & Decoder-only & General & $5.0$ B & $512$ & $94.54$ & $11\,890.56$ \\
\multicolumn{1}{l}{m30} & \texttt{Camel 5B} \cite{camel} & \begin{tabular}[c]{@{}l@{}}Decoder-only\\ (instruction-tuned)\end{tabular} & General & $5.0$ B & $1\,024$ & $96.33$ & \multicolumn{1}{c}{--} \\
\multicolumn{1}{l}{m31} & \texttt{GPT-J 6B}  \cite{gptj} & Decoder-only & General & 6.0 B & 2\,048 & $132.50$ & $16\,495.20$ \\
\multicolumn{1}{l}{m32} & \texttt{Instruct GPT-J}  \cite{igptj} & \begin{tabular}[c]{@{}l@{}}Decoder-only\\ (instruction-tuned)\end{tabular} & General & $6.0$ B & $2\,048$ & $132.58$ & \multicolumn{1}{c}{--} \\
\multicolumn{1}{l}{m33} & \texttt{Falcon-7B}  \cite{falcon} & Decoder-only & General & $7.0$ B & $2\,048$ & $151.15$ & $17\,496.83$ \\
\multicolumn{1}{l}{m34} & \texttt{Falcon-7B-Instruct}  \cite{falcon} & \begin{tabular}[c]{@{}l@{}}Decoder-only\\ (instruction-tuned)\end{tabular} & General & $7.0$ B & $2\,048$ & $151.10$ & \multicolumn{1}{c}{--} \\
\multicolumn{1}{l}{m35} & \texttt{MPT-7B}  \cite{MPT} & Decoder-only & General & $7.0$ B & $2\,048$ & $140.49$ &  $15\,384.00$ \\
\multicolumn{1}{l}{m36} & \texttt{MPT-7B-Instruct}  \cite{MPT} & \begin{tabular}[c]{@{}l@{}}Decoder-only\\ (instruction-tuned)\end{tabular} & General & $7.0$ B & $2\,048$ & $140.56$ &  \multicolumn{1}{c}{--}\\
\multicolumn{1}{l}{m37} & \texttt{LLaMA-7B}  \cite{llama} & Decoder-only & General & $7.0$ B & $2\,048$ & $143.56$ & $19\,203.39$ \\
\multicolumn{1}{l}{m38} & \texttt{LLaMA 2-7B}  \cite{llama2} & Decoder-only & General & $7.0$ B & $2\,048$ & $144.27$ & $19\,225.13$ \\
\multicolumn{1}{l}{m39} & \texttt{Alpaca 7B}  \cite{alpaca} & \begin{tabular}[c]{@{}l@{}}Decoder-only\\ (instruction-tuned)\end{tabular} & General & $7.0$ B & $512$ & $146.31$ & \multicolumn{1}{c}{--} \\
\multicolumn{1}{l}{m40} & \texttt{LLaMA 2-CHAT-7B}  \cite{llama2} & \begin{tabular}[c]{@{}l@{}}Decoder-only\\ (instruction-tuned)\end{tabular} & General & $7.0$ B & $2\,048$ & $144.50$ &  \multicolumn{1}{c}{--}\\
\multicolumn{1}{l}{m41} & \texttt{OpenLLaMA 3B}  \cite{openllama} & Decoder-only & General & 3.0 B & 2\,048 & \multicolumn{1}{c}{--} & $9\,736.33$ \\
\multicolumn{1}{l}{m42} & \texttt{OpenLLaMA 3Bv2}  \cite{openllama} & Decoder-only & General & 3.0 B & 2\,048 & \multicolumn{1}{c}{--} & $9\,914.52$ \\
\multicolumn{1}{l}{m43} & \texttt{OpenLLaMA 7B}  \cite{openllama} & Decoder-only & General & 7.0 B & 2\,048 & \multicolumn{1}{c}{--} & $17\,433.58$ \\
\multicolumn{1}{l}{m44} & \texttt{OpenLLaMA 7Bv2}  \cite{openllama} & Decoder-only & General & 7.0 B & 2\,048 & \multicolumn{1}{c}{--} & $27\,589.57$ \\
\multicolumn{1}{l}{m45} & \texttt{OpenLLaMA 13B}  \cite{openllama} & Decoder-only & General & 13.0 B & 2\,048 & \multicolumn{1}{c}{--} & $7\,125.28$ \\
\multicolumn{1}{l}{m46} & \texttt{GPT-2-PubMed Medium}  \cite{gpt2p} & Decoder-only & Biomedical & 355 M & 1\,024 & \multicolumn{1}{c}{--} & $2\,023.37$ \\
\multicolumn{1}{l}{m47} & \texttt{GPT-2-PubMed Large}  \cite{gpt2p} & Decoder-only & Biomedical & 774 M & 1\,024 & \multicolumn{1}{c}{--} &  $3\,213.49$\\
\multicolumn{1}{l}{m48} & \texttt{BioGPT}  \cite{biogpt} & Decoder-only & Biomedical & 347 M & 1\,024 & \multicolumn{1}{c}{--} &  $1\,680.22$\\
\multicolumn{1}{l}{m49} & \texttt{BioGPT-Large}  \cite{biogpt} & Decoder-only & Biomedical & 1.5 B & 1\,024 & \multicolumn{1}{c}{--} & $4\,840.45$ \\
\multicolumn{1}{l}{m50} & \texttt{Galactica 1.3B}  \cite{galactica} & Decoder-only & Biomedical & 1.3 B & 2\,048 & \multicolumn{1}{c}{--} & $3\,941.80$ \\
\multicolumn{1}{l}{m51} & \texttt{Galactica 6.7B}  \cite{galactica} & Decoder-only & Biomedical & 6.7 B & 2\,048 & \multicolumn{1}{c}{--} & $15\,118.26$ \\
\multicolumn{1}{l}{m52} & \texttt{MedAlpaca 7b}  \cite{medalpaca} & Decoder-only & Clinical & $7.0$ B & $512$ & $146.88$ & \multicolumn{1}{c}{--}\\ \midrule
\end{tabular}}
\end{table}

%% file: sections/92_appendix_prompts.tex
The prompts used for the text classification task via contextual embedding similarity, natural language inference (NLI), and multiple-choice question answering (QA) are presented.

\subsection{Prompts for text classification via contextual embedding similarity and NLI}
The prompts proposed for text classification using contextual embedding similarity and Natural Language Inference (NLI) are exclusively applied to the label (in the case of NLI, to the hypothesis). Table \ref{tab:prompt1} lists the prompts used. Prompt template ID 0 is the default to generate the hypothesis in the zero-shot text classification using the NLI setting, as documented in \href{https://joeddav.github.io/blog/2020/05/29/ZSL.html}{HuggingFace}.

\begin{table}[ht]
\centering
\caption{Prompt templates to be used as contextual embedding similarity and NLI prompts. The column ``Dataset'' specifies the dataset in which the prompt template is applied.}
\label{tab:prompt1}
\begin{tabular}{cll}
\hline 
\textbf{ID} & \multicolumn{1}{c}{\textbf{Prompt template}} & \multicolumn{1}{c}{\textbf{Dataset}} \\\hline
0 & This example is \{label\}. & Transcriptions, MS-CXR \\
1 & This is an example of \{label\}. & Transcriptions, MS-CXR \\
2 & This report belongs to the category \{label\}. & Transcriptions \\
3 & This report belongs to the medical speciality \{label\}. & Transcriptions \\
4 & This report belongs to the medical speciality: \{label\}. & Transcriptions \\
5 & The diagnosis is \{label\}. & MS-CXR \\
6 & There is evidence of \{label\}. & MS-CXR \\
7 & These findings are consistent with \{label\}. & MS-CXR\\\hline
\end{tabular}
\end{table}

\subsection{Prompts for text classification via multiple-choice QA}
The proposed prompts for text classification via multiple-choice question answering are based on the default prompt templates specific to various of the considered instruction-tuned models. These templates are systematically assessed using a set of questions, enabling us to quantify the influence of the question wording. For the MS-CXR dataset, we also incorporate role-based questions. The prompts, their corresponding datasets, and specific requirements are summarized in Table \ref{tab:prompt2} and Table \ref{tab:prompt2q}. 

Each class or label is encoded with an uppercase letter denoting the option, followed by its name. For instance, if the first label is ``$y_1$'', it is represented as ``(A) $y_1$'' within the prompt. In the context of the transcriptions dataset, there are 29 distinct labels. However, due to their large number, we include the top 10 most frequent labels and categorize the remaining labels under an additional ``Other'' option. Specifically, for the transcriptions dataset, we employ templates t01, t02, t03, t04, and t07 along with questions q07, q08, and q09. Whereas for MS-CXR dataset, we employ templates t01, t02, t03, t04, t07, t11, and t13, and questions q03, q04, and q05.

\begin{table}[ht]
\centering
\caption{Questions to be used for the multiple-choice QA templates. The column ``Dataset'' specifies the target dataset.}
\label{tab:prompt2q}
\resizebox{.82\textwidth}{!}{
\begin{tabular}{lll}
\hline 
\multicolumn{1}{c}{\textbf{ID}} & \multicolumn{1}{c}{\textbf{Question}} & \multicolumn{1}{c}{\textbf{Dataset}} \\ \hline
q01 & What is the most plausible diagnosis? & MS-CXR \\
q02 & What is the patient's diagnosis? & MS-CXR\\
q03 & What is the diagnosis? & MS-CXR\\
q04 & Which one of the following is the diagnosis? & MS-CXR \\
q05 & Which one is the patient's diagnosis? & MS-CXR\\
q06 & Which of the options is the most likely to be the diagnosis? & MS-CXR\\
q07 & Which category does the report belong to? & Transcriptions \\
q08 & What is the field that best suits the report? & Transcriptions\\
q09 & Which one is the topic of the report? & Transcriptions\\\hline
\end{tabular}}
\end{table}

\begin{table*}[ht]
\centering
\caption{Prompt structures to be used as multiple-choice QA prompts. Regarding the column ``Requirements'', ``report'' refers to the text sample, ``options'' to the labels provided as choices, and ``question'' to the question itself (see Table \ref{tab:prompt2q}). Note that the term ``question'' sometimes appears capitalized, indicating that the question begins with an uppercase letter when integrated into the template.}
\label{tab:prompt2}
\resizebox*{!}{.92\textheight}{
\begin{tabular}{clll}
 &  &  &  \\[-1.0em]\hline 
 &  &  &  \\[-1.0em]
\textbf{ID} & \multicolumn{1}{c}{\textbf{Prompt structure}} & \multicolumn{1}{c}{\textbf{Requirements}} & \multicolumn{1}{c}{\textbf{Dataset}} \\
 &  &  &  \\[-1.0em] \hline
 &  &  &  \\[-1.0em]
t01 & \makecell[l]{Below is an instruction that describes a task, paired with an input that provides further context. Write a response that \\appropriately completes the request.\\\\\#\#\# Instruction:\\\{question\} Select one of the following options:\\\{options\}\\\\\#\#\# Input:\\\{report\}\\\\\#\#\# Response:\\(} & \makecell[l]{report, options,\\QUESTION} & \makecell[l]{Transcriptions, \\MS-CXR} \\
 &  &  &  \\[-1.0em] \hline
 &  &  &  \\[-1.0em]
t02 & \makecell[l]{Context: \{report\}\\\\Question: \{question\}\\\\Options:\\\{options\}\\\\Answer: (} & \makecell[l]{report, options,\\QUESTION} & \makecell[l]{Transcriptions, \\MS-CXR} \\
 &  &  &  \\[-1.0em] \hline
 &  &  &  \\[-1.0em]
t03 & \makecell[l]{Context: \{report\}\\\\Question: Based on the context, \{question\}\\\\Options:\\\{options\}\\\\Answer: (} & \makecell[l]{report, options,\\question} & \makecell[l]{Transcriptions, \\MS-CXR} \\
 &  &  &  \\[-1.0em] \hline
 &  &  &  \\[-1.0em]
t04 & \makecell[l]{\{report\}. Which one of the following, if true, most strengthens the argument? \{options\}. (} & report, options & \makecell[l]{Transcriptions, \\MS-CXR} \\
 &  &  &  \\[-1.0em] \hline
 &  &  &  \\[-1.0em]
t05 & \makecell[l]{Read the following and answer the question.\\\\\{report\}\\\\\{question\}\\\{options\}\\\\(} & \makecell[l]{report, options,\\QUESTION} & \makecell[l]{Transcriptions, \\MS-CXR} \\
 &  &  &  \\[-1.0em] \hline
 &  &  &  \\[-1.0em]
t06 & \makecell[l]{\{report\}\\\\What's the best answer to this question: \{question\}\\\\\{options\}\\\\(} & \makecell[l]{report, options,\\QUESTION} & \makecell[l]{Transcriptions, \\MS-CXR} \\
 &  &  &  \\[-1.0em] \hline
 &  &  &  \\[-1.0em]
t07 & \makecell[l]{\{report\}\\\\\{question\}\\\\\{options\}\\\\(} & \makecell[l]{report, options,\\QUESTION} & \makecell[l]{Transcriptions, \\MS-CXR} \\
 &  &  &  \\[-1.0em] \hline
 &  &  &  \\[-1.0em]
t08 & \makecell[l]{Read this chest x-ray report: ``\{report\}''\\\\Now answer this question: ``\{question\}''\\\\\{options\}\\\\(} & \makecell[l]{report, options,\\QUESTION} & \makecell[l]{MS-CXR} \\
 &  &  &  \\[-1.0em] \hline
 &  &  &  \\[-1.0em]
t09 & \makecell[l]{Knowing that ``\{report\}'', how would one answer ``\{question\}''\\\\\{options\}\\\\(} & \makecell[l]{report, options,\\QUESTION} & \makecell[l]{Transcriptions, \\MS-CXR} \\
 &  &  &  \\[-1.0em] \hline
 &  &  &  \\[-1.0em]
t10 & \makecell[l]{\{report\}\\Based on the above text, what's the best answer to this question: \{question\}\\\\\{options\}\\\\(} & \makecell[l]{report, options,\\QUESTION} & \makecell[l]{Transcriptions, \\MS-CXR} \\
 &  &  &  \\[-1.0em] \hline
 &  &  &  \\[-1.0em]
t11 & \makecell[l]{You are a doctor and have the following information about a patient from a chest x-ray: \{report\}. Which one of the \\following, if true, most strengthens the argument? \{options\}. (} & report, options & \makecell[l]{MS-CXR} \\
 &  &  &  \\[-1.0em] \hline
 &  &  &  \\[-1.0em]
t12 & \makecell[l]{You are a doctor and have the following information about a patient from a chest x-ray: \{report\}. \{question\} \{options\}. (} & \makecell[l]{report, options,\\QUESTION} & \makecell[l]{MS-CXR} \\
 &  &  &  \\[-1.0em] \hline
 &  &  &  \\[-1.0em]
t13 & \makecell[l]{I want you to act as a virtual doctor. I will describe my symptoms and you will choose the most probable diagnosis among \\the following: \{options\}. You should only reply with the chosen diagnosis, and nothing else. My request is ``\{report\}''. (} & report, options & \makecell[l]{MS-CXR} \\
 &  &  &  \\[-1.0em] \hline
 &  &  &  \\[-1.0em]
t14 & \makecell[l]{I want you to act as a virtual doctor. I will describe my symptoms and you will choose a diagnosis among the possible diag-\\noses. You should only reply with the chosen diagnosis, and nothing else. Do not write explanations. The possible diagnoses \\are: \{options\}. My request is ``\{report\}''. (} & report, options & \makecell[l]{MS-CXR}\\
 &  &  &  \\[-1.0em] \hline
\end{tabular}}
\end{table*}

%% file: sections/93_appendix_supplementary.tex
This appendix presents supplementary figures and tables to support the results presented. They are displayed first by dataset and then by task or approach.  These results do not are not obtained by bootstrapping, but singular values for the complete inference dataset.

\subsection{Text classification task}

\subsubsection{Contextual embedding similarity}
\begin{figure}[ht]
\centering
\begin{subfigure}{\textwidth}
    \centering
    \includegraphics[width=.98\textwidth]{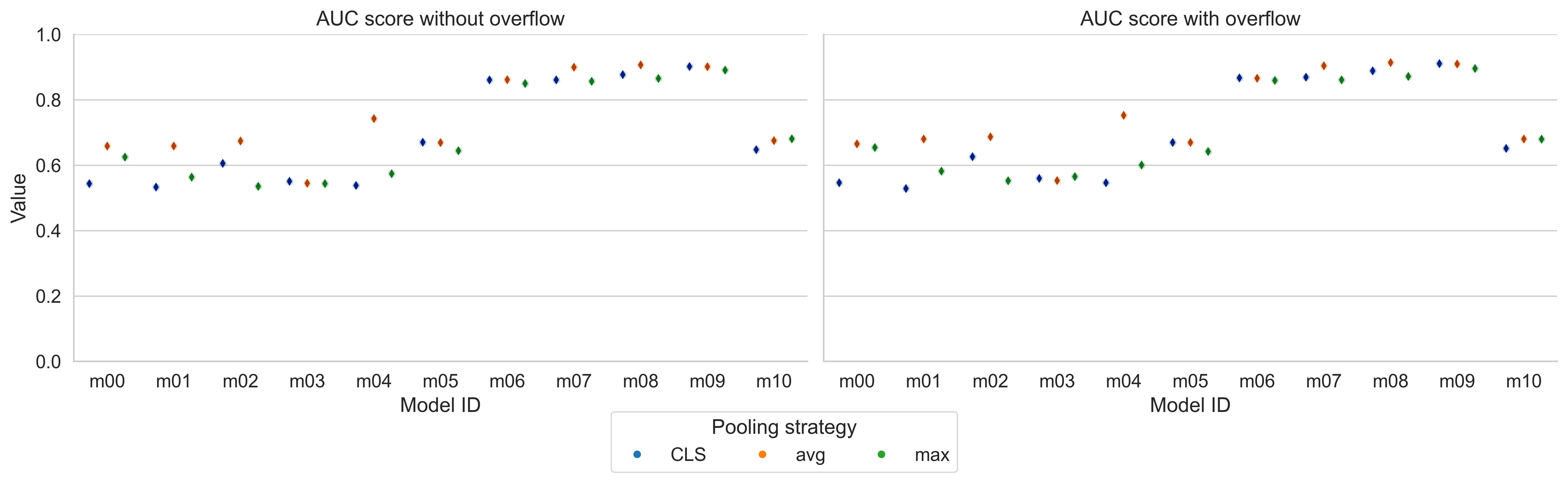}
    \caption{Performance scores without template}
    \label{fig:transcriptions_egwo}
\end{subfigure}
\begin{subfigure}{\textwidth}
    \centering
    \includegraphics[width=.98\textwidth]{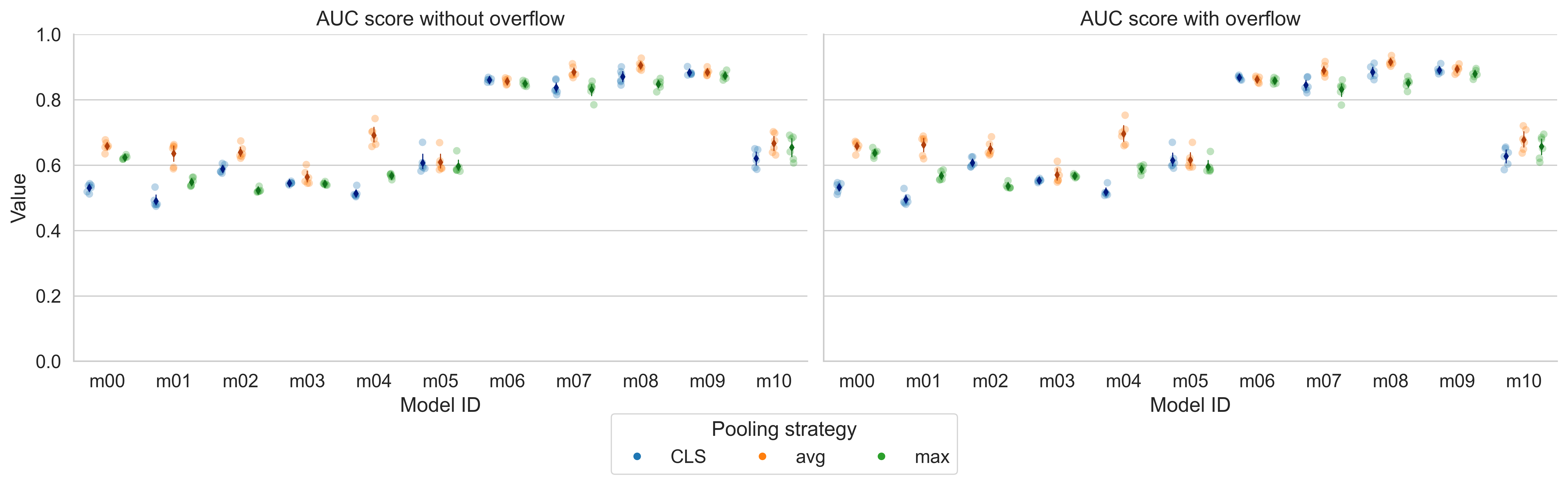}
    \caption{Performance scores with template}
    \label{fig:transcriptions_egw}
\end{subfigure}
\caption{Performance scores for transcriptions dataset using contextual embedding similarity, disaggregated by template and overflow usage, and pooling strategy. Using overflow leads to improvement on $87.88\%$ of the cases, being a case a model + pool strategy. The impact of using overflow is, on average, $0.90$ points of the AUC score. Thus, most of the trends observed when not using overflow are kept. Regarding the pooling strategy, when using overflow, average pooling produces the best results (8 of 11 models), followed by CLS pooling. SapBERT and BioLORD models stand out due to their performance, having more than a 10-point difference in the AUC score with the rest of the models when the best configurations are compared.}
\label{fig:transcriptions_eg}
\end{figure}

\begin{figure}[H]
\centering
\begin{subfigure}{\textwidth}
    \centering
    \includegraphics[width=.98\textwidth]{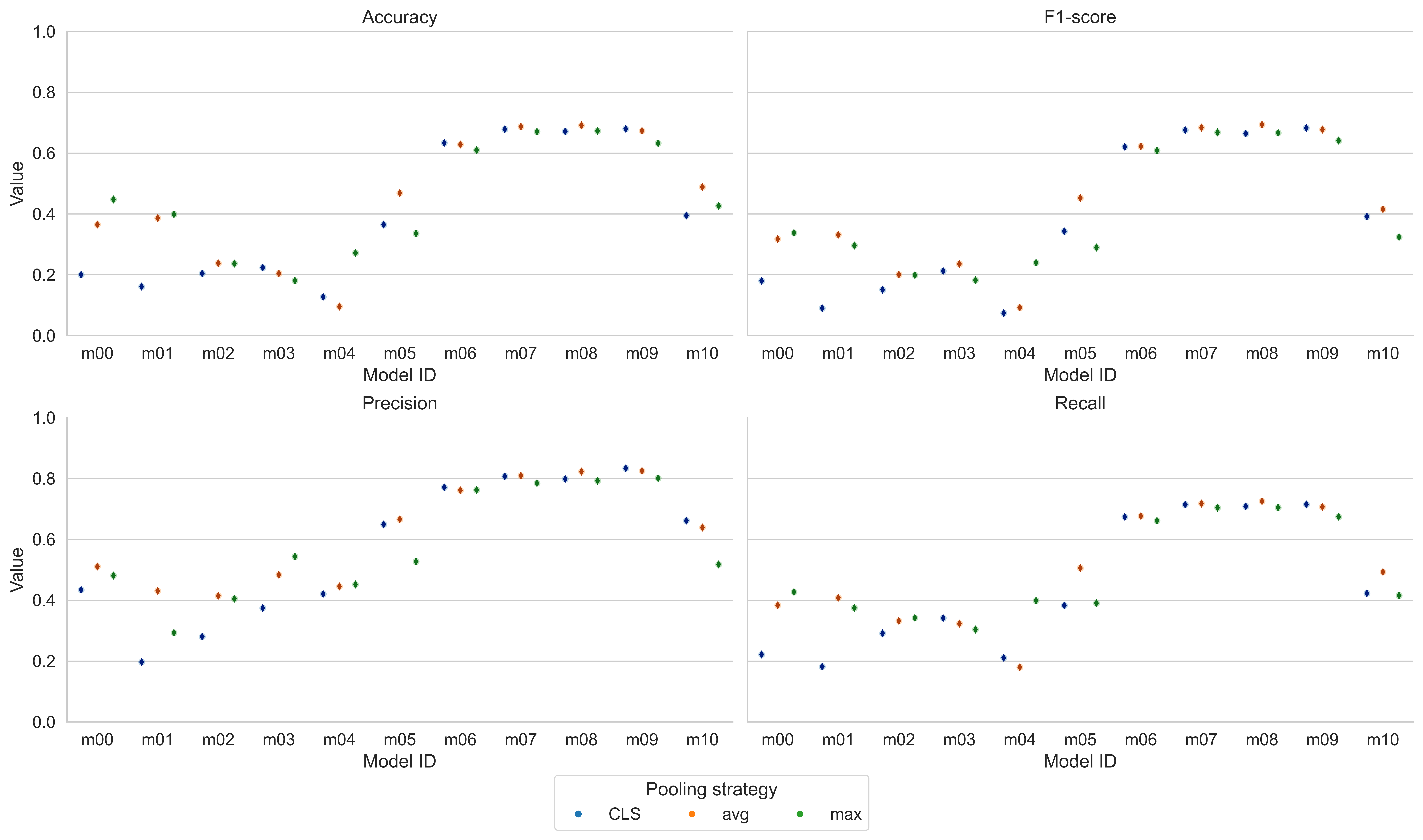}
    \caption{Performance scores without template}
    \label{fig:ms_egwo}
\end{subfigure}
\begin{subfigure}{\textwidth}
    \centering
    \includegraphics[width=.98\textwidth]{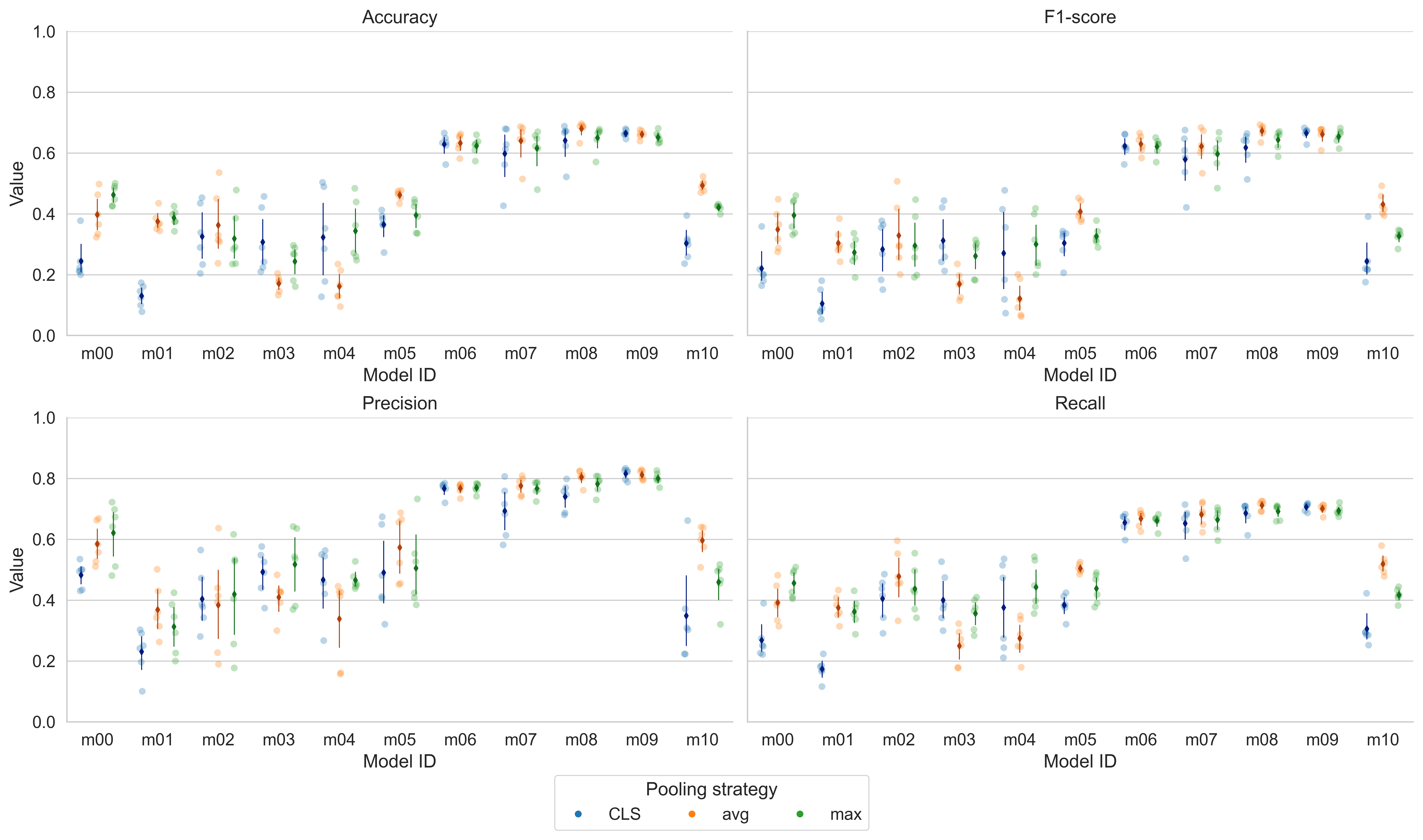}
    \caption{Performance scores with template}
    \label{fig:ms_egw}
\end{subfigure}
\caption{Performance scores for MS-CXR dataset using contextual embedding similarity, disaggregated by template usage and pooling strategy. SapBERT and BioLORD models stand out due to their performance, having at least a difference of 12, 16, 10, and 16 points in accuracy, F1-score, precision, and recall, respectively, with the rest of the models when the best configuration is compared. However, this gap no longer exists when templates are allowed to be used. That means that templates have a primordial role in determining model performance, both to boost or hinder it. Specifically, it boosts the performance on between $69.70\%$ to $84.85\%$ of the cases depending on the performance metric, being a case a model + pool strategy. Regarding the pooling strategy, average pooling produces the best results (5-8 of 11 models, depending on the metric).}
\label{fig:ms_eg}
\end{figure}

Results are depicted in \Cref{fig:r4,fig:ms_eg,fig:transcriptions_eg}. The mapping between the models and their ID is\\
\begin{center}
\begin{tabular}{ll}
m00: \texttt{BERT$_{\texttt{BASE}}$} & $\qquad$ m06: \texttt{SapBERT}\\
m01: \texttt{BERT$_{\texttt{LARGE}}$} & $\qquad$ m07: \texttt{BioLORD-STAMB2-v1}\\
m02: \texttt{BiomedBERT (abstracts + full text)} & $\qquad$ m08: \texttt{BioLORD-STAMB2-v1-STS2}\\
m03: \texttt{BiomedBERT (abstracts only)} & $\qquad$ m09: \texttt{BioLORD-PMB}\\
m04: \texttt{BiomedBERT-large (abstracts only)} & $\qquad$ m10: \texttt{Bio+Clinical BERT}\\
m05: \texttt{SciBERT} & \\
\end{tabular}
\end{center}

\newpage
\subsubsection{Natural language inference} 
\begin{figure}[H]
\centering
\includegraphics[width=.98\textwidth]{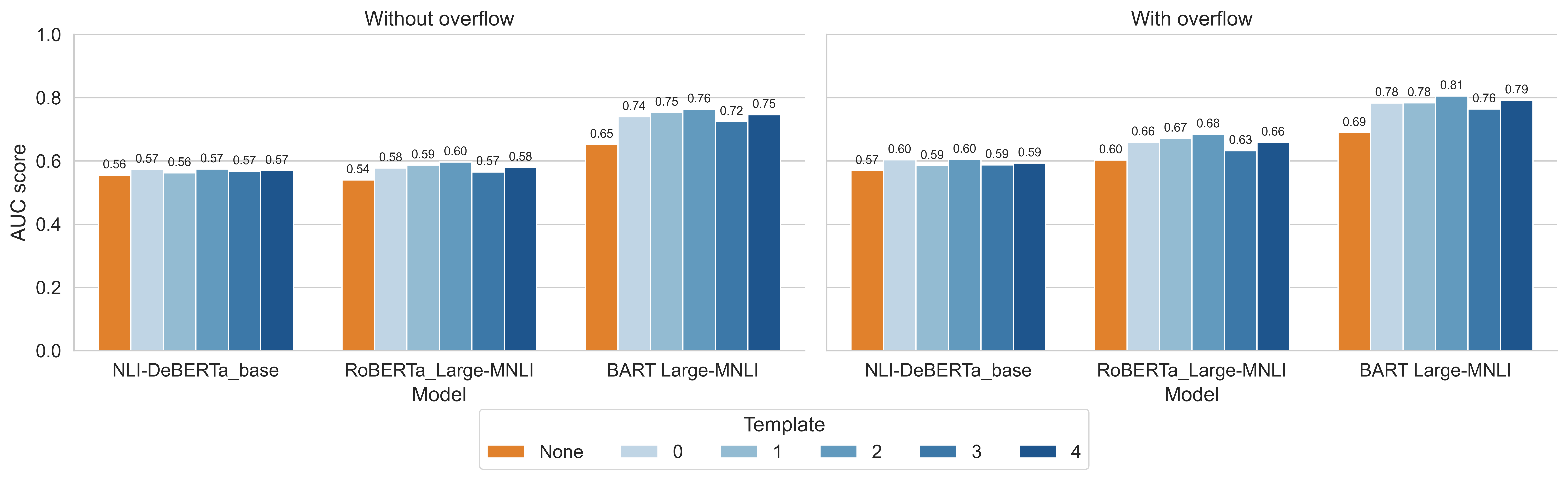}
\caption{Performance scores for transcriptions dataset using natural language inference, disaggregated by template and overflow usage. Using overflow leads to improvement on all three models, having an impact of $2.15$, $4.87$, $8.03$ on \texttt{NLI-DeBERTa$_\texttt{base}$}, \texttt{RoBERTa$_\texttt{LARGE}$-MNLI}, and \texttt{BART Large-MNLI} respectively. Noteworthy to observe is the benefit that using overflow represents for \texttt{RoBERTa$_\texttt{LARGE}$-MNLI}: from performing lower than \texttt{NLI-DeBERTa$_\texttt{base}$} to outperforming it when using overflow. Using a template consistently improves performance, particularly template 2, resulting in AUC scores of up to $80.54$.}
\label{fig:transcriptions_n}
\end{figure}

\begin{figure}[H]
\centering
\includegraphics[width=.98\textwidth]{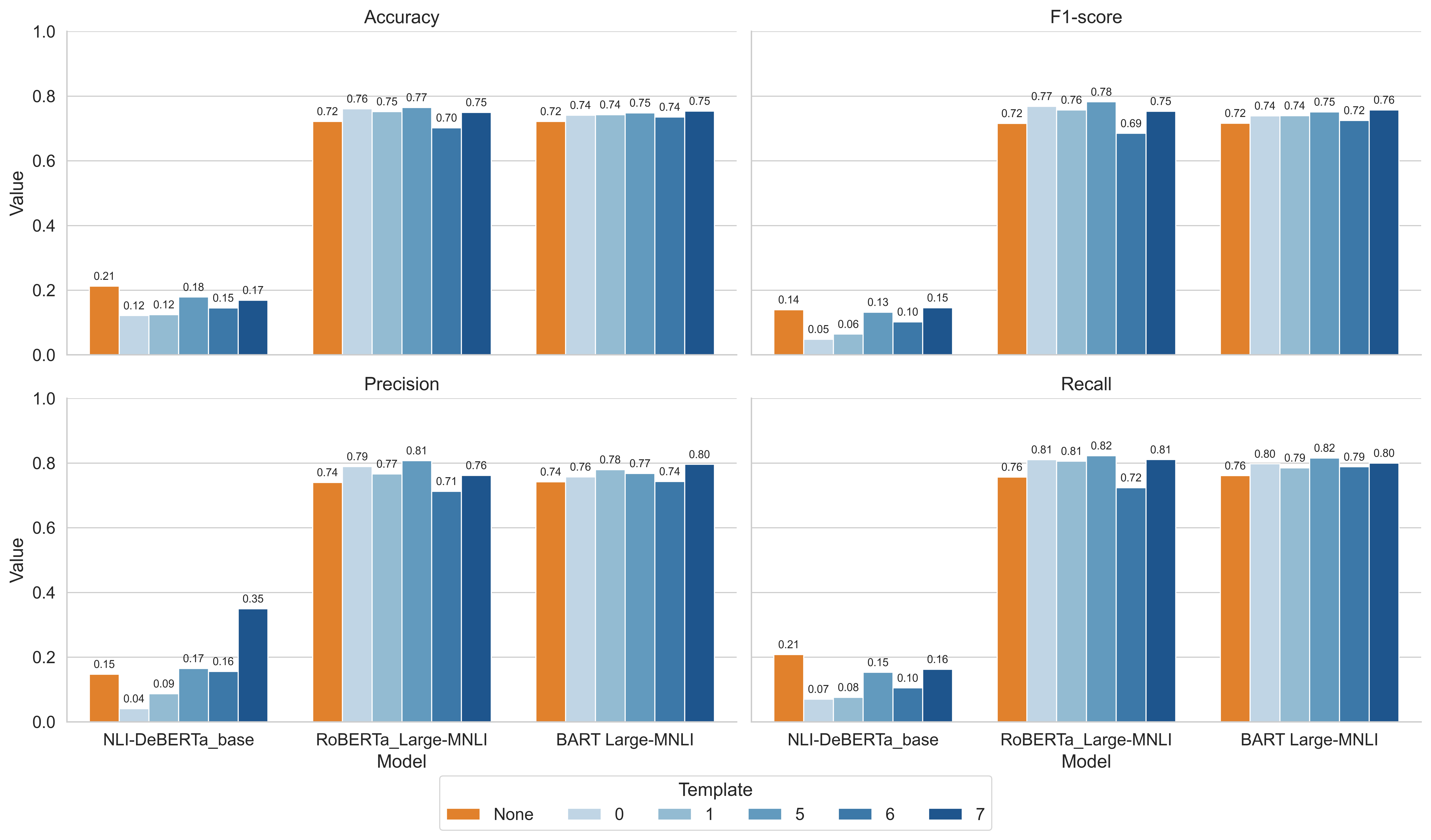}
\caption{Performance scores for MS-CXR dataset using natural language inference, disaggregated by template. Using overflow improves the large models, impacting the F1-score on $-3.43$, $2.78$, $2.18$ on \texttt{NLI-DeBERTa$_\texttt{base}$}, \texttt{RoBERTa$_\texttt{LARGE}$-MNLI}, and \texttt{BART Large-MNLI}, respectively. Template 5 is a good choice for the large models ($6.65$ for \texttt{RoBERTa$_\texttt{LARGE}$-MNLI} and $3.49$ for \texttt{BART Large-MNLI}), while template 6 is not ($-3.09$ for \texttt{RoBERTa$_\texttt{LARGE}$-MNLI} and $0.86$ for \texttt{BART Large-MNLI}). The choice of template can represent that \texttt{RoBERTa$_\texttt{LARGE}$-MNLI} is better than \texttt{BART Large-MNLI}, as they both have performances quite similar.}
\label{fig:ms_n}
\end{figure}

Results are depicted in \Cref{fig:ms_n,fig:transcriptions_n}. The mapping between the models and their ID is\\
\begin{center}
\begin{tabular}{lll}
m11: \texttt{NLI-DeBERTa$_\texttt{base}$} & m12: \texttt{RoBERTa$_\texttt{LARGE}$-MNLI} & m13: \texttt{BART Large-MNLI} \\
\end{tabular}
\end{center}

\subsubsection{Multiple choice question answering}
\begin{figure}[H]
\centering
\includegraphics[width=.73\textwidth]{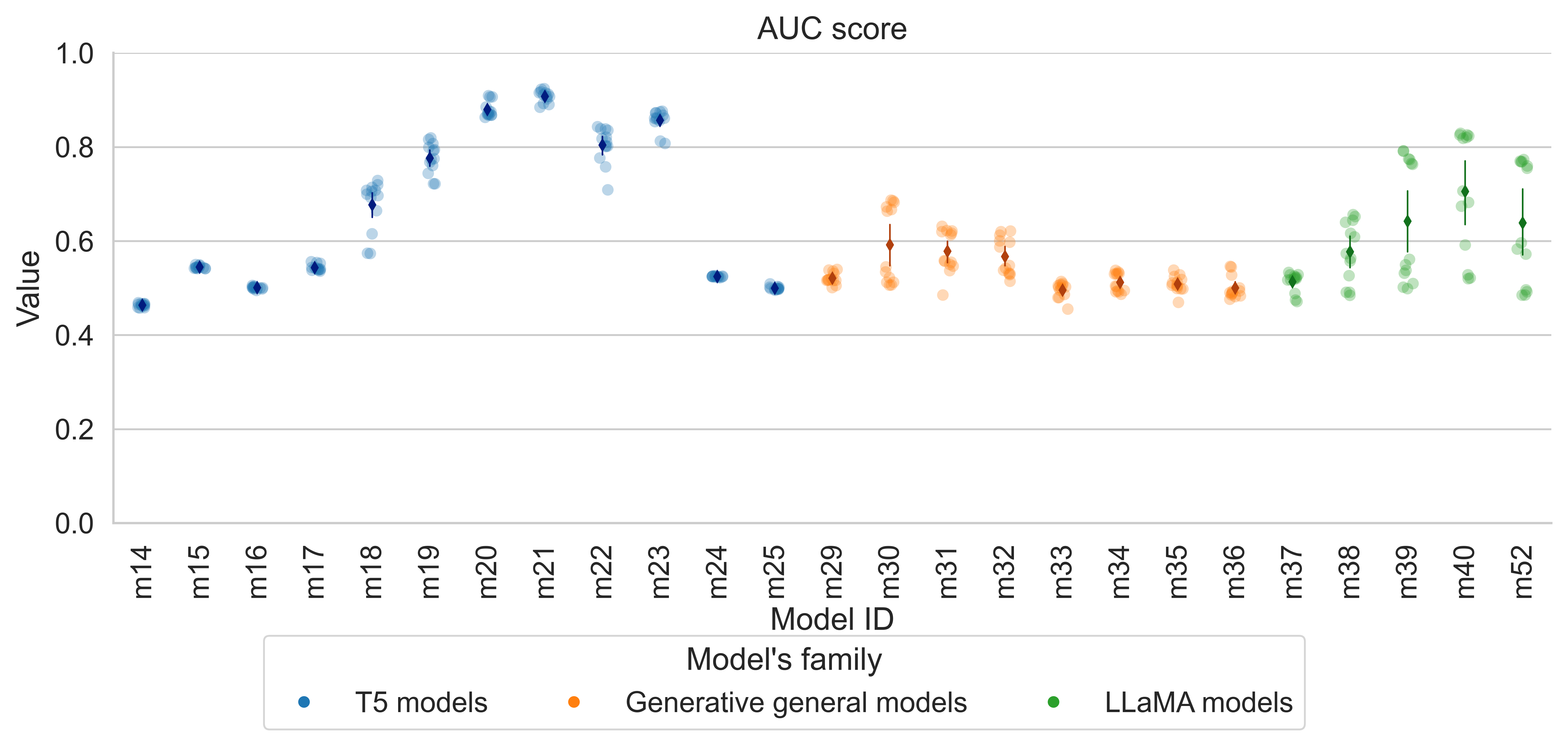}
\caption{Performance scores for transcriptions dataset using multiple-choice question answering. The T5 family of models represents text-to-text models, whereas the rest of the models represent autoregressive models. Instruction tuning usually leads to a performance increase, with an impact of $21.45$ on the AUC score when comparing the best performance per model. Considering the model size, text-to-text models perform similarly or better than their autoregressive counterparts, and \texttt{Flan-T5-XXL} is the best-performing model of all. Regarding the sensitivity of the models to the prompt used, the high sensitivity of the models is reflected in the clustering of the intra-model performance together with the visible variability of the latter.}
\label{fig:transcriptions_cg}
\end{figure}

\begin{figure}[H]
\centering
\includegraphics[width=.98\textwidth]{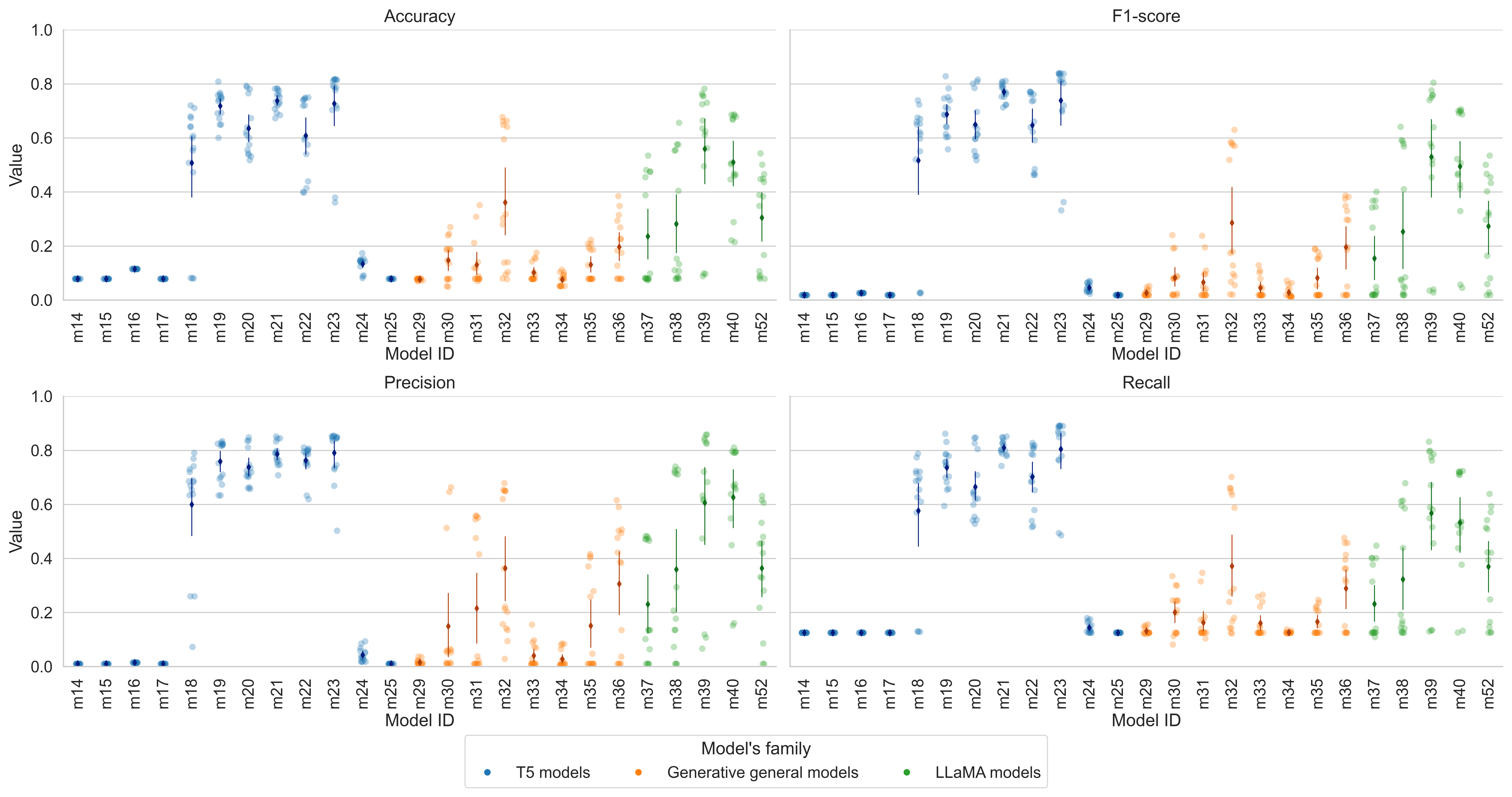}
\caption{Performance scores for MS-CXR dataset using multiple-choice question answering. The T5 family of models represents text-to-text models, whereas the rest of the models represent autoregressive models. Instruction-tuning leads to a performance increase, except for \texttt{Falcon-7B}, with an impact of $43.55$ on the AUC score when comparing the best performance per model. Taking into account the model size, text-to-text models perform similarly or better than their autoregressive counterparts, having that T0++ achieves the best scores for accuracy, F1-score, and recall. However, \texttt{Alpaca 7B} does it for precision. On the other hand, models that are not suitable for this task are T5 models, in all their size versions, and \texttt{ClinicalT5-large}.  Regarding the sensitivity of the models to the prompt used, the high sensitivity of the models is reflected in the clustering of the intra-model performance together with the visible variability of the latter.}
\label{fig:ms_cg}
\end{figure}

\begin{figure}[H]
\centering
\includegraphics[width=\textwidth]{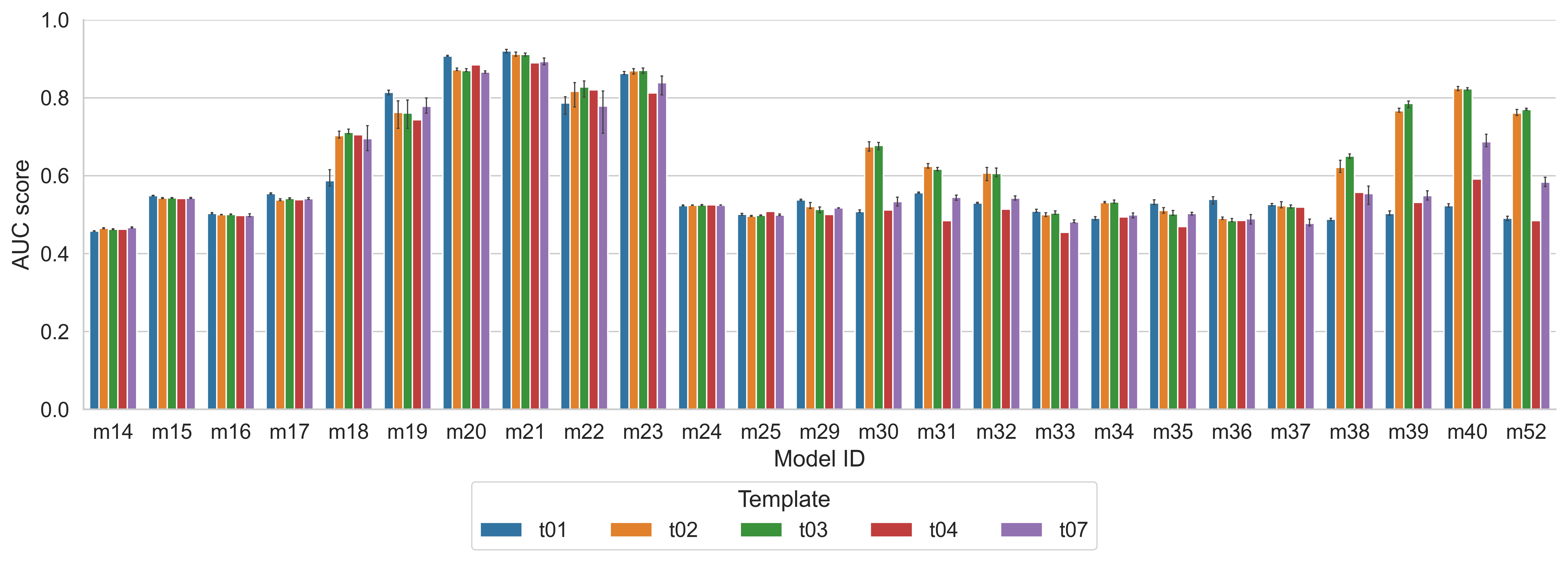}
\caption{Prompt analysis for the transcriptions dataset using multiple-choice question answering, disaggregated by template. Neither a template nor a question works best for all the models. Regarding the impact measured in terms of standard deviations, templates have an average impact of $3.84$. On its side, questions have an impact of $0.80$ on a template. Thus, the template's wording plays a more important role than the questions itself. In general, prompting has a great impact on performance.}
\label{fig:transcriptions_cp}
\end{figure}

\begin{figure}[H]
\centering
\includegraphics[width=\textwidth]{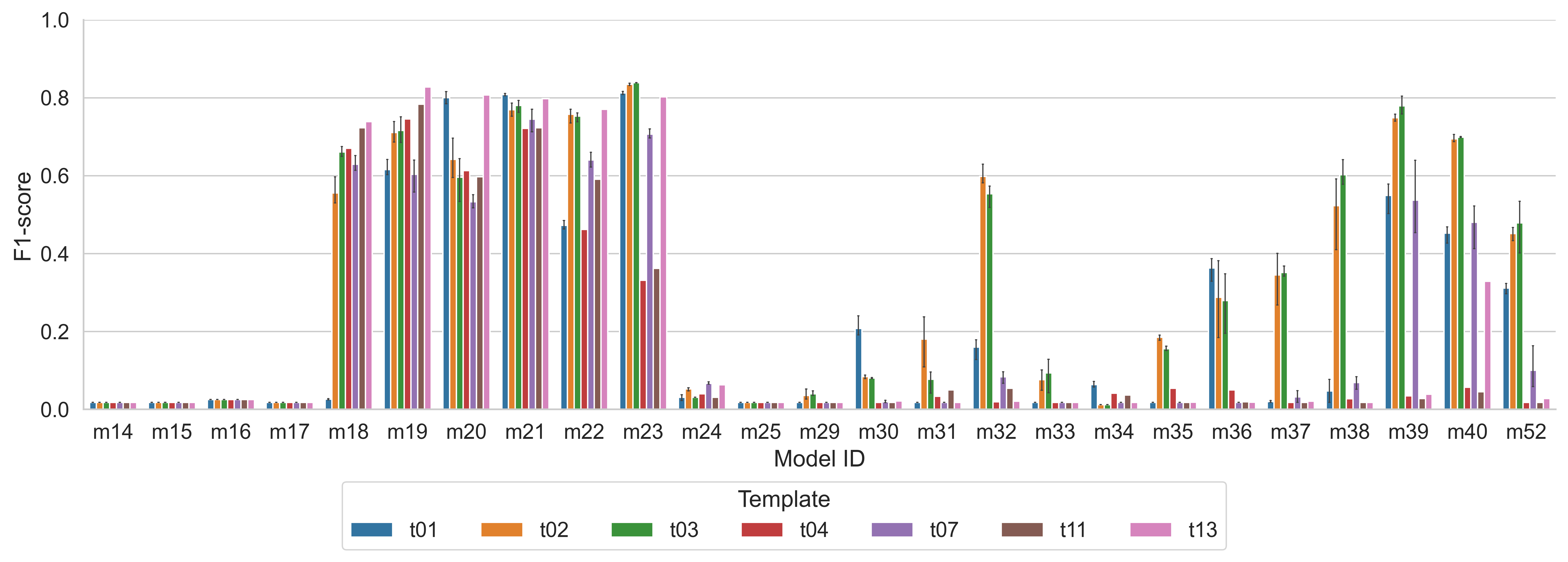}
\caption{Prompt analysis for the MS-CXR dataset using multiple-choice question answering, disaggregated by template. Neither a template nor a question works best for all the models. However, some templates are least suitable for this task and dataset. For example, template 4 is not good for the T0 and LLaMA family; template 7 is not for Flan-T5 models, and some generative no LLaMA models. Also, for some models, the role prompting strategy does not give good results, with emphasis in template 11. Regarding the impact, measured in terms of F1-score standard deviations, templates have an average impact of $11.02$. On its side, questions have an impact of $1.74$ on a template. Thus, the template's wording plays a more important role than the questions itself. In general, prompting greatly impacts performance, even decisive in terms of the ranking of models.}
\label{fig:ms_cp}
\end{figure}

Results are depicted in \Cref{fig:ms_cg,fig:transcriptions_cg,fig:ms_cp,fig:transcriptions_cp}. The mapping between the models and their ID is\\

\begin{center}
\begin{tabular}{llll}
m14: \texttt{T5-V1.1-Base} & m21: \texttt{Flan-T5-XXL} & m31: \texttt{GPT-J 6B} & m38: \texttt{LLaMA 2-7B}\\
m15: \texttt{T5-V1.1-Large} & m22: \texttt{T0-3B} & m32: \texttt{Instruct GPT-J} & m39: \texttt{Alpaca 7B}\\
m16: \texttt{T5-V1.1-3B} & m23: \texttt{T0++} & m33: \texttt{Falcon-7B} & m40: \texttt{LLaMA 2-CHAT-7B}\\
m17: \texttt{T5-V1.1-11B} & m24: \texttt{ClinicalT5-base} & m34: \texttt{Falcon-7B-Instruct} & m52: \texttt{MedAlpaca 7b}\\
m18: \texttt{Flan-T5-Base} & m25: \texttt{ClinicalT5-large} & m35: \texttt{MPT-7B} & \\
m19: \texttt{Flan-T5-Large} & m29: \texttt{Palmyra Base 5B} & m36: \texttt{MPT-7B-Instruct} & \\
m20: \texttt{Flan-T5-XL} & m30: \texttt{Camel 5B}& m37: \texttt{LLaMA-7B} & \\
\end{tabular}
\end{center}
\vspace{25pt}

\newpage
\subsection{Conditional text generation task}

\begin{figure*}[ht]
\centering
\includegraphics[width=\textwidth]{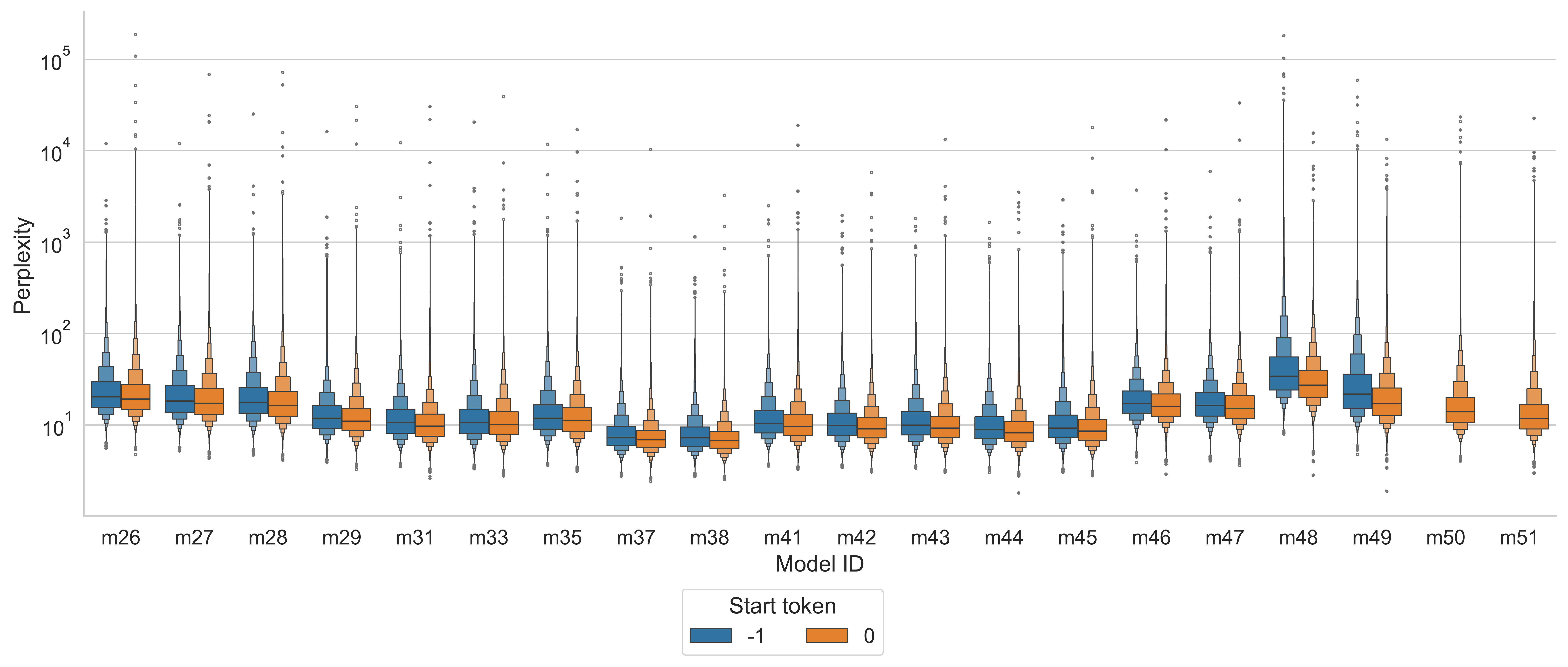}
\caption{Performance scores for the MIMIC-CXR dataset, disaggregated by BOS token usage. The perplexities are displayed in logarithmic scale. Not using the BOS token is beneficial for $77.78\%$ (14 out of 18) of the models, with the exceptions of the GPT-2 models (m26-m28) and \texttt{Palmyra Base 5B} (m29). Concerning outliers, their presence is quite strong. Moderate outliers, above quantile $0.75$ by $1.5$ times the IQR, represent between $7\%$ and $11\%$ of the data, with BioGPT models (m48-m49) having the highest percentages. Extreme outliers, above quantile $0.75$ by three times the IQR, make up between $4\%$ and $7\%$ of the data, with most models exhibiting percentages around $4\%$ and $5\%$.}
\label{fig:mimic_p}
\end{figure*}

Results are depicted in \Cref{fig:mimic_p}. The mapping between the models and their ID is\\

\begin{center}
\begin{tabular}{llll}
m26: \texttt{GPT-2 Medium} & m35: \texttt{MPT-7B} & m44: \texttt{OpenLLaMA 7Bv2} & m50: \texttt{Galactica 1.3B}\\
m27: \texttt{GPT-2 Large} & m37: \texttt{LLaMA-7B} & m45: \texttt{OpenLLaMA 13B} & m51: \texttt{Galactica 6.7B}\\
m28: \texttt{GPT-2 XL} & m38: \texttt{LLaMA 2-7B} & m46: \texttt{GPT-2-PubMed Medium} & \\
m29: \texttt{Palmyra Base 5B} & m42: \texttt{OpenLLaMA 3B} & m47: \texttt{GPT-2-PubMed Large} & \\
m31: \texttt{GPT-J 6B} & m42: \texttt{OpenLLaMA 3Bv2} & m48: \texttt{BioGPT-Large} & \\
m33: \texttt{Falcon-7B} & m43: \texttt{OpenLLaMA 7B} & m49: \texttt{BioGPT-Large} & \\
\end{tabular}
\end{center}

%% file: main.bbl
\begin{thebibliography}{100}
\providecommand{\url}[1]{#1}
\csname url@samestyle\endcsname
\providecommand{\newblock}{\relax}
\providecommand{\bibinfo}[2]{#2}
\providecommand{\BIBentrySTDinterwordspacing}{\spaceskip=0pt\relax}
\providecommand{\BIBentryALTinterwordstretchfactor}{4}
\providecommand{\BIBentryALTinterwordspacing}{\spaceskip=\fontdimen2\font plus
\BIBentryALTinterwordstretchfactor\fontdimen3\font minus \fontdimen4\font\relax}
\providecommand{\BIBforeignlanguage}[2]{{%
\expandafter\ifx\csname l@#1\endcsname\relax
\typeout{** WARNING: IEEEtran.bst: No hyphenation pattern has been}%
\typeout{** loaded for the language `#1'. Using the pattern for}%
\typeout{** the default language instead.}%
\else
\language=\csname l@#1\endcsname
\fi
#2}}
\providecommand{\BIBdecl}{\relax}
\BIBdecl

\bibitem{wiggings}
W.~F. Wiggins and A.~S. Tejani, ``{On the Opportunities and Risks of Foundation Models for Natural Language Processing in Radiology},'' \emph{Radiology: Artificial Intelligence}, vol.~4, no.~4, p. e220119, Jul. 2022.

\bibitem{shah}
N.~H. Shah, D.~Entwistle, and M.~A. Pfeffer, ``{Creation and Adoption of Large Language Models in Medicine},'' \emph{JAMA}, vol. 330, no.~9, pp. 866--869, Sep. 2023.

\bibitem{wei_ec}
J.~Wei \emph{et~al.}, ``{Emergent Abilities of Large Language Models},'' \emph{Transactions on Machine Learning Research}, 2022.

\bibitem{omiye}
J.~A. Omiye, H.~Gui, S.~J. Rezaei, J.~Zou, and R.~Daneshjou, ``Large language models in medicine: The potentials and pitfalls : A narrative review,'' \emph{Ann. Intern. Med.}, vol. 177, no.~2, pp. 210--220, Feb. 2024.

\bibitem{lievin}
V.~Liévin, C.~E. Hother, A.~G. Motzfeldt, and O.~Winther, ``Can large language models reason about medical questions?'' \emph{Patterns}, vol.~5, no.~3, p. 100943, 2024.

\bibitem{transformer}
A.~Vaswani \emph{et~al.}, ``{Attention is All you Need},'' in \emph{Advances in Neural Information Processing Systems}, I.~Guyon \emph{et~al.}, Eds., vol.~30.\hskip 1em plus 0.5em minus 0.4em\relax Curran Associates, Inc., 2017.

\bibitem{kuling}
G.~Kuling, B.~Curpen, and A.~L. Martel, ``{BI-RADS BERT and Using Section Segmentation to Understand Radiology Reports},'' \emph{Journal of Imaging}, vol.~8, no.~5, p. 131, 2022.

\bibitem{singhal}
K.~Singhal \emph{et~al.}, ``Large language models encode clinical knowledge,'' \emph{Nature}, vol. 620, no. 7972, pp. 172--180, Aug. 2023.

\bibitem{zhao}
W.~X. Zhao \emph{et~al.}, ``{A Survey of Large Language Models},'' 2023, \textit{arXiv:2303.18223 [cs.CL]}.

\bibitem{hoffmann}
J.~Hoffmann \emph{et~al.}, ``An empirical analysis of compute-optimal large language model training,'' in \emph{Advances in Neural Information Processing Systems}, S.~Koyejo, S.~Mohamed, A.~Agarwal, D.~Belgrave, K.~Cho, and A.~Oh, Eds., vol.~35.\hskip 1em plus 0.5em minus 0.4em\relax Curran Associates, Inc., 2022, pp. 30\,016--30\,030.

\bibitem{liu_rad}
Q.~Liu \emph{et~al.}, ``{Exploring the Boundaries of {GPT}-4 in Radiology},'' in \emph{Proceedings of the 2023 Conference on Empirical Methods in Natural Language Processing}.\hskip 1em plus 0.5em minus 0.4em\relax Association for Computational Linguistics, Dec. 2023, pp. 14\,414--14\,445.

\bibitem{phi3}
M.~Abdin \emph{et~al.}, ``{Phi-3 Technical Report: A Highly Capable Language Model Locally on Your Phone},'' 2024, \textit{arXiv:2404.14219 [cs.CL]}.

\bibitem{he}
K.~He \emph{et~al.}, ``{A Survey of Large Language Models for Healthcare: from Data, Technology, and Applications to Accountability and Ethics},'' 2024, \textit{arXiv:2310.05694 [cs.CL]}.

\bibitem{tang}
L.~Tang \emph{et~al.}, ``Evaluating large language models on medical evidence summarization,'' \emph{npj Digital Medicine}, vol.~6, no.~1, p. 158, Aug. 2023.

\bibitem{elmo}
M.~E. Peters \emph{et~al.}, ``{Deep Contextualized Word Representations},'' in \emph{Proceedings of the 2018 Conference of the North {A}merican Chapter of the Association for Computational Linguistics: Human Language Technologies, Volume 1 (Long Papers)}, M.~A. Walker, H.~Ji, and A.~Stent, Eds.\hskip 1em plus 0.5em minus 0.4em\relax Association for Computational Linguistics, Jun. 2018, pp. 2227--2237.

\bibitem{bert}
J.~Devlin, M.-W. Chang, K.~Lee, and K.~Toutanova, ``{BERT: Pre-training of Deep Bidirectional Transformers for Language Understanding},'' in \emph{Proceedings of the 2019 Conference of the North {A}merican Chapter of the Association for Computational Linguistics: Human Language Technologies, Volume 1 (Long and Short Papers)}, J.~Burstein, C.~Doran, and T.~Solorio, Eds.\hskip 1em plus 0.5em minus 0.4em\relax Association for Computational Linguistics, Jun. 2019, pp. 4171--4186.

\bibitem{flant5}
H.~W. Chung \emph{et~al.}, ``{Scaling Instruction-Finetuned Language Models},'' \emph{Journal of Machine Learning Research}, vol.~25, no.~70, pp. 1--53, 2024.

\bibitem{brown}
T.~Brown \emph{et~al.}, ``{Language Models are Few-Shot Learners},'' in \emph{Advances in Neural Information Processing Systems}, H.~Larochelle, M.~Ranzato, R.~Hadsell, M.~Balcan, and H.~Lin, Eds., vol.~33.\hskip 1em plus 0.5em minus 0.4em\relax Curran Associates, Inc., 2020, pp. 1877--1901.

\bibitem{gpt2}
\BIBentryALTinterwordspacing
A.~Radford, J.~Wu, R.~Child, D.~Luan, D.~Amodei, and I.~Sutskever, ``{Language Models are Unsupervised Multitask Learners},'' {OpenAI}, Tech. Rep., 2019. [Online]. Available: \url{https://api.semanticscholar.org/CorpusID:160025533}
\BIBentrySTDinterwordspacing

\bibitem{palm}
A.~Chowdhery \emph{et~al.}, ``{PaLM: Scaling Language Modeling with Pathways},'' \emph{Journal of Machine Learning Research}, vol.~24, no. 240, pp. 1--113, 2023.

\bibitem{rae}
J.~W. Rae \emph{et~al.}, ``{Scaling Language Models: Methods, Analysis \& Insights from Training Gopher},'' 2022, \textit{arXiv:2112.11446 [cs.CL]}.

\bibitem{wei}
J.~Wei \emph{et~al.}, ``{Finetuned Language Models are Zero-Shot Learners},'' in \emph{International Conference on Learning Representations}, 2022.

\bibitem{hendrycks}
D.~Hendrycks \emph{et~al.}, ``{Measuring Massive Multitask Language Understanding},'' in \emph{International Conference on Learning Representations}, 2021.

\bibitem{scalelaw}
J.~Kaplan \emph{et~al.}, ``{Scaling Laws for Neural Language Models},'' 2020, \textit{arXiv:2001.08361 [cs.LG]}.

\bibitem{scalelaw2}
Y.~Bahri, E.~Dyer, J.~Kaplan, J.~Lee, and U.~Sharma, ``Explaining neural scaling laws,'' \emph{Proceedings of the National Academy of Sciences}, vol. 121, no.~27, p. e2311878121, 2024.

\bibitem{galactica}
R.~Taylor \emph{et~al.}, ``{Galactica: A Large Language Model for Science},'' 2022, \textit{arXiv:2211.09085 [cs.CL]}.

\bibitem{llama}
H.~Touvron \emph{et~al.}, ``{LLaMA: Open and Efficient Foundation Language Models},'' 2023, \textit{arXiv:2302.13971 [cs.CL]}.

\bibitem{llama2}
------, ``{Llama 2: Open Foundation and Fine-Tuned Chat Models},'' 2023, \textit{arXiv:2307.09288 [cs.CL]}.

\bibitem{claude}
\BIBentryALTinterwordspacing
Antropic, ``{Introducing the next generation of Claude},'' Mar. 2024. [Online]. Available: \url{https://www.anthropic.com/news/claude-3-family}
\BIBentrySTDinterwordspacing

\bibitem{gemini}
{Gemini Team} \emph{et~al.}, ``{Gemini 1.5: Unlocking multimodal understanding across millions of tokens of context},'' 2024, \textit{arXiv:2403.05530 [cs.CL]}.

\bibitem{mistral}
A.~Q. Jiang \emph{et~al.}, ``{Mistral 7B},'' 2023, \textit{arXiv:2310.06825 [cs.CL]}.

\bibitem{bubeck}
S.~Bubeck \emph{et~al.}, ``{Sparks of Artificial General Intelligence: Early experiments with GPT-4},'' 2023, \textit{arXiv:2303.12712 [cs.CL]}.

\bibitem{nori}
H.~Nori, N.~King, S.~M. McKinney, D.~Carignan, and E.~Horvitz, ``{Capabilities of GPT-4 on Medical Challenge Problems},'' 2023, \textit{arXiv:2303.13375 [cs.CL]}.

\bibitem{mao}
R.~Mao, G.~Chen, X.~Zhang, F.~Guerin, and E.~Cambria, ``{GPTEval: A Survey on Assessments of ChatGPT and GPT-4},'' in \emph{Proceedings of the 2024 Joint International Conference on Computational Linguistics, Language Resources and Evaluation (LREC-COLING 2024)}.\hskip 1em plus 0.5em minus 0.4em\relax ELRA and ICCL, May 2024, pp. 7844--7866.

\bibitem{lopez}
J.~{López Espejel}, E.~H. Ettifouri, M.~S. {Yahaya Alassan}, E.~M. Chouham, and W.~Dahhane, ``{GPT-3.5, GPT-4, or BARD? Evaluating LLMs reasoning ability in zero-shot setting and performance boosting through prompts},'' \emph{Natural Language Processing Journal}, vol.~5, p. 100032, 2023.

\bibitem{liang}
P.~Liang \emph{et~al.}, ``{Holistic Evaluation of Language Models},'' \emph{Transactions on Machine Learning Research}, 2023.

\bibitem{liu__}
H.~Liu, R.~Ning, Z.~Teng, J.~Liu, Q.~Zhou, and Y.~Zhang, ``{Evaluating the Logical Reasoning Ability of ChatGPT and GPT-4},'' 2023, \textit{arXiv:2304.03439 [cs.CL]}.

\bibitem{schaeffer}
R.~Schaeffer, B.~Miranda, and S.~Koyejo, ``{Are Emergent Abilities of Large Language Models a Mirage?}'' in \emph{Advances in Neural Information Processing Systems}, A.~Oh, T.~Naumann, A.~Globerson, K.~Saenko, M.~Hardt, and S.~Levine, Eds., vol.~36.\hskip 1em plus 0.5em minus 0.4em\relax Curran Associates, Inc., 2023, pp. 55\,565--55\,581.

\bibitem{agrawal}
M.~Agrawal, S.~Hegselmann, H.~Lang, Y.~Kim, and D.~Sontag, ``Large language models are few-shot clinical information extractors,'' in \emph{Proceedings of the 2022 Conference on Empirical Methods in Natural Language Processing}, Y.~Goldberg, Z.~Kozareva, and Y.~Zhang, Eds.\hskip 1em plus 0.5em minus 0.4em\relax Association for Computational Linguistics, Dec. 2022, pp. 1998--2022.

\bibitem{sanh}
V.~Sanh \emph{et~al.}, ``{Multitask Prompted Training Enables Zero-Shot Task Generalization},'' in \emph{International Conference on Learning Representations}, 2022.

\bibitem{lampinen}
A.~Lampinen \emph{et~al.}, ``Can language models learn from explanations in context?'' in \emph{Findings of the Association for Computational Linguistics: EMNLP 2022}, Y.~Goldberg, Z.~Kozareva, and Y.~Zhang, Eds.\hskip 1em plus 0.5em minus 0.4em\relax Association for Computational Linguistics, Dec. 2022, pp. 537--563.

\bibitem{kojima}
T.~Kojima, S.~S. Gu, M.~Reid, Y.~Matsuo, and Y.~Iwasawa, ``{Large Language Models are Zero-Shot Reasoners},'' in \emph{Advances in Neural Information Processing Systems}, S.~Koyejo, S.~Mohamed, A.~Agarwal, D.~Belgrave, K.~Cho, and A.~Oh, Eds., vol.~35.\hskip 1em plus 0.5em minus 0.4em\relax Curran Associates, Inc., 2022, pp. 22\,199--22\,213.

\bibitem{wei_cot}
J.~Wei \emph{et~al.}, ``{Chain-of-Thought Prompting Elicits Reasoning in Large Language Models},'' in \emph{Advances in Neural Information Processing Systems}, S.~Koyejo, S.~Mohamed, A.~Agarwal, D.~Belgrave, K.~Cho, and A.~Oh, Eds., vol.~35.\hskip 1em plus 0.5em minus 0.4em\relax Curran Associates, Inc., 2022, pp. 24\,824--24\,837.

\bibitem{joshi}
M.~Joshi, E.~Choi, D.~Weld, and L.~Zettlemoyer, ``{TriviaQA: A Large Scale Distantly Supervised Challenge Dataset for Reading Comprehension},'' in \emph{Proceedings of the 55th Annual Meeting of the Association for Computational Linguistics (Volume 1: Long Papers)}, R.~Barzilay and M.-Y. Kan, Eds.\hskip 1em plus 0.5em minus 0.4em\relax Association for Computational Linguistics, Jul. 2017, pp. 1601--1611.

\bibitem{rev_scarcity}
I.~Jahan, M.~T.~R. Laskar, C.~Peng, and J.~X. Huang, ``A comprehensive evaluation of large language models on benchmark biomedical text processing tasks,'' \emph{Computers in Biology and Medicine}, vol. 171, p. 108189, 2024.

\bibitem{quantization}
B.~Jacob \emph{et~al.}, ``Quantization and training of neural networks for efficient integer-arithmetic-only inference,'' in \emph{Proceedings of the IEEE Conference on Computer Vision and Pattern Recognition (CVPR)}, June 2018.

\bibitem{quant1}
G.~Xiao, J.~Lin, M.~Seznec, H.~Wu, J.~Demouth, and S.~Han, ``{S}mooth{Q}uant: Accurate and efficient post-training quantization for large language models,'' in \emph{Proceedings of the 40th International Conference on Machine Learning}, ser. Proceedings of Machine Learning Research, A.~Krause, E.~Brunskill, K.~Cho, B.~Engelhardt, S.~Sabato, and J.~Scarlett, Eds., vol. 202.\hskip 1em plus 0.5em minus 0.4em\relax PMLR, July 2023, pp. 38\,087--38\,099.

\bibitem{quant2}
Y.~Tay, M.~Dehghani, D.~Bahri, and D.~Metzler, ``Efficient transformers: A survey,'' \emph{ACM Comput. Surv.}, vol.~55, no.~6, December 2022.

\bibitem{quant3}
S.~Li \emph{et~al.}, ``Evaluating quantized large language models,'' in \emph{Forty-first International Conference on Machine Learning}, 2024.

\bibitem{quant4}
S.~Kim \emph{et~al.}, ``Squeeze{LLM}: Dense-and-sparse quantization,'' in \emph{Forty-first International Conference on Machine Learning}, 2024.

\bibitem{quant5}
J.~Guo \emph{et~al.}, ``Compressing large language models by joint sparsification and quantization,'' in \emph{Forty-first International Conference on Machine Learning}, 2024.

\bibitem{quant6}
R.~Jin \emph{et~al.}, ``A comprehensive evaluation of quantization strategies for large language models,'' in \emph{Findings of the Association for Computational Linguistics ACL 2024}, L.-W. Ku, A.~Martins, and V.~Srikumar, Eds.\hskip 1em plus 0.5em minus 0.4em\relax Association for Computational Linguistics, August 2024, pp. 12\,186--12\,215.

\bibitem{lee}
P.~Lee, S.~Bubeck, and J.~Petro, ``{Benefits, Limits, and Risks of GPT-4 as an AI Chatbot for Medicine},'' \emph{New England Journal of Medicine}, vol. 388, no.~13, pp. 1233--1239, 2023.

\bibitem{fink}
M.~A. Fink, ``{Goße Sprachmodelle wie ChatGPT und GPT-4 für eine patientenzentrierte Radiologie [Large language models such as ChatGPT and GPT-4 for patient-centered care in radiology]},'' \emph{Radiologie}, vol.~63, no.~9, pp. 665--671, Sep. 2023.

\bibitem{lyu}
Q.~Lyu \emph{et~al.}, ``Translating radiology reports into plain language using {ChatGPT} and {GPT-4} with prompt learning: results, limitations, and potential,'' \emph{Visual Computing for Industry, Biomedicine, and Art}, vol.~6, no.~1, p.~9, May 2023.

\bibitem{adams}
L.~C. Adams \emph{et~al.}, ``{Leveraging GPT-4 for Post Hoc Transformation of Free-text Radiology Reports into Structured Reporting: A Multilingual Feasibility Study},'' \emph{Radiology}, vol. 307, no.~4, p. e230725, 2023.

\bibitem{bhayana}
R.~Bhayana, R.~R. Bleakney, and S.~Krishna, ``{GPT-4 in Radiology: Improvements in Advanced Reasoning},'' \emph{Radiology}, vol. 307, no.~5, p. e230987, 2023.

\bibitem{wu}
Z.~Wu \emph{et~al.}, ``{Exploring the Trade-Offs: Unified Large Language Models vs Local Fine-Tuned Models for Highly-Specific Radiology NLI Task},'' 2023, \textit{arXiv:2304.09138 [cs.CL]}.

\bibitem{ranjit}
M.~Ranjit, G.~Ganapathy, R.~Manuel, and T.~Ganu, ``{Retrieval Augmented Chest X-Ray Report Generation using OpenAI GPT models},'' in \emph{Proceedings of the 8th Machine Learning for Healthcare Conference}, ser. Proceedings of Machine Learning Research, K.~Deshpande \emph{et~al.}, Eds., vol. 219.\hskip 1em plus 0.5em minus 0.4em\relax PMLR, Aug. 2023, pp. 650--666.

\bibitem{mesko}
B.~Meskó and E.~J. Topol, ``{The imperative for regulatory oversight of large language models (or generative AI) in healthcare},'' \emph{npj Digital Medicine}, vol.~6, no.~1, p. 120, Jul. 2023.

\bibitem{gala}
D.~Gala and A.~N. Makaryus, ``{The Utility of Language Models in Cardiology: A Narrative Review of the Benefits and Concerns of ChatGPT-4},'' \emph{International Journal of Environmental Research and Public Health}, vol.~20, no.~15, 2023.

\bibitem{atallah}
S.~B. Atallah, N.~R. Banda, A.~Banda, and N.~A. Roeck, ``{How large language models including generative pre-trained transformer (GPT) 3 and 4 will impact medicine and surgery},'' \emph{Techniques in Coloproctology}, vol.~27, no.~8, pp. 609--614, Aug. 2023.

\bibitem{cheng}
K.~Cheng, Q.~Guo, Y.~He, Y.~Lu, S.~Gu, and H.~Wu, ``{Exploring the Potential of GPT-4 in Biomedical Engineering: The Dawn of a New Era},'' \emph{Annals of Biomedical Engineering}, vol.~51, no.~8, pp. 1645--1653, Aug. 2023.

\bibitem{biobert}
J.~Lee \emph{et~al.}, ``{BioBERT: a pre-trained biomedical language representation model for biomedical text mining},'' \emph{Bioinformatics}, vol.~36, no.~4, pp. 1234--1240, Sep. 2019.

\bibitem{scibert}
I.~Beltagy, K.~Lo, and A.~Cohan, ``{SciBERT: A Pretrained Language Model for Scientific Text},'' in \emph{Proceedings of the 2019 Conference on Empirical Methods in Natural Language Processing and the 9th International Joint Conference on Natural Language Processing (EMNLP-IJCNLP)}, K.~Inui, J.~Jiang, V.~Ng, and X.~Wan, Eds.\hskip 1em plus 0.5em minus 0.4em\relax Association for Computational Linguistics, Nov. 2019, pp. 3613--3618.

\bibitem{pubmedbert}
Y.~Gu \emph{et~al.}, ``{Domain-Specific Language Model Pretraining for Biomedical Natural Language Processing},'' \emph{ACM Trans. Comput. Heal.}, vol.~3, no.~1, pp. 2:1--2:23, Oct. 2022.

\bibitem{biomegatron}
H.~Shin \emph{et~al.}, ``{BioMegatron: Larger Biomedical Domain Language Model},'' in \emph{Proceedings of the 2020 Conference on Empirical Methods in Natural Language Processing (EMNLP)}, B.~Webber, T.~Cohn, Y.~He, and Y.~Liu, Eds.\hskip 1em plus 0.5em minus 0.4em\relax Association for Computational Linguistics, Nov. 2020, pp. 4700--4706.

\bibitem{scholarbert}
Z.~Hong, A.~Ajith, J.~G. Pauloski, E.~Duede, K.~Chard, and I.~T. Foster, ``{The Diminishing Returns of Masked Language Models to Science},'' in \emph{Findings of the Association for Computational Linguistics: {ACL} 2023}, A.~Rogers, J.~L. Boyd{-}Graber, and N.~Okazaki, Eds.\hskip 1em plus 0.5em minus 0.4em\relax Association for Computational Linguistics, 2023, pp. 1270--1283.

\bibitem{biogpt}
R.~Luo \emph{et~al.}, ``{BioGPT}: generative pre-trained transformer for biomedical text generation and mining,'' \emph{Briefings in Bioinformatics}, vol.~23, no.~6, Sep. 2022.

\bibitem{bioclinicalbert}
E.~Alsentzer \emph{et~al.}, ``{Publicly Available Clinical BERT Embeddings},'' in \emph{Proceedings of the 2nd Clinical Natural Language Processing Workshop}.\hskip 1em plus 0.5em minus 0.4em\relax Association for Computational Linguistics, Jun. 2019, pp. 72--78.

\bibitem{medalpaca}
T.~Han \emph{et~al.}, ``{MedAlpaca -- An Open-Source Collection of Medical Conversational AI Models and Training Data},'' 2023, \textit{arXiv:2304.08247 [cs.CL]}.

\bibitem{pmcllama}
C.~Wu, W.~Lin, X.~Zhang, Y.~Zhang, W.~Xie, and Y.~Wang, ``{PMC-LLaMA: toward building open-source language models for medicine},'' \emph{Journal of the American Medical Informatics Association: JAMIA}, vol.~31, no.~9, pp. 1833--1843, Apr. 2024.

\bibitem{medpalm2}
K.~Singhal \emph{et~al.}, ``{Towards Expert-Level Medical Question Answering with Large Language Models},'' 2023, \textit{arXiv:2305.09617 [cs.CL]}.

\bibitem{gatortron}
X.~Yang \emph{et~al.}, ``A large language model for electronic health records,'' \emph{npj Digital Medicine}, vol.~5, no.~1, p. 194, Dec. 2022.

\bibitem{gatortrongpt}
C.~Peng \emph{et~al.}, ``A study of generative large language model for medical research and healthcare,'' \emph{npj Digital Medicine}, vol.~6, no.~1, p. 210, Nov. 2023.

\bibitem{clinicalgpt}
G.~Wang, G.~Yang, Z.~Du, L.~Fan, and X.~Li, ``{ClinicalGPT: Large Language Models Finetuned with Diverse Medical Data and Comprehensive Evaluation},'' 2023, \textit{arXiv:2306.09968 [cs.CL]}.

\bibitem{huatuogpt}
H.~Zhang \emph{et~al.}, ``{HuatuoGPT, Towards Taming Language Model to Be a Doctor},'' in \emph{Findings of the Association for Computational Linguistics: EMNLP 2023}, H.~Bouamor, J.~Pino, and K.~Bali, Eds.\hskip 1em plus 0.5em minus 0.4em\relax Association for Computational Linguistics, Dec. 2023, pp. 10\,859--10\,885.

\bibitem{llavamed}
C.~Li \emph{et~al.}, ``{LLaVA-Med: Training a Large Language-and-Vision Assistant for Biomedicine in One Day},'' in \emph{Advances in Neural Information Processing Systems}, A.~Oh, T.~Naumann, A.~Globerson, K.~Saenko, M.~Hardt, and S.~Levine, Eds., vol.~36.\hskip 1em plus 0.5em minus 0.4em\relax Curran Associates, Inc., 2023, pp. 28\,541--28\,564.

\bibitem{medagi}
J.~Zhou, X.~Chen, and X.~Gao, ``{Path to Medical AGI: Unify Domain-specific Medical LLMs with the Lowest Cost},'' 2023, \textit{arXiv:2306.10765 [cs.AI]}.

\bibitem{ophglm}
W.~Gao \emph{et~al.}, ``{OphGLM: Training an Ophthalmology Large Language-and-Vision Assistant based on Instructions and Dialogue},'' 2023, \textit{arXiv:2306.12174 [cs.CV]}.

\bibitem{visualmedalpaca}
\BIBentryALTinterwordspacing
C.~Shu, B.~Chen, F.~Liu, Z.~Fu, E.~Shareghi, and N.~Collier, ``{Visual Med-Alpaca: A Parameter-Efficient Biomedical LLM with Visual Capabilities},'' 2013. [Online]. Available: \url{https://github.com/cambridgeltl/visual-med-alpaca}
\BIBentrySTDinterwordspacing

\bibitem{medflaming}
M.~Moor \emph{et~al.}, ``{Med-Flamingo: a Multimodal Medical Few-shot Learner},'' in \emph{Proceedings of the 3rd Machine Learning for Health Symposium}, ser. Proceedings of Machine Learning Research, S.~Hegselmann \emph{et~al.}, Eds., vol. 225.\hskip 1em plus 0.5em minus 0.4em\relax PMLR, Dec. 2023, pp. 353--367.

\bibitem{chexzero}
E.~Tiu, E.~Talius, P.~Patel, C.~P. Langlotz, A.~Y. Ng, and P.~Rajpurkar, ``{Expert-level detection of pathologies from unannotated chest X-ray images via self-supervised learning},'' \emph{Nature Biomedical Engineering}, vol.~6, no.~12, pp. 1399--1406, Dec. 2022.

\bibitem{zhou}
H.~Zhou \emph{et~al.}, ``{A Survey of Large Language Models in Medicine: Progress, Application, and Challenge},'' 2024, \textit{arXiv:2311.05112 [cs.CL]}.

\bibitem{soni}
S.~Soni and K.~Roberts, ``{Evaluation of Dataset Selection for Pre-Training and Fine-Tuning Transformer Language Models for Clinical Question Answering},'' in \emph{Proceedings of the 12th Language Resources and Evaluation Conference}, N.~Calzolari \emph{et~al.}, Eds.\hskip 1em plus 0.5em minus 0.4em\relax European Language Resources Association, May 2020, pp. 5532--5538.

\bibitem{lehman}
E.~Lehman \emph{et~al.}, ``Do we still need clinical language models?'' in \emph{Proceedings of the Conference on Health, Inference, and Learning}, ser. Proceedings of Machine Learning Research, B.~J. Mortazavi, T.~Sarker, A.~Beam, and J.~C. Ho, Eds., vol. 209.\hskip 1em plus 0.5em minus 0.4em\relax {PMLR}, Aug. 2023, pp. 578--597.

\bibitem{rev_long}
Y.~Li, R.~M. Wehbe, F.~S. Ahmad, H.~Wang, and Y.~Luo, ``{A comparative study of pretrained language models for long clinical text},'' \emph{Journal of the American Medical Informatics Association}, vol.~30, no.~2, pp. 340--347, 11 2022.

\bibitem{deberta}
\BIBentryALTinterwordspacing
{Sentence Transformers - Cross-Encoders}, ``cross-encoder/nli-deberta-base,'' 2021. [Online]. Available: \url{https://huggingface.co/cross-encoder/nli-deberta-base}
\BIBentrySTDinterwordspacing

\bibitem{roberta}
Y.~Liu \emph{et~al.}, ``{RoBERTa: A Robustly Optimized BERT Pretraining Approach},'' 2019, \textit{arXiv:1907.11692 [cs.CL]}.

\bibitem{sapbert}
F.~Liu, E.~Shareghi, Z.~Meng, M.~Basaldella, and N.~Collier, ``{Self-Alignment Pretraining for Biomedical Entity Representations},'' in \emph{Proceedings of the 2021 Conference of the North American Chapter of the Association for Computational Linguistics: Human Language Technologies {NAACL-HLT} 2021}, K.~Toutanova \emph{et~al.}, Eds.\hskip 1em plus 0.5em minus 0.4em\relax Association for Computational Linguistics, Jun. 2021, pp. 4228--4238.

\bibitem{biolord}
F.~Remy, K.~Demuynck, and T.~Demeester, ``{BioLORD: Learning Ontological Representations from Definitions for Biomedical Concepts and their Textual Descriptions},'' in \emph{Findings of the Association for Computational Linguistics: EMNLP 2022}, Y.~Goldberg, Z.~Kozareva, and Y.~Zhang, Eds.\hskip 1em plus 0.5em minus 0.4em\relax Association for Computational Linguistics, Dec. 2022, pp. 1454--1465.

\bibitem{t5}
C.~Raffel \emph{et~al.}, ``{Exploring the Limits of Transfer Learning with a Unified Text-to-Text Transformer},'' \emph{Journal of Machine Learning Research}, vol.~21, no. 140, pp. 1--67, 2020.

\bibitem{t5v1}
\BIBentryALTinterwordspacing
Google, ``google/t5-v1\_1,'' 2023. [Online]. Available: \url{https://huggingface.co/google}
\BIBentrySTDinterwordspacing

\bibitem{bart}
\BIBentryALTinterwordspacing
{AI at Meta}, ``facebook/bart-large-mnli,'' 2023. [Online]. Available: \url{https://huggingface.co/facebook/bart-large-mnli}
\BIBentrySTDinterwordspacing

\bibitem{t0}
V.~Sanh \emph{et~al.}, ``{Multitask Prompted Training Enables Zero-Shot Task Generalization},'' in \emph{The Tenth International Conference on Learning Representations, {ICLR} 2022}.\hskip 1em plus 0.5em minus 0.4em\relax OpenReview.net, 2022.

\bibitem{clinicalt5}
Q.~Lu, D.~Dou, and T.~Nguyen, ``{ClinicalT5: A Generative Language Model for Clinical Text},'' in \emph{Findings of the Association for Computational Linguistics: EMNLP 2022}, Y.~Goldberg, Z.~Kozareva, and Y.~Zhang, Eds.\hskip 1em plus 0.5em minus 0.4em\relax Association for Computational Linguistics, Dec. 2022, pp. 5436--5443.

\bibitem{palmyra}
\BIBentryALTinterwordspacing
{Writer Engineering team}, ``{Palmyra-base Parameter Autoregressive Language Model},'' Jan. 2023. [Online]. Available: \url{https://dev.writer.com}
\BIBentrySTDinterwordspacing

\bibitem{openllama}
\BIBentryALTinterwordspacing
X.~Geng and H.~Liu, ``{OpenLLaMA: An Open Reproduction of LLaMA},'' May 2023. [Online]. Available: \url{https://github.com/openlm-research/open_llama}
\BIBentrySTDinterwordspacing

\bibitem{camel}
\BIBentryALTinterwordspacing
{Writer Engineering team}, ``{Camel-5B InstructGPT},'' Apr. 2023. [Online]. Available: \url{https://dev.writer.com}
\BIBentrySTDinterwordspacing

\bibitem{gptj}
\BIBentryALTinterwordspacing
B.~Wang and A.~Komatsuzaki, ``{GPT-J-6B: A 6 Billion Parameter Autoregressive Language Model},'' May 2021. [Online]. Available: \url{https://github.com/kingoflolz/mesh-transformer-jax}
\BIBentrySTDinterwordspacing

\bibitem{igptj}
\BIBentryALTinterwordspacing
{NLP Cloud}, ``nlpcloud/instruct-gpt-j-fp16,'' 2023. [Online]. Available: \url{https://huggingface.co/nlpcloud/instruct-gpt-j-fp16}
\BIBentrySTDinterwordspacing

\bibitem{falcon}
E.~Almazrouei \emph{et~al.}, ``{The Falcon Series of Open Language Models},'' 2023, \textit{arXiv: 2311.16867 [cs.CL]}.

\bibitem{MPT}
\BIBentryALTinterwordspacing
{MosaicML NLP Team}, ``{Introducing MPT-7B: A New Standard for Open-Source, Commercially Usable LLMs},'' May 2023. [Online]. Available: \url{www.mosaicml.com/blog/mpt-7b}
\BIBentrySTDinterwordspacing

\bibitem{alpaca}
\BIBentryALTinterwordspacing
R.~Taori \emph{et~al.}, ``{Stanford Alpaca: An Instruction-following LLaMA model},'' GitHub, 2023. [Online]. Available: \url{https://github.com/tatsu-lab/stanford_alpaca}
\BIBentrySTDinterwordspacing

\bibitem{gpt2p}
\BIBentryALTinterwordspacing
Y.~Papanikolaou, ``healx/gpt-2-pubmed,'' 2020. [Online]. Available: \url{https://huggingface.co/healx}
\BIBentrySTDinterwordspacing

\bibitem{ms-cxr-1}
\BIBentryALTinterwordspacing
B.~Boecking \emph{et~al.}, ``{MS-CXR: Making the Most of Text Semantics to Improve Biomedical Vision-Language Processing (version 0.1)},'' PhysioNet, 2022. [Online]. Available: \url{https://doi.org/10.13026/b90j-vb87}
\BIBentrySTDinterwordspacing

\bibitem{ms-cxr-2}
\BIBentryALTinterwordspacing
------, ``{Making the Most of Text Semantics to Improve Biomedical Vision--Language Processing},'' in \emph{Computer Vision -- ECCV 2022: 17th European Conference}.\hskip 1em plus 0.5em minus 0.4em\relax Cham: Springer Nature Switzerland, Oct. 2022, pp. 1--21. [Online]. Available: \url{https://doi.org/10.1007/978-3-031-20059-5_1}
\BIBentrySTDinterwordspacing

\bibitem{PhysioNet}
A.~L. Goldberger \emph{et~al.}, ``{PhysioBank, PhysioToolkit, and PhysioNet}: Components of a new research resource for complex physiologic signals,'' \emph{\textnormal{Circulation [Online]}}, vol. 101, no.~23, pp. e215--e220, Jun. 2000.

\bibitem{mimic-cxr-1}
\BIBentryALTinterwordspacing
A.~E.~W. Johnson, T.~Pollard, R.~Mark, S.~Berkowitz, and S.~Horng, ``{The MIMIC-CXR Database},'' PhysioNet, 2019. [Online]. Available: \url{https://doi.org/10.13026/C2JT1Q}
\BIBentrySTDinterwordspacing

\bibitem{mimic-cxr-2}
\BIBentryALTinterwordspacing
A.~E.~W. Johnson \emph{et~al.}, ``{MIMIC-CXR}, a de-identified publicly available database of chest radiographs with free-text reports,'' \emph{Sci Data}, vol.~6, no.~1, p. 317, 2019. [Online]. Available: \url{https://doi.org/10.1038/s41597-019-0322-0}
\BIBentrySTDinterwordspacing

\bibitem{mimicrepo}
A.~E.~W. Johnson, D.~J. Stone, L.~A. Celi, and T.~J. Pollard, ``{The MIMIC Code Repository: enabling reproducibility in critical care research},'' \emph{Journal of the American Medical Informatics Association}, vol.~25, no.~1, pp. 32--39, 2018.

\bibitem{mimiccode}
\BIBentryALTinterwordspacing
A.~Johnson \emph{et~al.}, ``{MIT-LCP/mimic-code: MIMIC Code v2.2.1},'' Zenodo, Jul. 2022. [Online]. Available: \url{https://doi.org/10.5281/zenodo.6818823}
\BIBentrySTDinterwordspacing

\end{thebibliography}
